\documentclass[journal, 12pt, draftclsnofoot, onecolumn]{IEEEtran}

\usepackage{amsthm,amsmath,amsfonts,bbm}
\usepackage{array}
\usepackage{textcomp}
\usepackage{stfloats}
\usepackage{url}
\usepackage{verbatim}
\usepackage{graphicx}
\usepackage{cite}

\usepackage{amssymb}
\usepackage{bm}
\usepackage{multirow} 
\usepackage{booktabs}
\usepackage{lipsum}
\usepackage{enumerate}
\usepackage{multicol}
\usepackage{multirow}
\usepackage{subfigure}
\usepackage{stfloats}
\usepackage{mathrsfs}

\usepackage[ruled,linesnumbered]{algorithm2e}

\usepackage[colorlinks,
linkcolor=blue,
anchorcolor=blue,
citecolor=blue]{hyperref}

\usepackage{color}
\definecolor{rosered}{rgb}{0.894,0,0.498}

\usepackage{halloweenmath}
\allowdisplaybreaks[4]

\newtheorem{lemma}{Lemma}

\newtheorem{proposition}{Proposition}

\newtheorem{theorem}{Theorem}

\newtheorem{assumption}{Assumption}
\newtheorem{corollary}{Corollary}

\begin{document}
	
\title{Capacity-Constrained Delayed Feedback for Online Conformal Risk Control}

\author{
Bowen Wang,~\IEEEmembership{Student Member,~IEEE} and Osvaldo Simeone,~\IEEEmembership{Fellow,~IEEE}
\thanks{The work of O. Simeone was supported by an EPSRC Open Fellowship (EP/W024101/1), by the EPSRC (EP/X011852/1), and by the European Research Council (ERC) (Grant No. 101198347). Views and opinions expressed are however those of the author(s) only and do not necessarily reflect those of the European Union or the European Research Council. Neither the European Union nor the granting authority can be held responsible for them.}
\thanks{Bowen Wang is with the Department of Engineering, King’s College London, London WC2R 2LS, U.K. (e-mail: bowen.wang@kcl.ac.uk).}
\thanks{Osvaldo Simeone is with the Institute for Intelligent Networked Systems (INSI) at Northeastern University London, London E1 8PH, U.K. (e-mail: o.simeone@northeastern.edu).}
}

\maketitle

\begin{abstract}
	Online conformal risk control (OCRC) enables reliable uncertainty quantification in machine learning by converting model outputs into prediction sets with long-term risk guarantees.
	These guarantees, however, hinge on the assumption that the loss of each round is revealed immediately and losslessly.
	In practice, feedback may be delayed, and limited resources may prevent tracking all pending predictions, causing some feedback to be lost.
	To address this challenge, we develop capacity-aware OCRC (CA-OCRC), a framework that jointly determines how to allocate limited tracking feedback capacity and how to construct prediction sets.
	CA-OCRC is based on a randomized tracking policy that samples a proxy delay to decide whether, and for how long, feedback for the current round is tracked.
	CA-OCRC updates the prediction sets via inverse probability weighting, compensating for the possibility of feedback lost due to delays and limited tracking capacity.
	We analyze risk bounds for both arbitrary and stochastic delays, revealing a trade-off between receiving more feedback and avoiding capacity saturation.
	Building on these bounds, we optimize the proxy delay distribution, showing scaling laws for the tracking capacity that ensure the same long-term risk as OCRC with unlimited feedback.
	Experiments on medical imaging, natural language processing, computer vision, and wireless communications show the effectiveness of CA-OCRC as compared to existing OCRC mechanisms.
\end{abstract}


\section{Introduction}

\subsection{Context and Motivation}
\IEEEPARstart{A}{rtificial} intelligence (AI) models, such as large language models (LLMs), are increasingly deployed in applications ranging from finance and healthcare to robotics, autonomous systems, and wireless communications \cite{wisniewski2020application,lekeufack2024conformal,Osvaldo2026COMG,Park2026TPAMI}.
Particularly in high-stakes settings, their reliable deployment requires not only predictive accuracy, but also principled methods to quantify predictive uncertainty and to control the risks associated with predictions \cite{Wang2026TPAMI,Osvaldo2026JSAIT}.

Conformal prediction (CP) \cite{angelopoulos2024theoretical,Park2024TPAMI} quantifies uncertainty by transforming the outputs of a pre-trained model into prediction sets that contain the true label with user-specified probability.
However, coverage alone may not adequately capture the task-specific consequences of prediction errors.
Conformal risk control (CRC) \cite{angelopoulos2024conformal,Einbinder2025TPAMI,blot2024automatically} generalizes CP from the miscoverage indicator to general loss functions, guaranteeing that the expected loss incurred by the prediction sets does not exceed a user-specified target risk level.

The guarantees provided by CP and CRC rely on the exchangeability of calibration and test data \cite{angelopoulos2024theoretical,angelopoulos2024conformal}, which may be violated in sequential environments with evolving data distributions.
Online CP (OCP) \cite{gibbs2021adaptive,angelopoulos2024online,Xu2023TPAMI} and online CRC (OCRC) \cite{feldman2023achieving,zecchin2024localized,Zhu2025TSP,Binucci2025TSP} address this problem by adaptively updating the prediction sets based on feedback about the losses incurred in past rounds.
This way, OCP and OCRC ensure that the long-term average risk approaches an arbitrary target risk level for any, possibly adversarial, data sequence.

The guarantees of OCP and OCRC hinge on timely feedback, as the loss incurred at each round is assumed to be revealed to the predictor by the end of the round \cite{gibbs2021adaptive,angelopoulos2024online,Xu2023TPAMI,feldman2023achieving,zecchin2024localized,Zhu2025TSP,Binucci2025TSP}.
In practical applications, however, such feedback may not be immediately available and it may lost. 
For instance, in healthcare, the outcomes of medical decisions may only become known after subsequent examinations or patient follow-up \cite{cifra2021bridging}, while in open-domain question answering, the correctness of the answers returned to user questions may only be assessed after a slow human annotation process \cite{kwiatkowski2019natural,csaba2024label,pham2023continual}.

Tracking each prediction until its feedback becomes available may also require substantial monitoring and storage resources.
In healthcare, patient follow-up is constrained by limited monitoring capacity \cite{cifra2021bridging}, while, in question answering, only a limited number of questions can be submitted for human annotation at any time, owing to its high cost \cite{csaba2024label}.
Similar constraints arise in other applications, such as wireless communication systems, which must manage pending transmissions with finite memory and computational resources \cite{wiesmayr2025salad}, and online recommender and advertising systems, in which user conversions may occur up to a month after an item is displayed \cite{chapelle2014modeling}.
It may thus be infeasible to track all predictions with pending feedback simultaneously, resulting in the loss of some feedback signals.

\begin{figure}[t]
	\centering
	\includegraphics[width=0.8\linewidth]{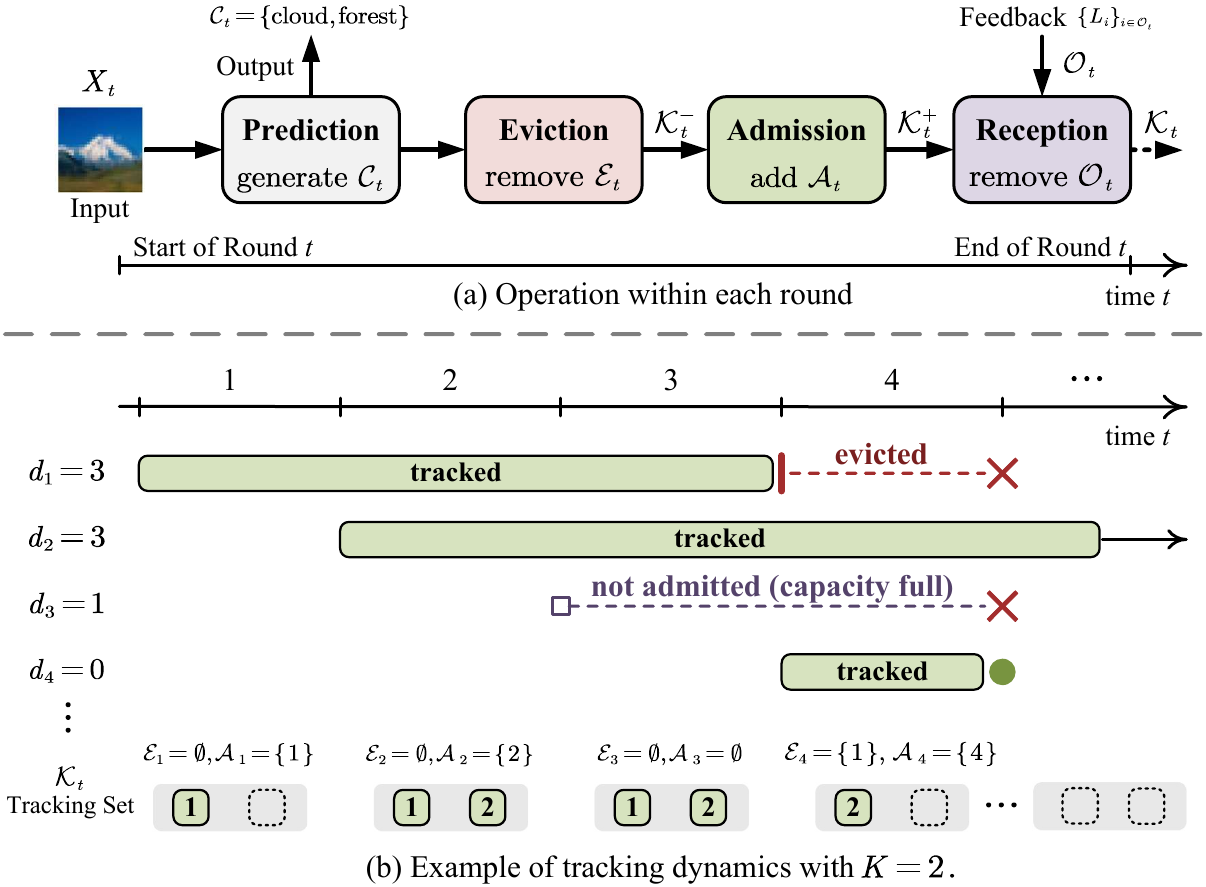}
	\caption{Illustration of sequential prediction with delayed feedback under a finite tracking capacity. 
	(a) Operations within round $t$: The predictor observes the input $X_t$ and generates the prediction set $\mathcal{C}_t$.
	It then evicts the rounds in the set $\mathcal{E}_t\subseteq\mathcal{K}_{t-1}$ from the current set $\mathcal{K}_{t-1}$ of tracked rounds, and decides whether to admit round $t$, i.e., $\mathcal{A}_t\in\{\emptyset,\{t\}\}$, subject to the capacity constraint $|\mathcal{K}_t^+|\le K$. 
	Finally, the feedback signals $\{L_i\}_{i\in\mathcal{O}_t}$ of the tracked rounds whose delays expire at round $t$ are received, and the rounds in $\mathcal{O}_t$ are removed from the tracking set.
	(b) Example with $K=2$, showing for each round $t$ the decisions $\mathcal{E}_t$ and $\mathcal{A}_t$, together with the resulting tracking set $\mathcal{K}_t$ at the end of the round. At $t=3$, the tracking set is full, so round~$t=3$ is not admitted, and its feedback, arriving at the end of round~4, is lost. At $t=4$, round~1 is evicted to admit round~4. 
	The feedback of round~4 is received immediately  ($d_4=0$), whereas that of round~1, arriving at the end of the same round, is lost ($d_1=3$). Round~2 remains tracked, awaiting its feedback at the end of round~5 ($d_2=3$).}
	\label{Fig_0}
\end{figure}

Motivated by this challenge, as illustrated in Fig.~\ref{Fig_0}, we investigate OCRC, and thus also the special case of OCP, under capacity-constrained delayed feedback.
The goal is to jointly design the prediction sets and the feedback tracking policy, which determines which rounds to track and for how long, so that the long-term risk approaches a user-specified target level.

\subsection{Related Work}

\subsubsection{Online Conformal Prediction with Delayed Feedback}
Reference \cite{Hallberg2024PLMR} studied delayed feedback in OCP by extending adaptive conformal inference (ACI) \cite{gibbs2021adaptive} to multi-step-ahead time-series forecasting, in which the feedback of a forecast is observed only after its forecast horizon has elapsed.
Subsequently, the work \cite{ding2025calibrated} extended multi-level quantile calibration to delayed feedback.
More recently, the paper \cite{halabi2026adaptive} analyzed ACI under fixed delays, establishing long-term coverage bounds by decomposing the delayed feedback into interleaved ACI instances.
In all these studies, however, the tracking capacity is implicitly unlimited, since all pending feedback signals are tracked until their arrival.
At the opposite extreme, intermittent OCP \cite{zhao2025conformalized,wang2025mirror}, in which the feedback of each round is either received immediately or lost, can be viewed as a special case of delayed feedback with unit tracking capacity (see Sec.~\ref{sec:iocrc}).

\subsubsection{Online Convex Optimization with Delayed Feedback}
Early work on delayed online convex optimization (OCO) \cite{orabona2019modern,shalev2025online} showed that online gradient descent (OGD) \cite{zinkevich2003online} achieves sublinear regret under fixed delays \cite{langford2009slow}. 
This analysis was extended to arbitrary delays in \cite{quanrud2015online}. 
Subsequent studies investigated dynamic regret in non-stationary environments \cite{wan2024non}, and unknown delays under general norms \cite{wu2024online}.
More recently, the work \cite{qiu2025exploiting} went beyond OGD and used the follow-the-regularized-leader (FTRL) approach to obtain improved regret bounds for strongly convex and exp-concave losses.
These studies assume that all delayed feedback can eventually be observed, without imposing tracking-capacity constraints.

More closely related to our work, reference \cite{ryabchenko2025capacity} introduced capacity-constrained delayed online learning for finite-action problems, characterizing the trade-off between tracking capacity and regret through tracking policies. 
This framework was subsequently extended to OCO in \cite{ryabchenko2026capacity} using randomized scheduling and delayed FTRL. 
However, these works focus on regret minimization, which does not directly imply long-term risk-control guarantees \cite{angelopoulos2025gradient}. 
The tracking policy proposed in this paper also builds on randomized scheduling, as in \cite{ryabchenko2026capacity}, but it targets long-term risk control rather than regret minimization.

Overall, existing OCP and OCRC schemes with delayed feedback do not account for a limited tracking capacity, while capacity-constrained OCO methods do not provide risk guarantees. This work addresses both limitations.

\subsection{Main Contributions}

The main contributions of this paper are summarized as follows:

\begin{itemize}
	\item \textbf{Capacity-aware OCRC:}
	We propose capacity-aware OCRC (CA-OCRC), an online framework for long-term risk control under delayed feedback and finite tracking capacity.
	CA-OCRC consists of two components: \textit{(i)} a tracking policy and \textit{(ii)} a threshold update for the prediction sets.
	The tracking policy samples a random proxy delay for each round, and it tracks the round, if capacity permits, until its feedback is received or its proxy delay expires.
	The threshold update strategy employs a two-level inverse probability weighting mechanism to compensate for permanently missing feedback.

	\item \textbf{Risk guarantees:}
	We first derive bounds on the long-term risk that hold for any delay (Theorem~\ref{thm:worstcase_miscoverage}), and then extend the analysis to stochastic delays, yielding distribution-dependent bounds (Theorem~\ref{thm_iid_miscoverage}).
	Both bounds reveal a trade-off in the design of the tracking policy: larger tracking probabilities yield more feedback, but they also increase the tracking load, and hence the probability of capacity saturation.

	\item \textbf{Proxy delay distribution design:}
	Building on the derived bounds, we optimize the proxy delay distribution.
	Under arbitrary delays, we show that the risk bound is minimized by a Bernoulli-type proxy delay distribution, under which CA-OCRC tracks each round until the maximum delay with a fixed probability, and does not track it otherwise.
	Under stochastic delays, we obtain an optimized proxy delay distribution in closed form based on the derived bound.
	With this distribution, CA-OCRC keeps tracking pending rounds by accounting on which feedback signals are more likely to arrive.

	\item \textbf{Scaling laws:}
	With the optimized proxy delay distributions, the risk gap of CA-OCRC decays with the time horizon $T$ as $\mathcal{O}((d_{\max}/T)^{K/(K+1)})$ under arbitrary delays, and as $\mathcal{O}((\mathbb{E}[d_t]/T)^{K/(K+1)})$ under stochastic delays, for any capacity $K\geq 1$, where $d_{\max}$ is the maximum delay and $\mathbb{E}[d_t]$ is the mean delay.
	Compared with the rate $\mathcal{O}(d_{\max}/T)$ of OCRC (see Sec. \ref{sec:ocrc}), which tracks all pending feedback, the capacity constraint thus reduces the exponent of the convergence rate from $1$ to $K/(K+1)$.
	In particular, a capacity of the order of $\ln(T/d_{\max})$, which may be much smaller than $d_{\max}$, ensures CA-OCRC to obtain the same rate as OCRC.

	\item \textbf{Experimental validation:}
	We validate CA-OCRC on tumor segmentation and open-domain question answering under stochastic delays, as well as on hierarchical image classification and beam selection under temporally correlated delays.
	The results show that CA-OCRC achieves risk levels close to the target, with faster convergence as the capacity increases, while keeping the tracking load below the capacity.
	Compared with capacity-constrained benchmarks, CA-OCRC achieves more reliable risk control, while the distribution-aware proxy design further improves capacity utilization, convergence speed, and prediction-set efficiency under stochastic delays.
\end{itemize}

\subsection{Organization}
The remainder of this paper is organized as follows.
Sec.~\ref{sec:2} formulates the sequential prediction problem with delayed feedback and finite tracking capacity. Sec.~\ref{sec:3} reviews OCRC and intermittent OCRC (I-OCRC).
Sec.~\ref{Sec:4} introduces CA-OCRC, and Secs.~\ref{sec:5} and~\ref{sec:6} give risk guarantees and proxy delay designs for arbitrary and stochastic delays, respectively.
Sec.~\ref{sec:7} presents numerical results, and Sec.~\ref{sec:8} concludes.

\section{Problem Formulation}\label{sec:2}

In this section, we formulate the problem of sequential prediction with delayed feedback under finite tracking capacity, which is outlined in Fig. \ref{Fig_0}.

\subsection{Setting}
We consider sequential prediction over rounds $t=1,2,\ldots,T$.
At each round $t$, the environment generates an input-label pair $(X_t,Y_t) \in \mathcal{X} \times \mathcal{Y}$, where $\mathcal{X}$ and $\mathcal{Y}$ denote the input and label spaces, respectively.
The data sequence $\{(X_t,Y_t)\}_{t=1}^{T}$ is arbitrary, and we make no probabilistic assumptions such as stationarity or exchangeability.

The predictor has access to a pre-trained model $f(\cdot)$. 
At the beginning of round $t$, the predictor observes the current input $X_t$ and, using the pre-trained model $f(\cdot)$, returns a prediction set $\mathcal{C}_t\subseteq\mathcal{Y}$ for the unknown label $Y_t$. 
The performance of the set prediction is measured by the loss
\begin{equation}\label{eq:loss}
	L_t = \mathcal{L}(\mathcal{C}_t, Y_t) ,
\end{equation}
where the loss function $\mathcal{L}(\mathcal{C}, y) \in [0, 1]$ quantifies the cost incurred by the set $\mathcal{C} \subseteq \mathcal{Y}$ when the true label is $y$.
A standard example is the miscoverage indicator
\begin{equation}\label{eq:mis-indicator}
    \mathcal{L}(\mathcal{C}_t, Y_t) = \mathbbm{1}\{Y_t\notin\mathcal{C}_t\} ,
\end{equation}
which equals one when the true label $Y_t$ is not contained in the set $\mathcal{C}_t$, and zero otherwise.
Other examples include the false negative rate (FNR) in multi-label classification problems, and weighted variants of miscoverage \cite{feldman2023achieving,zecchin2024localized,angelopoulos2024conformal}.

Prior art assumed that feedback about the loss $L_t$ was available with zero delay\cite{angelopoulos2024online,gibbs2021adaptive,feldman2023achieving,zecchin2024localized}, i.e., at the end of round $t$, though it could possibly be lost \cite{zhao2025conformalized,wang2025mirror} or distorted \cite{Matteo2026JSAIT}.
In contrast, we assume that the predictor can access the loss $L_t$:
\textit{(i)} after $d_t \in \mathbb{Z}_{\geq 0}$ time steps, i.e., at the end of round $t+d_t$;
and \textit{(ii)} only provided that feedback $L_t$ was tracked throughout the time interval $[t, t+d_t]$.
As extreme cases, setting $d_t=0$ recovers the standard scenario with feedback available at the end of the current round, while the delays $d_t \ge T - t$ represent a feedback free scenario in which feedback does not become available before the end of the considered time horizon $T$.

At any given round $t$, the predictor can only track at most $K\in\mathbb{Z}_{\geq 0}$ of the feedback signals $\{L_{t'}\}$ corresponding to past time instants $t' \in \{1, \dots, t\}$. 
As mentioned, feedback signals must be tracked continuously in order to be eligible to be received after the corresponding delay.
We focus on the nontrivial case $K \le d_{\max}$, where $d_{\max}$ is the largest possible feedback delay. 
Settings with $K \geq d_{\max}+1$, in fact, permit tracking all the feedback signals, since at any round $t$ only the $d_{\max}+1$ rounds $t-d_{\max},\ldots,t$ can have pending feedback.

The tracked feedback signals at the end of round $t$ are defined by a \emph{tracking set} $\mathcal{K}_t \subseteq \{1,...,t\}$, which evolves recursively over time via eviction and admission policies to be optimized. 
To elaborate, initialize the set as $\mathcal{K}_0 = \emptyset$.
As shown in Fig. \ref{Fig_0}, at any round $t$, the tracking set evolves from $\mathcal{K}_{t-1}$ to $\mathcal{K}_t$ through the following steps.

\subsubsection{Eviction} The predictor evicts a subset $\mathcal{E}_t\subseteq\mathcal{K}_{t-1}$ of previously tracked rounds, yielding the post-eviction set
\begin{equation}\label{eq:tracking-minus}
	\mathcal{K}_t^{-} = \mathcal{K}_{t-1}\setminus\mathcal{E}_t, \text{ with } \mathcal{E}_t\subseteq\mathcal{K}_{t-1} .
\end{equation}
Once a round $i \in \mathcal{E}_t$ is evicted, its feedback $L_{i}$ can no longer be received by the predictor.

\subsubsection{Admission} 
If capacity permits, i.e., if $|\mathcal{K}_t^-| < K$, the predictor can next decide to track the current round $t$.
Formally, the admission set at round $t$, $\mathcal{A}_t$, is empty if capacity is saturated, i.e., if $|\mathcal{K}_t^-| = K$, while it may include $t$ otherwise. 
Accordingly, the post-eviction tracking set $\mathcal{K}_t^{-}$ is updated as
\begin{equation}
	\mathcal{K}_t^{+} = \mathcal{K}_t^{-}\cup \mathcal{A}_t \text{ with } \mathcal{A}_t \in \{\emptyset , \{t\} \} ,
	\label{eq:tracking-plus}
\end{equation}
where $\mathcal{A}_t = \emptyset$ if $|\mathcal{K}_t^-| = K$.

\subsubsection{Feedback Reception}
At the end of round $t$, the predictor receives all tracked feedback signals $L_{t'}$, with $t' \in \mathcal{K}_t^+$, for which the delay $d_{t'}$ satisfies the equality $t' + d_{t'} = t$. 
Accordingly, the set of observed feedback signals is
\begin{equation}\label{eq:physical-arrivals}
	\mathcal{O}_t
	= \{i\leq t : i+d_i=t , \text{ and } i \in \mathcal{K}_t^+ \}.
\end{equation}
Note that, by this convention, feedback $L_t$ with a zero delay $d_t=0$ is observed only if round $t$ is admitted, i.e., if $\mathcal{A}_t = \{ t \}$.
The rounds in $\mathcal{O}_t$ are then removed from the tracking set $\mathcal{K}_t^{+}$, yielding the updated tracking set
\begin{equation}\label{eq:tracking-end}
	\mathcal{K}_t
	= \mathcal{K}_t^{+} \setminus \mathcal{O}_t.
\end{equation}

Overall, at the beginning of round $t$, the predictor has access to the tracking set $\mathcal{K}_{t-1}$, to the feedback signals $\mathcal{O}_{1:t-1} = \bigcup_{t'=1}^{t-1} \mathcal{O}_{t'}$ received so far, as well as to the input $X_t$ and to the prediction $f(X_t)$. 
The predictor can use this information in an arbitrary way to design the prediction set $\mathcal{C}_t$, of the eviction set $\mathcal{E}_t$, and of the admission set $\mathcal{A}_t$.

\subsection{Delay Models}
We consider the following two alternative models for the feedback delays.
\begin{assumption}\label{assum:delay}
The feedback delays satisfy one of the following alternatives:
\begin{enumerate}
	\item[\textnormal{(a)}] \textnormal{Arbitrary Delays:} The feedback delays $d_{1} , \ldots , d_{T}$ are arbitrary, while satisfying the upper bound
	\begin{equation}\label{eq_assum_1a}
		0 \le d_t \le d_{\max}, \; t=1,\ldots,T ,
	\end{equation}
	for a known constant $d_{\max} \geq 0$.

	\item[\textnormal{(b)}] \textnormal{Stochastic Delays:} The feedback delays $d_{1} , \ldots , d_{T}$ are independent and identically distributed, i.e.,
	\begin{equation}
		d_t \sim G ,
	\end{equation}
	according to a survival function
	\begin{equation}\label{eq_assum_1b}
		G(a) = {\Pr}_G [d_t \ge a] , \text{ for } a \in \{0, 1, \dots, d_{\max}\} .
	\end{equation}
\end{enumerate}
\end{assumption}

\subsection{Design Goal}\label{sec:design-goal}

For a target risk level $\alpha\in(0,1)$, the goal is to jointly design a policy that maps the observations $\mathcal{O}_{1:t-1}$, the input $X_t$, and the prediction $f(X_t)$ to a prediction set $\mathcal{C}_t$, an eviction set $\mathcal{E}_t \subseteq \mathcal{K}_{t-1}$ in \eqref{eq:tracking-minus}, and an admission set $\mathcal{A}_t$ in \eqref{eq:tracking-plus}.
The design objective is to ensure the long-term risk gap condition
\begin{equation}\label{eq_miscov}
    \lim_{T \to \infty} \left\{ \mathrm{RiskGap}(T) = \left| \mathbb{E} \left[ \frac{1}{T}\sum_{t=1}^{T} L_t \right] - \alpha  \right| \right\} = 0 ,
\end{equation}
where the expectation is taken with respect to the internal randomness of the policy under Assumption~\ref{assum:delay}(a), and also with respect to the stochastic delays under Assumption~\ref{assum:delay}(b).
Note that the long-term risk in \eqref{eq_miscov} is evaluated over all prediction rounds, including rounds whose labels are never observed by the predictor.

The following lemma connects the two settings with arbitrary and stochastic delays. 
\begin{lemma}\label{lem:arb-implies-iid}
	Suppose that, for some deterministic constant $\varepsilon$, the upper bound 
	\begin{equation}\label{eq_lemma_1}
		\mathrm{RiskGap}(T) \le \varepsilon
	\end{equation}
	is achievable under Assumption~\ref{assum:delay}(a).
	Then, the same bound \eqref{eq_lemma_1} also holds under Assumption~\ref{assum:delay}(b).
\end{lemma}

Lemma \ref{lem:arb-implies-iid} shows that a uniform guarantee for arbitrary bounded delays automatically yields the same guarantee under bounded stochastic delays.
In fact, the same policy that ensures the upper bound \eqref{eq_lemma_1} under Assumption~\ref{assum:delay}(a) automatically also satisfies the bound \eqref{eq_lemma_1} under Assumption~\ref{assum:delay}(b).
That said, as we will see in Sec. \ref{sec:6}, knowledge of the delay distribution $G$ under Assumption~\ref{assum:delay}(b) can probably lead to more efficient, tailored policies, yielding distribution-dependent bounds on the risk gap.

\section{Background: Online Conformal Risk Control}\label{sec:3}

In this section, we review two existing solutions that address extreme special cases of the setting described in Sec. \ref{sec:2}, namely online conformal risk control (OCRC) \cite{feldman2023achieving,zecchin2024localized}, which assumes an infinite-capacity predictor and can handle arbitrary finite delays, and intermittent OCRC (I-OCRC), i.e., the direct extension of intermittent OCP \cite{zhao2025conformalized,wang2025mirror} to general losses, which applies to a unit-capacity predictor under stochastic delays.

\subsection{Online Conformal Risk Control}\label{sec:ocrc}
OCRC assumes no capacity constraints, i.e., $K \geq T$.
Accordingly, it tracks all rounds, carrying out no evictions and admitting every current round, i.e., $\mathcal{E}_t=\emptyset$ and $\mathcal{A}_t=\{t\}$ for all rounds $t$.
The prediction set $\mathcal{C}_t$ is constructed as \cite{angelopoulos2024online,gibbs2021adaptive}
\begin{equation}\label{eq_pred_set}
    \mathcal{C}_t(X_t;r_t)
    =\{y\in\mathcal{Y}:S_t(X_t,y)\le r_t\} ,
\end{equation}
where $S_t(X_t,y):\mathcal{X}\times\mathcal{Y}\to\mathbb{R}$ is a score based on the pre-trained model $f(\cdot)$, with a larger score indicating poorer agreement between a candidate label $y$ and the prediction $f(X_t)$, and $r_t$ is a threshold.
Typical examples include the residual score $S_t(X_t, y) = |y - f(X_t)|$ for regression and $S_t(X_t, y) = 1 - \log f_y(X_t)$ for classification, where $f_y(X_t)$ is the predictive probability for label $y$ given input $X_t$. 
Starting from an initial threshold $r_t$, the threshold is updated as
\begin{equation}\label{eq:unconstrained-update}
    r_{t+1}
    =r_t+\eta\sum_{t': t' + d_{t'} = t}(L_{t'}-\alpha),
\end{equation}
where $\eta > 0$ is a learning rate.
Note that the original analysis in \cite{angelopoulos2024online,gibbs2021adaptive,feldman2023achieving,zecchin2024localized} assumed zero delays, so that the update reduces to $r_{t+1}=r_t+\eta(L_t-\alpha)$.

The update \eqref{eq:unconstrained-update} is a direct extension of the standard OCRC \cite{feldman2023achieving,zecchin2024localized} to delayed feedback, and its analysis relies on the following assumptions \cite{angelopoulos2024online,gibbs2021adaptive,feldman2023achieving}.

\begin{assumption}\label{assum:score}
	The score function satisfies $S_t(X_t,y)\in[0,B]$ for every label $y\in\mathcal{Y}$ and every round $t$, where $B<\infty$.
\end{assumption}

\begin{assumption}\label{assum:loss}
	The loss function satisfies $\mathcal{L}(\mathcal{Y}, y) = 0$ and $\mathcal{L}(\emptyset, y) = 1$ for every label $y \in \mathcal{Y}$, i.e., the full set $\mathcal{C}_t = \mathcal{Y}$ incurs no loss, while the empty set $\mathcal{C}_t = \emptyset$ incurs the maximum loss.
\end{assumption}

By Assumption~\ref{assum:score}, the prediction set \eqref{eq_pred_set} equals the full set $\mathcal{Y}$ for $r_t\ge B$ and the empty set $\emptyset$ for $r_t<0$, so that, by Assumption~\ref{assum:loss}, the corresponding losses equal $0$ and $1$, respectively, irrespective of the true label $Y_t$.
Under these assumptions, the following guarantee can be proved in a manner similar to \cite{angelopoulos2024online,gibbs2021adaptive,feldman2023achieving}.

\begin{lemma}\label{lem:unconstrained-coverage}
Under Assumption \ref{assum:delay}(a), Assumption \ref{assum:score}, and Assumption~\ref{assum:loss} with $r_1\in[0,B]$, the risk gap of OCRC satisfies the upper bound
\begin{equation}\label{eq:unconstrained-coverage}
	\mathrm{RiskGap}(T)
	\le \frac{B}{\eta T}+\frac{d_{\max}+1}{T}
\end{equation}
for any sequence of data points and delays.
\end{lemma}
\begin{proof}
	See Supplementary Material Appendix~\ref{app:lem_baselines}.
\end{proof}

Accordingly, the risk gap of OCRC vanishes as $T \to \infty$ whenever the maximum delay $d_{\max}$ is constant or grows sublinearly with $T$, i.e., $d_{\max} = o(T)$.

\subsection{Intermittent OCRC}\label{sec:iocrc}

At the opposite extreme, consider a setting with unit tracking capacity, i.e., $K=1$.
The eviction and admission sets are chosen as $\mathcal{E}_t=\mathcal{K}_{t-1}$ and $\mathcal{A}_t=\{t\}$, respectively, so that only the current round is tracked.
Under Assumption \ref{assum:delay}(b), the current feedback is thus received with probability 
\begin{equation}
	{\Pr}_G [d_t = 0] = G(0) - G(1)
\end{equation}
and is otherwise lost.
I-OCRC produces the prediction set \eqref{eq_pred_set} with the threshold updated as \cite{zhao2025conformalized,wang2025mirror}
\begin{equation}\label{eq_i_ocp}
	r_{t+1}
	= r_t + \eta \frac{\mathbbm{1}\{ d_t = 0 \} }{ G(0) - G(1) } (L_t-\alpha) .
\end{equation}
The factor $(G(0) - G(1))^{-1} = ({\Pr}_G[d_t = 0])^{-1}$ applies a form of inverse probability weighting \cite{elvira2021advances}, accounting for the probability of feedback loss.

\begin{lemma}[\hspace{-0.05em}\cite{wang2025mirror,zhao2025conformalized}]\label{lemm_i_ocp}
Under Assumption~\ref{assum:delay}(b), Assumption \ref{assum:score}, and Assumption~\ref{assum:loss} with $r_1\in[0,B]$, the risk gap of I-OCRC is upper bounded by
\begin{equation}\label{eq_unit_coverage}
	\mathrm{RiskGap}(T)
	\le \frac{B}{\eta T}+\frac{1}{ ( G(0) - G(1) ) T} .
\end{equation}
\end{lemma}
\begin{proof}
	See Supplementary Material Appendix~\ref{app:lem_baselines}, where the analysis in \cite{wang2025mirror,zhao2025conformalized} is extended to general losses.
\end{proof}

Accordingly, the risk gap of I-OCRC vanishes as $T \to \infty$ whenever immediate feedback occurs with a positive probability, i.e., ${\Pr}_G[d_t = 0] > 0$, although the bound \eqref{eq_unit_coverage} degrades as ${\Pr}_G[d_t = 0]$ decreases.

\section{Capacity-Aware OCRC}\label{Sec:4}
In this section, we introduce capacity-aware online conformal risk control (CA-OCRC), a policy that can satisfy the long-term risk requirement \eqref{eq_miscov} for arbitrary values of the capacity $K$.
CA-OCRC is summarized in Algorithm~\ref{alg:ca-ocp} and detailed next.

\suppressfloats[t]
\begin{algorithm}[t]
    \DontPrintSemicolon
    \SetKwInOut{Input}{Input}
    \Input{Target risk level $\alpha$, learning rate $\eta$, capacity $K$, and proxy delay distribution $F$.}
    Initialize $r_1=0$, $\bar{\bm d}=\mathbf{0}_T$, and $\mathcal{K}_0=\emptyset$\;
    \For{$t=1,2,\dots,T$}{
        Observe $X_t$ and return $\mathcal{C}_t(X_t;r_t)$ via \eqref{eq_pred_set}\;
        Set $\mathcal{E}_t=\{k\in\mathcal{K}_{t-1}:k+\bar d_k<t\}$, where $\bar d_k=\bar{\bm d}[k]$\;
        Set $\mathcal{K}_t^-=\mathcal{K}_{t-1}\setminus\mathcal{E}_t$\;
        \uIf{$\mathcal{C}_t\in\{\emptyset,\mathcal{Y}\}$}{
            Calculate $L_t=\mathbbm{1}\{\mathcal{C}_t=\emptyset\}$\;
			Set $\bar{d}_t = - 1$ and $\mathcal{A}_t = \emptyset$
        }
        \Else{
            Sample $\bar d_t\sim F$ independently and store $\bar{\bm d}[t]=\bar d_t$\;
			Update $\mathcal{A}_t$ via \eqref{eq_ca_ocp_admit}
        }
		\lIf*{$|\mathcal{K}_t^-|<K$}{$\mathcal{K}_t^+ = \mathcal{K}_t^- \cup \mathcal{A}_t$ }\lElse{$\mathcal{K}_t^+ = \mathcal{K}_t^-$}
        Observe feedback $\{L_{t'}\}_{t' \in \mathcal{O}_t}$\;
        Set $\mathcal{K}_t=\mathcal{K}_t^+\setminus\mathcal{O}_t$\;
        Update $r_{t+1}$ via \eqref{eq_ca_ocp}\;
    }
    \caption{Capacity-Aware OCRC (CA-OCRC)}
    \label{alg:ca-ocp}
\end{algorithm}

\subsection{Tracking Policy}\label{sec:policy}

CA-OCRC implements randomized eviction and admission policies based on proxy delays generated for each round $t$.
Inspired by \cite{ryabchenko2026capacity}, the general principle is that, if capacity permits, each round $t$ is tracked until the end of round $t+\bar{d}_t$, where $\bar{d}_t$ is the random proxy delay for round $t$.
The proxy delays $\bar{d}_t$ are drawn independently for each round $t$, following a survival function $F(d) = \Pr_F [\bar{d}_t \ge d] $ to be designed.
The optimization of the function $F(d)$ is covered in Sec. \ref{sec:5} and Sec. \ref{sec:6} under Assumption~\ref{assum:delay}(a) and Assumption~\ref{assum:delay}(b), respectively.

At each round $t$, CA-OCRC first checks whether the prediction set $\mathcal{C}_t$ is trivially empty or full, i.e., if $\mathcal{C}_t \in \{\emptyset, \mathcal{Y}\}$.
In either case, there is no need to track feedback $L_t$, since, by Assumption~\ref{assum:loss}, the loss $L_t$ is known to equal $1$ for $\mathcal{C}_t = \emptyset$ and $0$ for $\mathcal{C}_t = \mathcal{Y}$.
In such cases, a proxy delay $\bar{d}_t = -1$ is selected, indicating that round $t$ is not tracked:
\begin{equation}
	\bar{d}_t = -1 \text{ if } \mathcal{C}_t \in \{\emptyset, \mathcal{Y}\} .
\end{equation}

For a nontrivial prediction set $\mathcal{C}_t\notin\{\emptyset,\mathcal{Y}\}$, CA-OCRC instead samples the random proxy delay $\bar{d}_t$, following the proxy delay distribution $F$, which is supported on the set $\{-1,0,1,\ldots,\bar{d}_{\max}\}$, where $\bar{d}_{\max}\le d_{\max}$ is the maximum proxy delay:
\begin{equation}
	\bar{d}_t \sim F \text{ if } \mathcal{C}_t \notin \{\emptyset, \mathcal{Y}\} .
\end{equation}

Based on the proxy delays, CA-OCRC evicts the tracked rounds whose proxy delays have expired, i.e.,
\begin{equation}\label{eq_ca_ocp_evict}
	\mathcal{E}_t=\{i\in\mathcal{K}_{t-1}: i+\bar{d}_i<t\} ,
\end{equation}
and selects round $t$ for admission only if its proxy delay is nonnegative, i.e., 
\begin{equation}\label{eq_ca_ocp_admit}
	\mathcal{A}_t = \left\{
		\begin{array}{ll}
			\{t\}, & \text{if } \bar{d}_t \ge 0,\\
			\emptyset, & \text{otherwise} .
	\end{array}\right.
\end{equation}

\subsection{Threshold Update}\label{sec:update}
CA-OCRC produces the prediction set \eqref{eq_pred_set} with the threshold updated as
\begin{equation}\label{eq_ca_ocp}
	r_{t+1}
	= r_t + \eta \varphi_t (L_t-\alpha)
	+ \eta \sum_{t'\in\mathcal{O}_t}\frac{L_{t'}-\alpha}{q_G(\bar{d}_{\max}) \,F(d_{t'})} ,
\end{equation}
where $\varphi_t = \mathbbm{1} \{\mathcal{C}_t \in \{\emptyset,\mathcal Y\}\}$ equals 1 when $\mathcal{C}_t$ is a trivial prediction set, and 0 otherwise; and
\begin{equation}\label{eq_qF}
	q_G(\bar{d}_{\max}) = {\Pr}_{G} [d_t\le\bar{d}_{\max}] = 1 - G(\bar{d}_{\max}+1)
\end{equation}
is the probability that the true feedback delay $d_t$ does not exceed the maximum proxy delay $\bar{d}_{\max}$.
Under Assumption \ref{assum:delay}(a), i.e., deterministic delays, we set $\bar{d}_{\max} = d_{\max}$ and $q_G(\bar{d}_{\max}) = 1$, and thus CA-OCRC is applicable under both Assumption \ref{assum:delay}(a) and Assumption \ref{assum:delay}(b).

The OCRC update $(L_t - \alpha)$ in \eqref{eq:unconstrained-update} is applied only to rounds with trivial prediction sets $\mathcal{C}_t \in \{\emptyset, \mathcal{Y}\}$.
In contrast, for each past received feedback signal $L_{t'}$ with $t' < t$, the update $(L_{t'} - \alpha)$ is applied at each time $t$ and is scaled by the factor $(q_G(\bar{d}_{\max}) F(d_{t'}))^{-1}$, thus applying a two-level form of inverse probability weighting \cite{elvira2021advances}.

This scaling resembles that applied by the I-OCRC update \eqref{eq_i_ocp}.
In particular, the factor $q_G(\bar{d}_{\max})^{-1}$ compensates for the randomness of the feedback delays as in I-OCRC, since the feedback signals with delays larger than $\bar{d}_{\max}$ are never received.
Moreover, the first-level factor $(F(d_{t'}))^{-1}$ compensates for the randomness of the proxy delays, which determine the eviction and admission policies \eqref{eq_ca_ocp_evict} and \eqref{eq_ca_ocp_admit}.

\section{Theoretical Results under Arbitrary Delays}\label{sec:5}

In this section, we analyze the long-term risk of CA-OCRC under Assumption~\ref{assum:delay}(a), i.e., for arbitrary delays with $\bar{d}_{\max}=d_{\max}$ and $q_G(\bar{d}_{\max})=1$ (see Sec.~\ref{sec:update}).
Based on the analysis, we also propose a design method for the proxy delay distribution $F$.

\subsection{Risk Guarantee}
Since CA-OCRC evicts the rounds whose proxy delays have expired, a round $i$ can be tracked at a later round $t>i$ only if its proxy delay satisfies $\bar{d}_i\ge t-i$.
In the absence of capacity constraints, the expected number of previous rounds tracked by CA-OCRC at any round is thus upper bounded by the \emph{tracking load}
\begin{equation}\label{eq:wc-load}
	\ell_F=\sum_{a=1}^{d_{\max}}F(a)=\mathbb{E}_F \left[(\bar{d})_+\right],
\end{equation}
where $(x)_+=\max\{x,0\}$.
To control the probability of capacity saturation, i.e., $|\mathcal{K}_t^-|=K$, we impose the design constraint that the tracking load \eqref{eq:wc-load} not exceed the capacity $K$.
This design choice is formalized by the following assumption.

\begin{assumption}\label{assum:wc-load}
	The proxy delay distribution $F$ satisfies the underload condition $\ell_F\le K$.
\end{assumption}

The following theorem shows that, under Assumption~\ref{assum:wc-load}, CA-OCRC satisfies the long-term risk condition \eqref{eq_miscov} up to an additive term that accounts for capacity saturation.

\begin{theorem}[Risk bound under arbitrary delays]\label{thm:worstcase_miscoverage}
	Under Assumption~\ref{assum:delay}(a), Assumption \ref{assum:score}, Assumption~\ref{assum:loss}, and Assumption \ref{assum:wc-load} with $r_1\in[0,B]$, the risk gap \eqref{eq_miscov} of CA-OCRC satisfies the upper bound
	\begin{equation}\label{eq:wc-bound}
		\mathrm{RiskGap}(T)
		\le \frac{B}{\eta T}
		+\frac{K+1}{F(d_{\max}) T}
		+e^{-Kh\left(\ell_F/K\right)}
	\end{equation}
	for any sequence of data points, where 
	\begin{equation}\label{eq_hx}
		h(x) = x-1-\ln x .
	\end{equation}
\end{theorem}
\begin{proof}
	See Supplementary Material Appendix~\ref{app:thm_worstcase_miscoverage}.
\end{proof}

The first two terms in the bound \eqref{eq:wc-bound} are similar to the bound \eqref{eq_unit_coverage} for I-OCRC.
In fact, the probability $F(d_{\max})$ provides a lower bound on the probability of receiving feedback in any given round. 
In contrast, the last term in the bound \eqref{eq:wc-bound} accounts for the probability of capacity saturation, which is absent in I-OCRC since removing any previously tracked round allows the current round to be tracked.
By Lemma~\ref{lem:arb-implies-iid}, the bound \eqref{eq:wc-bound} also holds under the stochastic delays in Assumption~\ref{assum:delay}(b).

\subsection{Optimization of the Proxy Delay Distribution}
We propose to optimize the proxy delay distribution $F$ by minimizing the upper bound in Theorem \ref{thm:worstcase_miscoverage}. 
This yields the problem
\begin{subequations}\label{eq:wc-opt}
	\begin{align}
		\min_{\{F(a)\}_{a=-1}^{d_{\max}}} & \frac{K+1}{ F(d_{\max}) T } + e^{-K h \left( \ell_F / K\right)} \\
		\text{s.t.} \quad & \ell_F = \sum_{a=1}^{d_{\max}} F(a) \le K , \\
		& 1 = F(-1) \ge F(0) \ge \ldots \ge F(d_{\max}) \ge 0 .
	\end{align}
\end{subequations}

The following proposition describes a solution to problem~\eqref{eq:wc-opt}.

\begin{proposition}\label{prop:wc-opt}
	An optimal solution to problem~\eqref{eq:wc-opt} is given by the proxy delay distribution
	\begin{equation}\label{eq:wc_opt_F}
		F^\star(a) = p^\star, \text{ for } a \in \{1,\ldots,d_{\max}\} ,
	\end{equation}
	with any $F^\star(0) \in [p^\star, 1]$ and $F^\star(-1)=1$, where $p^\star$ is obtained by solving the one-dimensional problem
	\begin{equation}\label{eq:wc-opt-p}
		p^\star = 
		\mathop{\arg\min}\limits_{0<p\le K/d_{\max}} \frac{K + 1}{p T} + e^{ -K h \left({d_{\max}p} / {K}\right) } .
	\end{equation}
\end{proposition}
\begin{proof}
	See Supplementary Material Appendix~\ref{app:wc-design}.
\end{proof}

Under the optimized proxy delay distribution in Proposition~\ref{prop:wc-opt}, CA-OCRC tracks each round until the maximum possible delay, i.e., $\bar{d}_t = d_{\max}$, with probability $p^\star$, while not tracking the round otherwise, i.e., $\bar{d}_t = -1$.

\subsection{Scaling Law}
The next result, which follows from Theorem~\ref{thm:worstcase_miscoverage} and Proposition~\ref{prop:wc-opt}, shows that, under the optimized proxy delay distribution in Proposition~\ref{prop:wc-opt}, the risk gap of CA-OCRC vanishes as $T\to\infty$ for any fixed capacity $K$, whenever the maximum delay scales sublinearly with $T$, i.e., $d_{\max}=o(T)$.

\begin{corollary}\label{cor:wc-rate}
	Under Assumption~\ref{assum:delay}(a), Assumption \ref{assum:score}, and Assumption~\ref{assum:loss} with $r_1\in[0,B]$ and $1\le K\le d_{\max}\le T$, the risk gap of CA-OCRC with the proxy delay distribution in Proposition~\ref{prop:wc-opt} satisfies the upper bound
	\begin{equation}\label{eq:wc-rate}
		\mathrm{RiskGap}(T) \le
		\frac{B}{\eta T} 
		+ (2e+1)\left(\frac{d_{\max}}{T}\right)^{\frac{K}{K+1}} .
	\end{equation}
\end{corollary}
\begin{proof}
	See Supplementary Material Appendix~\ref{app:wc-design}.
\end{proof}

Based on the bound in \eqref{eq:wc-rate}, the main effect of the capacity constraint is to reduce the exponent of the convergence rate from $1$ to $K/(K+1)$.
In particular, even with unit-capacity, i.e., $K=1$, CA-OCRC attains the rate $\mathcal{O}(\sqrt{d_{\max}/T})$ under arbitrary delays, while I-OCRC provides no guarantees under Assumption~\ref{assum:delay}(a) (see Lemma~\ref{lemm_i_ocp}).
Furthermore, a capacity $K$ of the order of $\ln(T/d_{\max})$ suffices for CA-OCRC to attain the same rate $\mathcal{O}(d_{\max}/T)$ as OCRC as per the upper bound in \eqref{eq:unconstrained-coverage}.
This capacity may be much smaller than the maximum delay $d_{\max}$, while OCRC requires a capacity $K\ge d_{\max}+1$ to accommodate all pending feedback.

\section{Theoretical Results under Stochastic Delays}\label{sec:6}

In this section, we analyze the long-term risk of CA-OCRC under Assumption~\ref{assum:delay}(b), i.e., for stochastic delays.
By Lemma~\ref{lem:arb-implies-iid}, the guarantee in Theorem~\ref{thm:worstcase_miscoverage} also applies to this setting.
Here, we show that knowledge of the delay distribution $G$ yields a refined guarantee, and that it can be leveraged to tailor the proxy delay distribution $F$ to the feedback delays.

\subsection{Risk Guarantee}
Recall that a tracked round leaves the tracking set as soon as its feedback is received.
Accordingly, a round $i$ can be tracked at a later round $t>i$ only if both its proxy delay and its feedback delay are no smaller than $t-i$, i.e., $\bar{d}_i\ge t-i$ and $d_i\ge t-i$.
Since the proxy delays are drawn independently of the feedback delays, in the absence of capacity constraints, the expected number of previous rounds tracked by CA-OCRC at any round is thus upper bounded by the \emph{stochastic tracking load}
\begin{equation}\label{eq:stoch-load}
	\ell_{F,G} = \sum_{a=1}^{d_{\max}}F(a)\,G(a)=\mathbb{E}_{F,G} \left[\min\{(\bar{d}_t)_+,d_t\}\right] .
\end{equation}
The stochastic tracking load \eqref{eq:stoch-load} is no larger than the tracking load $\ell_F$ in \eqref{eq:wc-load}, with equality holding when all delays equal $d_{\max}$, i.e., when $G(a)=1$ for $a\in\{1,\ldots,d_{\max}\}$.
As in Sec.~\ref{sec:5}, we impose the design constraint that the stochastic tracking load not exceed the capacity $K$, as formalized by the following assumption.

\begin{assumption}\label{assum:stoch-load}
	The proxy delay distribution $F$ satisfies the stochastic underload condition $\ell_{F,G}\le K$.
\end{assumption}

Assumption~\ref{assum:stoch-load} is always satisfied if we have $\bar{d}_{\max}<K$, since in this case, at most $\bar{d}_{\max}<K$ previous rounds can be tracked, and thus capacity saturation never occurs.

The following theorem provides the risk bound under stochastic delays.

\begin{theorem}[Risk bound under stochastic delays]\label{thm_iid_miscoverage}
	Under Assumption~\ref{assum:delay}(b), Assumption~\ref{assum:score}, Assumption~\ref{assum:loss}, and Assumption~\ref{assum:stoch-load} with $r_1\in[0,B]$, the risk gap \eqref{eq_miscov} of CA-OCRC satisfies the upper bound
	\begin{equation}\label{eq:iid-bound}
		\mathrm{RiskGap}(T)
		\le \frac{B}{\eta T}
		+\frac{R_{F,G}}{T}
		+\mathbbm{1}\{\bar{d}_{\max}\ge K\}\,e^{-Kh\left(\ell_{F,G}/K\right)}
	\end{equation}
	for any sequence of data points, where $h(\cdot)$ is defined in \eqref{eq_hx},
	\begin{equation}\label{eq:iid-R}
		R_{F,G}=\bar{\alpha}\left(  \sum_{a = 0}^{\bar{d}_{\max}} a \frac{G(a) - G(a+1)}{q_G(\bar{d}_{\max})} +2\right)+\frac{K\bar{\alpha}^2}{2(1-\bar{\alpha})}\,Q_{F,G}
	\end{equation}
	with $\bar{\alpha}=\max\{\alpha,1-\alpha\}$,
	and
	\begin{equation}\label{eq:iid-Q}
		Q_{F,G} =\sum_{a=0}^{\bar{d}_{\max}} \frac{ G(a) - G(a+1) }{q_G(\bar{d}_{\max})^2\,F(a)} .
	\end{equation}
\end{theorem}
\begin{proof}
	See Supplementary Material Appendix~\ref{app:thm_iid_miscoverage}.
\end{proof}

Compared with the bound \eqref{eq:wc-bound} for deterministic delays in Theorem~\ref{thm:worstcase_miscoverage}, the bound \eqref{eq:iid-bound} contains similar term $\mathbbm{1}\{\bar{d}_{\max}\ge K\} e^{-Kh(\ell_{F,G}/K)}$, which bounds the probability of capacity saturation, being equal to zero if $\bar{d}_{\max}<K$.
It also includes the additional term $R_{F,G}/T$, which bounds the range of the threshold iterates.

\subsection{Optimization of the Proxy Delay Distribution}
As in Sec.~\ref{sec:5}, we propose to optimize the proxy delay distribution $F$ by minimizing the upper bound in Theorem~\ref{thm_iid_miscoverage}.
To this end, we fix the maximum proxy delay $\bar{d}_{\max}$ to a possible value of the feedback delays, i.e., with ${\Pr}_G [d_t=\bar{d}_{\max}]>0$, so that $q_G(\bar{d}_{\max})>0$, and we optimize $\bar{d}_{\max}$ via a grid search over all such values.
Removing the terms that do not depend on $F$, we obtain the problem
\begin{subequations}\label{eq:iid-opt}
	\begin{align}
		\min_{\{F(a)\}_{a=-1}^{\bar{d}_{\max}}} & \frac{R_{F,G}}{T}+\mathbbm{1}\{\bar{d}_{\max}\ge K\}\,e^{-Kh\left(\ell_{F,G}/K\right)} \label{eq:iid-opt-obj} \\
		\text{s.t.} \quad\;\; & \ell_{F,G}=\sum_{a=1}^{\bar{d}_{\max}}F(a)\,G(a)\le K , \label{eq:iid-opt-cap} \\
		& 1=F(-1)\ge F(0)\ge\ldots\ge F(\bar{d}_{\max})>0 . \label{eq:iid-opt-mono}
	\end{align}
\end{subequations}

Unlike problem~\eqref{eq:wc-opt}, whose objective depends on $F$ only through the value $F(d_{\max})$ and the tracking load $\ell_{F}$, the objective of problem~\eqref{eq:iid-opt} depends on the entire proxy delay distribution through the quantity $Q_{F,G}$ in \eqref{eq:iid-Q}.
The following proposition describes a solution to problem~\eqref{eq:iid-opt}.

\begin{proposition}\label{prop:iid-opt}
	An optimal solution to problem~\eqref{eq:iid-opt} is given by the proxy delay distribution with $F^\star(a)=1$ for $a\in\{-1,0,\ldots,\bar{d}_{\max}\}$ if $\bar{d}_{\max}<K$; and, if $\bar{d}_{\max}\ge K$, by the proxy delay distribution
	\begin{equation}\label{eq:iid_opt_F}
		F^\star(a)=\left\{
		\begin{array}{ll}
			1, & \text{for } a\in\{-1,0\},\\
			\min\left\{1,\theta^\star\sqrt{\kappa(a)}\right\}, & \text{for } a\in\{1,\ldots,\bar{d}_{\max}\},
		\end{array}\right. 
	\end{equation}
	where $\theta^\star>0$ is obtained by solving problem~\eqref{eq:iid-opt} over the one-parameter family of distributions of the form \eqref{eq:iid_opt_F}, and
	\begin{equation}\label{eq:hazard-iso}
		\kappa(a)=\min_{1\le s\le a} \; \max_{a\le s'\le\bar{d}_{\max}} \frac{ G(s) - G(s'+1) }{\sum_{i=s}^{s'}G(i)} .
	\end{equation}
\end{proposition}
\begin{proof}
	See Supplementary Material Appendix~\ref{app:iid-design}.
\end{proof}

If $K > \bar{d}_{\max}$, the proxy delay can be set to $\bar{d}_t=\bar{d}_{\max}$ for every round $t$ with a nontrivial prediction set, since  capacity saturation can never occur in this case.
In this setting, for $\bar{d}_{\max}=0$, we have $q_G(\bar{d}_{\max})=G(0) - G(1)$, and the CA-OCRC update \eqref{eq_ca_ocp} recovers I-OCRC in \eqref{eq_i_ocp}.
Accordingly, CA-OCRC can attain the same performance as I-OCRC, with the bound \eqref{eq:iid-bound} matching the bound \eqref{eq_unit_coverage} up to constant factors.

If $\bar{d}_{\max}\ge K$, instead, CA-OCRC with the distribution \eqref{eq:iid_opt_F} tracks a pending round at age $a$, i.e., $a$ rounds after its prediction, with a probability $F^\star(a)$ that increases with $\kappa(a)$.
Intuitively, the ratio $\kappa(a)$ compares the probability $G(s)-G(s'+1)$ that the feedback of a round arrives between ages $s$ and $s'$ with the tracking load $\sum_{i=s}^{s'}G(i)$ that the round contributes over these ages.
CA-OCRC thus keeps tracking a round while this ratio is high, and evicts it otherwise to free capacity for new rounds.
The parameter $\theta^\star$ sets the overall tracking level: a larger $\theta^\star$ yields more feedback, reducing the quantity $Q_{F,G}$, at the cost of a larger tracking load $\ell_{F,G}$.

\subsection{Scaling Law}
The next result, which follows from Theorem~\ref{thm_iid_miscoverage} and Proposition~\ref{prop:iid-opt}, shows that, under stochastic delays, the risk gap of CA-OCRC with the optimized proxy delay distribution is governed by the mean delay $\mathbb{E}_G [d_t]$, rather than by the maximum delay $d_{\max}$.

\begin{corollary}\label{cor:iid-rate}
	Under Assumption~\ref{assum:delay}(b), Assumption~\ref{assum:score}, and Assumption~\ref{assum:loss} with $r_1\in[0,B]$, the risk gap of CA-OCRC with the proxy delay distribution in Proposition~\ref{prop:iid-opt} and the optimized maximum proxy delay satisfies the upper bound
	\begin{equation}\label{eq:iid-rate}
		\mathrm{RiskGap}(T) \le \frac{B}{\eta T} + \frac{2\bar{\alpha}}{T} 
		+ C_{\alpha} \left(\frac{\mathbb{E}_G[d_t]}{T}\right)^{\frac{K}{K+1}} .
	\end{equation}
	where $\bar{\alpha} = \max \{ \alpha , 1 - \alpha\}$ and $C_{\alpha} = 1 + \bar{\alpha} + e\bar{\alpha}^2/(2 - 2\bar{\alpha})$.
\end{corollary}
\begin{proof}
	See Supplementary Material Appendix~\ref{app:iid-design}.
\end{proof}

By replacing the maximum delay $d_{\max}$ in Corollary~\ref{cor:wc-rate} with the mean delay $\mathbb{E}_G[d_t]$, Corollary~\ref{cor:iid-rate} suggests that random delays can yield significantly lower risk gaps as compared to deterministic delays, since the inequality $\mathbb{E}_G[d_t]\le d_{\max}$ holds.

\section{Numerical Results}\label{sec:7}

In this section, we evaluate CA-OCRC on tumor segmentation \cite{angelopoulos2023conformalge}, open-domain question answering \cite{angelopoulos2024conformal}, hierarchical image classification \cite{deng2012hedging}, and wireless beam selection \cite{zecchin2024localized} tasks.

\subsection{Setting}

\subsubsection{Datasets}\label{sec_7_A_1}
We consider four different experimental settings.

\noindent\textbf{Tumor Segmentation:}
We adopt the tumor segmentation setting of \cite{angelopoulos2023conformalge}, in which the input $X_t$ is a medical image, and the label $Y_t$ is the nonempty subset of pixels of the image that belong to a tumor.
As in \cite{angelopoulos2023conformalge}, the dataset is a compilation of samples from several open-source online repositories, namely Kvasir, CVC-300, CVC-ColonDB, CVC-ClinicDB, and ETIS-LaribDB.
The pre-trained model $f(\cdot)$ is the PraNet segmentation model \cite{fan2020pranet}, which outputs the probability $f_y(X_t)\in[0,1]$ that each pixel $y\in\mathcal{Y}$ belongs to a tumor.
Using the score $S_t(X_t,y)=1-f_y(X_t)$, which satisfies Assumption~\ref{assum:score} with $B=1$, the prediction set \eqref{eq_pred_set} is the segmentation mask $\mathcal{C}_t(X_t;r_t)=\{y\in\mathcal{Y}:f_y(X_t)\ge 1-r_t\}$.
Performance is measured by the false negative rate (FNR) loss $\mathcal{L}(\mathcal{C}_t,Y_t)=1 - {|\mathcal{C}_t\cap Y_t|} / {|Y_t|}$, which satisfies Assumption~\ref{assum:loss} since the label $Y_t$ is nonempty.

\noindent\textbf{Open-Domain Question Answering:}
We then consider the open-domain question answering setting of \cite{angelopoulos2024conformal}, in which the input $X_t \in \mathcal{X}$ is a natural-language question, and the label $Y_t \subset \mathcal{Y}$ is the set of ground-truth answers to the question.
The input domain $\mathcal{X}$ is the set of all token sequences, while the output domain $\mathcal{Y}$ is the set of all token sequences of length at most $m$.
As in \cite{angelopoulos2024conformal}, we use the $3610$ test questions of the Natural Questions (NQ) dataset \cite{kwiatkowski2019natural}, according to the split released with the dense passage retrieval (DPR) package \cite{karpukhin2020dense}.
The pre-trained model $f(\cdot)$ is the DPR retriever-reader pipeline, which returns a confidence score $f_y(X_t) \in [-M/2 , M/2]$, with $M=80$, for each answer $y$ in a set $\hat{\mathcal{Y}}_t\subset\mathcal{Y}$ of candidate answers.
Using the score $S_t(X_t,y) = 0.5 - f_y(X_t)/M$ for $y \in \hat{\mathcal{Y}}_t$ and $S_t(X_t,y)=1$ otherwise, which satisfies Assumption~\ref{assum:score} with $B=1$, the prediction set \eqref{eq_pred_set} is given by $\mathcal{C}_t(X_t;r_t) = \{y \in \hat{\mathcal{Y}}_t : f_y(X_t) \ge (0.5 - r_t) F \}$ for $r_t<1$, and by $\mathcal{C}_t(X_t;r_t) = \mathcal{Y}$ for $r_t\ge 1$.
Performance is measured by the smallest F1 loss for elements of the set $\mathcal{C}_t$, i.e., $\mathcal{L}(\mathcal{C}_t,Y_t)=1-\max_{y\in\mathcal{C}_t,\,y'\in Y_t}\mathrm{F1}(y,y')$, where $\mathrm{F1}(y,y')\in[0,1]$ is the token-level F1 score between the answers $y$ and $y'$ \cite{rajpurkar2016squad}.
Note that we have $\mathcal{L}(\emptyset,Y_t)=1$ and, since $Y_t$ is a nonempty subset of $\mathcal{Y}$, $\mathcal{L}(\mathcal{Y},Y_t)=0$, which satisfies Assumption~\ref{assum:loss}.

\noindent\textbf{Hierarchical Image Classification:}
We also adopt the hierarchical image classification setting of \cite{deng2012hedging}, in which the input $X_t$ is an image from the ImageNet validation set \cite{deng2009imagenet}, the output domain $\mathcal{Y}$ is the set of $1000$ classes, and the label $Y_t\in\mathcal{Y}$ is the ground-truth class.
The classes are the leaves of a tree of height $H=14$ in the WordNet hierarchy \cite{fellbaum1998wordnet}.
The prediction set contains all the classes below a node of the tree, so that nodes closer to the root yield larger prediction sets.
The pre-trained model $f(\cdot)$ is a ResNet-152 \cite{he2016deep}, which outputs the probability $f_y(X_t)\in[0,1]$ of each class $y\in\mathcal{Y}$.
We denote by $\hat{y}_t$ the top-1 class, and by $f_v(X_t)$ the total probability of the classes below a node $v$.
We use the score $S_t(X_t,y)=f_{v_t(y)}(X_t)$ for $y\neq\hat{y}_t$, where $v_t(y)$ is the lowest common ancestor of the classes $y$ and $\hat{y}_t$, and $S_t(X_t,\hat{y}_t)=0$, which satisfies Assumption~\ref{assum:score} with $B=1$.
Accordingly, for $r_t\ge 0$, the prediction set \eqref{eq_pred_set} contains all the classes below the node $v_t$, with $v_t=\hat{y}_t$ if no such node exists.
Performance is measured by the hierarchical loss $\mathcal{L}(\mathcal{C}_t,Y_t)=\mathrm{dist}(v_t,Y_t)/H$, where $\mathrm{dist}(v_t,Y_t)$ is the number of levels from the node $v_t$ up to its lowest common ancestor with the class $Y_t$.
Note that we set $\mathcal{L}(\emptyset,Y_t)=1$ and, since the root is an ancestor of every class, we have $\mathcal{L}(\mathcal{Y},Y_t)=0$, which satisfies Assumption~\ref{assum:loss}.

\noindent\textbf{Beam Selection:}
We finally consider the wireless beam selection setting of \cite{zecchin2024localized}, in which a base station with $8$ antennas selects a set of candidate beams from a discrete Fourier transform codebook \cite{zecchin2024localized} of $11$ beams, so that the output domain is $\mathcal{Y}=\{1,\ldots,11\}$.
The input $X_t$ is the position of the user, and the pre-trained model $f(\cdot)$ is a neural network that outputs a probability $f_b (X_t) \in[0,1]$ for each beam $b \in \mathcal{Y}$.
With the score $S_t(X_t,b) = 1-f_b(X_t)$, which satisfies Assumption~\ref{assum:score} with $B=1$, the prediction set \eqref{eq_pred_set} is given by $\mathcal{C}_t(X_t;r_t) = \{b \in \mathcal{Y} : f_b(X_t) \ge 1-r_t\}$.
Performance is measured by the SNR loss $\mathcal{L}(\mathcal{C}_t,Y_t)=1-\max_{b\in\mathcal{C}_t} \mathrm{SNR}_{t,b}/\max_{b'\in\mathcal{Y}}\mathrm{SNR}_{t,b'}$, where $\mathrm{SNR}_{t , b}$ is the signal-to-noise ratio of beam $b$ at round $t$.
Note that we have $\mathcal{L}(\emptyset,Y_t)=1$ and $\mathcal{L}(\mathcal{Y},Y_t)=0$, which satisfies Assumption~\ref{assum:loss}.

\subsubsection{Delay Model}\label{sec:7_delay}
We adopt an autoregressive gamma model for the delays.
Gamma distributions including the exponential and Erlang distributions provide a standard model for waiting times in queueing systems \cite{yao2019tsbatpro,chen2016dynamic,kleinrock1975queueing}, packet delays in communication networks \cite{kalman2004modeling,severinson2022dsag,liang2006network}, and conversion delays in online advertising \cite{chapelle2014modeling,ktena2019addressing}.
Specifically, as shown in Fig. \ref{Fig_0}(b), we assume that round $t$ spans the continuous-time interval $[t,t+1)$, and the feedback $L_t$ arrives at time $t+D_t$, where $D_t \ge 0$ is the underlying continuous-time gamma-distributed delay.
Therefore, any feedback $L_t$ arriving during round $t+\lfloor D_t\rfloor$ can only be used at the end of round $t + \lfloor D_t\rfloor$.

Formally, we adopt a beta-gamma first-order autoregressive (AR(1)) process \cite{lewis1989gamma}, obtaining the discrete-time delays $d_t$ by quantizing the continuous-time delays $D_t$ and clipping them at $d_{\max}$, yielding
\begin{equation}\label{eq:gamma-ar1}
	\left\{
	\begin{array}{l}
		D_{t+1} = U_{t+1} D_t + \xi_{t+1} , \\
		d_t = \min \{ \lfloor D_t \rfloor , d_{\max} \} ,
	\end{array}
	\right.
\end{equation}
where $D_1\sim\mathrm{Gamma}(\nu,\theta)$ follows the gamma distribution with shape $\nu>0$ and scale $\theta>0$, and the random variables $U_t \sim \mathrm{Beta} (\nu \rho  , \nu(1-\rho ))$ and $\xi_t \sim \mathrm{Gamma} (\nu(1-\rho ) , \theta)$, for $t \ge 2$, are mutually independent and independent of $D_1$.
The process \eqref{eq:gamma-ar1} preserves the gamma marginal distribution, i.e., $D_t\sim\mathrm{Gamma}(\nu,\theta)$ for all rounds $t$, while $\rho \in[0,1)$ is the correlation coefficient between consecutive continuous-time delays \cite{lewis1989gamma}.

Accordingly, the discrete-time delays $d_t$ have the survival function
\begin{equation}\label{eq:gamma-G}
	G(a) = \Pr[d_t\geq a] = 1-\Psi(a),
\end{equation}
for $a \in \{0,1,\ldots,d_{\max}\}$ and all rounds $t$, where $\Psi(\cdot)$ is the cumulative distribution function of $\mathrm{Gamma}(\nu,\theta)$.
The shape parameter $\nu$ controls the skewness of the gamma distribution, with $\nu = 1$ corresponding to the exponential distribution and integer values of $\nu$ yielding the Erlang distribution, while the scale parameter $\theta$ stretches the distribution without changing its shape \cite{lewis1989gamma}. 
The mean and variance of the delays are $\nu\theta$ and $\nu\theta^2$, respectively.
For $\rho =0$, we set $U_t=0$, and thus the delays are independent and identically distributed (i.i.d.) as in Assumption~\ref{assum:delay}(b), while for $\rho >0$, they are identically distributed but temporally correlated, and realization from process \eqref{eq:gamma-ar1} hence satisfy only Assumption~\ref{assum:delay}(a).

\subsubsection{Benchmarks}
We compare CA-OCRC against the following schemes:
\begin{itemize}
	\item The original OCRC \cite{feldman2023achieving,zecchin2024localized} without the capacity constraint, i.e., $K = \infty$ (see Sec.~\ref{sec:ocrc}), which tracks all rounds, and thus serves as an ideal benchmark;
	\item FIFO-OCRC, which tracks each round $t$ until the end of round $t+K-1$, thus managing the tracking set as a first-in first-out (FIFO) queue of the $K$ most recent rounds. 
	Under i.i.d. delays, FIFO-OCRC updates the threshold by scaling received feedback signals with factor $({\Pr}_G[d_t\le K-1])^{-1}$, thus extending I-OCRC (see Sec.~\ref{sec:iocrc}) from unit capacity to an arbitrary capacity $K$. Under temporally correlated delays, it instead applies the OCRC update \eqref{eq:unconstrained-update} to the received feedback signals.
\end{itemize}

The evaluated variants of CA-OCRC, which differ in the proxy delay distribution $F$, are described in Sec.~\ref{sec:exp_tumor} for i.i.d. delays and in Sec.~\ref{sec:exp_Image} for temporally correlated delays.

\begin{figure*}[t]
	\centering
	\subfigure[]{\label{Exp1_0a}
		\includegraphics[width=0.32\textwidth]{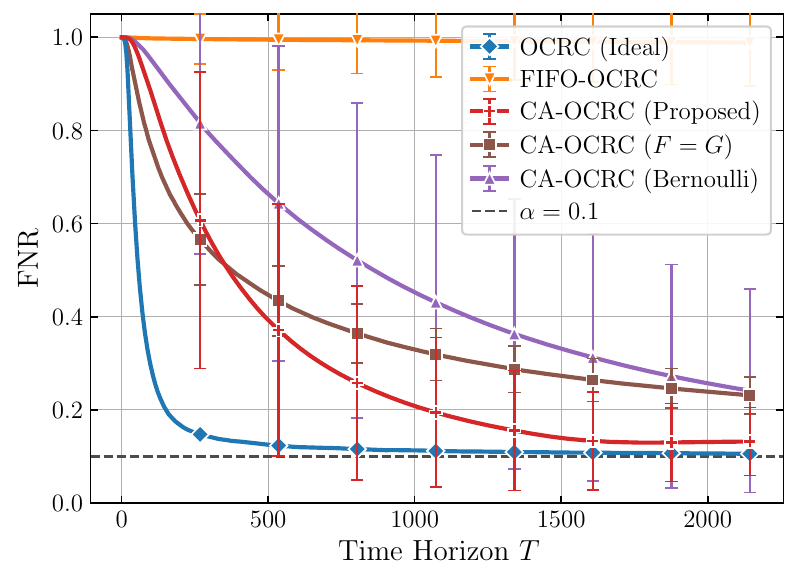}
	}
	\hspace{-1em}
	\subfigure[]{\label{Exp1_0b}
		\includegraphics[width=0.32\textwidth]{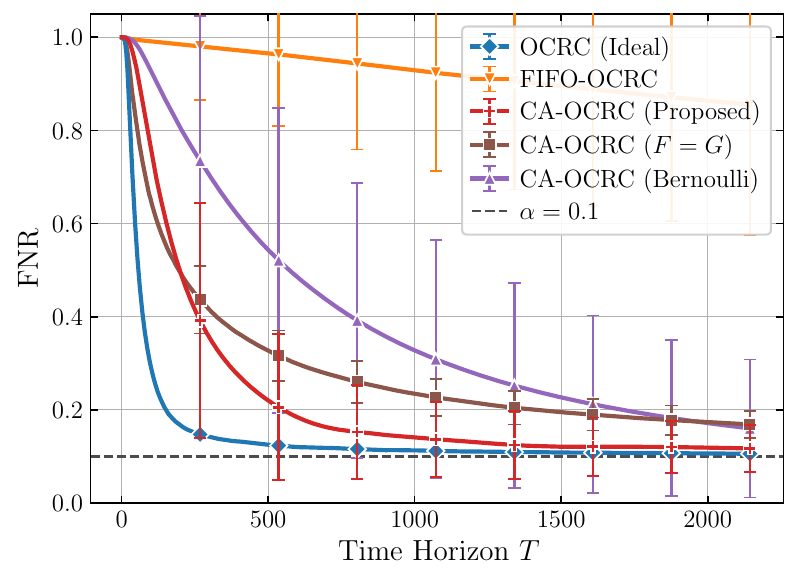}
	}
	\hspace{-1em}
	\subfigure[]{\label{Exp1_0c}
		\includegraphics[width=0.32\textwidth]{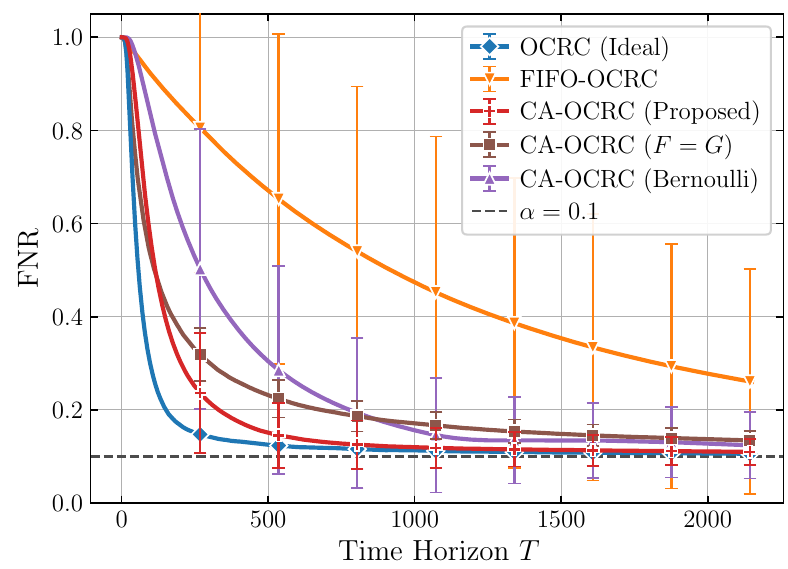}
	}\vspace{-0.5em}
	\caption{FNR as a function of the time horizon $T$ for the tumor segmentation task with $\nu=4$, $\theta=8$, $\rho =0$, $d_{\max}=150$, and target level $\alpha=0.1$ (dashed line), for capacities (a) $K=1$, (b) $K=2$, and (c) $K=4$. The results are averaged over $10^3$ independent trials, with the error bars denoting $\pm 1$ standard deviation.\vspace{-1em}}
	\label{Exp1_0}
\end{figure*}

\subsubsection{Performance Metrics}\label{sec:7_metrics}
Along with the long-term risk in \eqref{eq_miscov} and the set size (see the dataset descriptions in \ref{sec_7_A_1}), we report the tracking load, which is given by
\begin{equation}
	\mathrm{TrackingLoad}(T) = \frac{1}{T}\sum_{t=1}^{T}|\mathcal{K}_t^{+}| ,
\end{equation}
with $\mathcal{K}_t^{+}$ in \eqref{eq:tracking-plus}.
By construction, the tracking load of the capacity-constrained schemes is no larger than the capacity $K$, while, for OCRC, the set $\mathcal{K}_t^{+}$ contains all rounds with pending feedback.

\subsection{Tumor Segmentation}\label{sec:exp_tumor}

In this subsection, we evaluate CA-OCRC on the tumor segmentation task under i.i.d. delays, i.e., $\rho =0$.
The proposed scheme, referred to as ``CA-OCRC (Proposed)'', adopts the proxy delay distribution in Proposition~\ref{prop:iid-opt}.
We also consider two alternative proxy delay distributions.
The first, referred to as ``CA-OCRC (Bernoulli)'', is the distribution in Proposition~\ref{prop:wc-opt}, which does not use knowledge of the delay distribution $G$.
The second, referred to as ``CA-OCRC ($F=G$)'', is the heuristic choice $F=G$.
We set the maximum delay to $d_{\max}=150$, the target level to $\alpha=0.1$, and the time horizon to $T=2151$ rounds.
We set $\nu=4$ and $\theta=8$ in \eqref{eq:gamma-ar1}, yielding a mean delay of $\nu \theta=32$ rounds.

\textit{1) Impact of the Capacity:}
Fig.~\ref{Exp1_0} shows the FNR as a function of the time horizon $T$ for capacities $K\in\{1,2,4\}$.
With unlimited capacity, OCRC approaches the target level quickly by tracking all rounds.
In contrast, the FNR of FIFO-OCRC remains close to one for $K=1$, and it decreases only slowly for large capacity.
In fact, FIFO-OCRC receives the feedback of round $t$ only if $d_t\le K-1$, which rarely occurs under the considered delay distribution.
All versions of CA-OCRC converge faster as $K$ increases, in line with Corollary~\ref{cor:iid-rate}, with CA-OCRC (Proposed) being the fastest capacity-constrained scheme for all values of $K$.

\begin{figure*}[t]
	\centering
	\subfigure[]{\label{Exp1_1a}
		\includegraphics[width=0.32\textwidth]{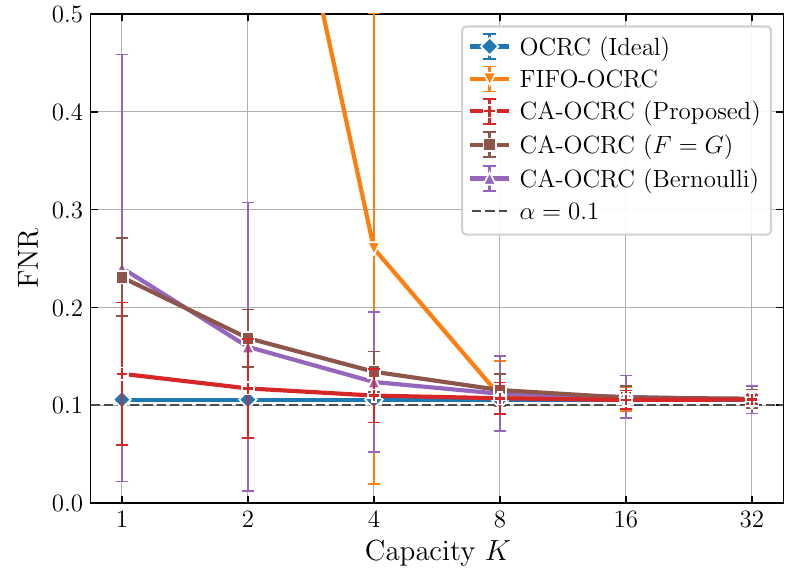}
	}
	\hspace{-1em}
	\subfigure[]{\label{Exp1_1b}
		\includegraphics[width=0.32\textwidth]{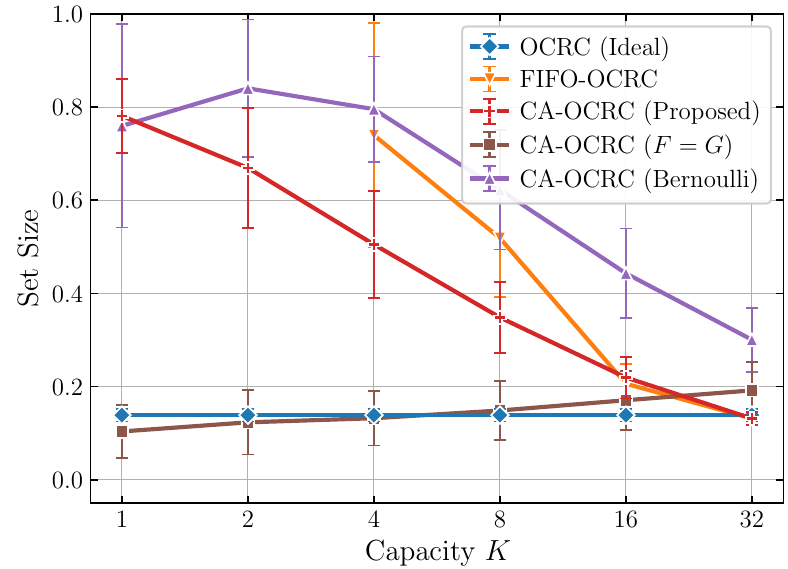}
	}
	\hspace{-1em}
	\subfigure[]{\label{Exp1_1c}
		\includegraphics[width=0.32\textwidth]{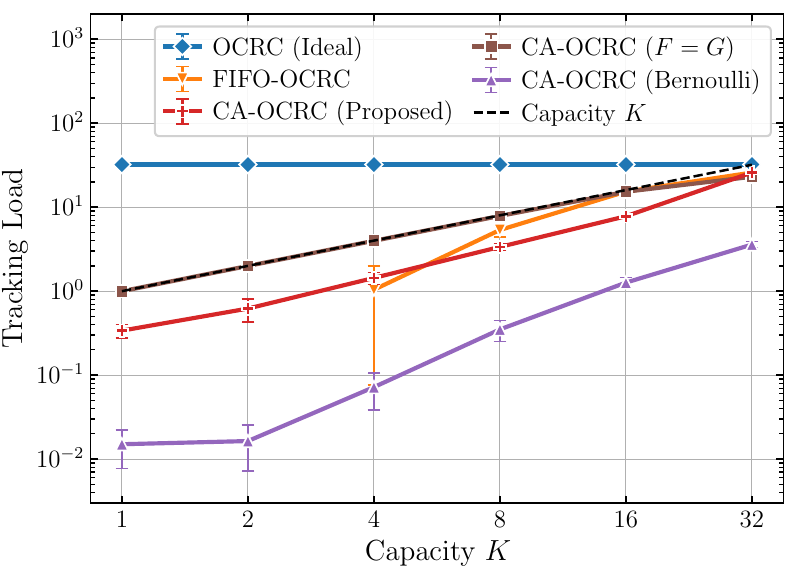}
	}\vspace{-0.5em}
	\caption{(a) FNR, (b) set size, and (c) tracking load as a function of the capacity $K$ for the tumor segmentation task with $\nu=4$, $\theta=8$, $\rho =0$, $d_{\max}=150$, target level $\alpha=0.1$, and time horizon $T=2151$. The results are averaged over $10^3$ independent trials, with the error bars denoting $\pm 1$ standard deviation.
	FIFO-OCRC, which fails for $K \le 2$ (see Fig. \ref{Exp1_1a}), is shown only for $K \ge 4$ in Figs. \ref{Exp1_1b} and \ref{Exp1_1c}.\vspace{-1em}}
	\label{Exp1_1}
\end{figure*}

Fig.~\ref{Exp1_1} shows that, as the capacity $K$ grows, the FNRs of all variants of CA-OCRC approach the target level, while that of FIFO-OCRC only achieves for $K \ge 8$.
The set sizes of CA-OCRC (Proposed) and CA-OCRC (Bernoulli) generally decrease with $K$, since more feedback signals become available for calibration, and both schemes keep the tracking load below $K$.
CA-OCRC (Bernoulli), being designed for arbitrary delays, has the lowest tracking load, while CA-OCRC (Proposed) exploits knowledge of $G$ to use it more effectively, yielding smaller masks that match those of OCRC for $K=32$.
Note that the small masks of CA-OCRC ($F=G$) for $K\le 4$ are not evidence of efficiency, as they come with FNR values above the target level.
In fact, stochastic tracking load of CA-OCRC ($F=G$) in \eqref{eq:stoch-load} equals $\ell_{F,G}=\sum_{a=1}^{d_{\max}}G(a)^2\approx 23$, and exceeds $K$ for $K\le 16$, violating Assumption~\ref{assum:stoch-load}, and the feedback lost at saturation causes the FNR to move away from the target level.

\begin{figure*}[t]
	\centering
	\subfigure[]{\label{Exp1_2a}
		\includegraphics[width=0.32\textwidth]{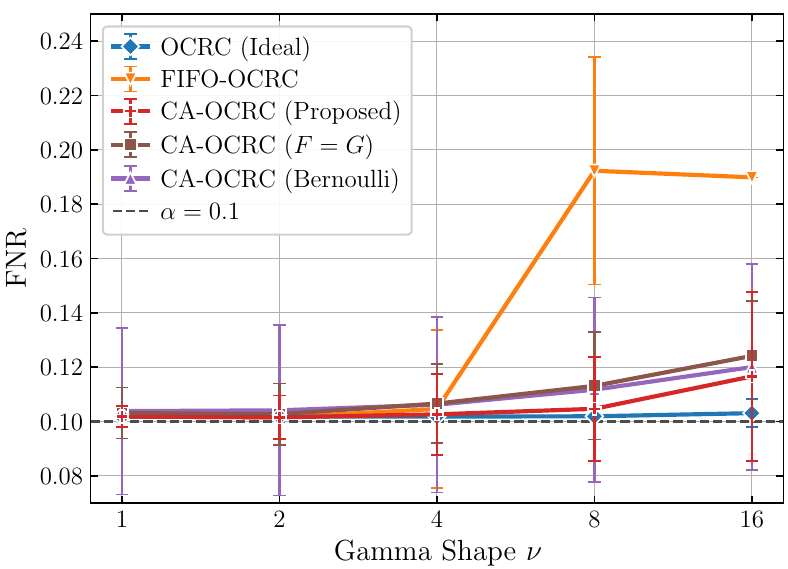}
	}
	\hspace{-1em}
	\subfigure[]{\label{Exp1_2b}
		\includegraphics[width=0.32\textwidth]{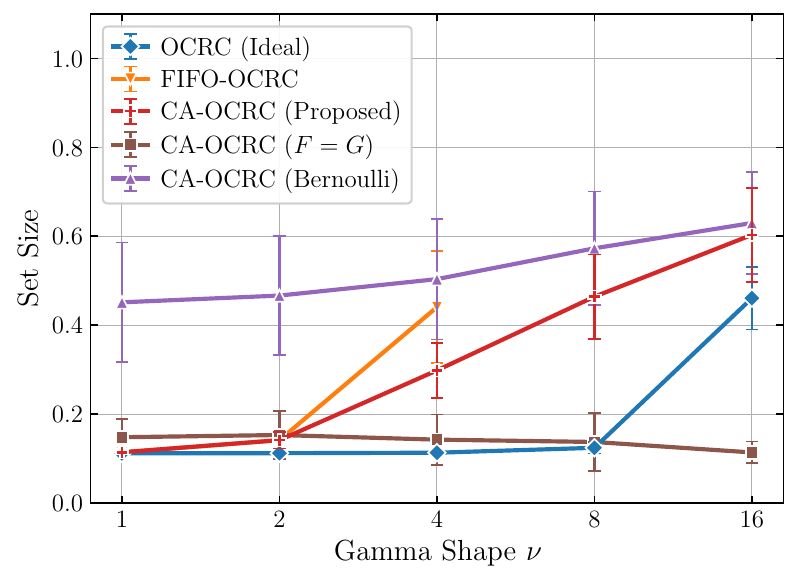}
	}
	\hspace{-1em}
	\subfigure[]{\label{Exp1_2c}
		\includegraphics[width=0.32\textwidth]{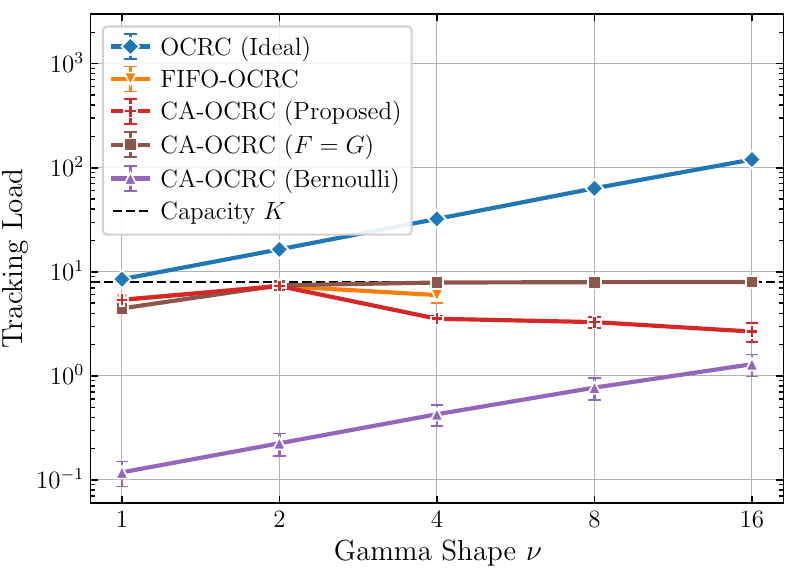}
	}\vspace{-0.5em}
	\caption{(a) FNR, (b) set size, and (c) tracking load as a function of the shape $\nu$ for the tumor segmentation task with $\theta=8$, $\rho =0$, $d_{\max}=150$, capacity $K=8$, target level $\alpha=0.1$, and time horizon $T=2151$. The results are averaged over $10^3$ independent trials, with the error bars denoting $\pm 1$ standard deviation.
	FIFO-OCRC, which fails for $\nu \ge 8$ (see Fig. \ref{Exp1_2a}), is shown only for $\nu \le 4$ in Figs. \ref{Exp1_2b} and \ref{Exp1_2c}.\vspace{-1em}}
	\label{Exp1_2}
\end{figure*}

\textit{2) Impact of the Delay Distribution:}
We now fix the capacity to $K=8$, and change the shape $\nu$ in \eqref{eq:gamma-ar1}, with scale $\theta=8$.
Larger values of the shape parameter $\nu$ yield longer and more concentrated delays, with mean $\nu\theta$ ranging from $8$ to $128$ rounds.
Accordingly, the tracking load of OCRC exceeds the capacity for for all values of $\nu$ considered (see Fig.~\ref{Exp1_2c}).
As seen in Fig.~\ref{Exp1_2a}, CA-OCRC (Proposed) and CA-OCRC (Bernoulli) keep the FNR close to the target level for all values of the shape parameter $\nu$, while FIFO-OCRC fails for $\nu \ge 8$.
Fig.~\ref{Exp1_2b} shows that the set sizes of OCRC and CA-OCRC (Proposed) increase with $\nu$, since longer delays make the feedback used by the threshold updates more outdated.

The advantage of CA-OCRC (Proposed) over CA-OCRC (Bernoulli) in terms of mask size is largest for short delays, and it shrinks as the shape parameter $\nu$ increases (see Fig.~\ref{Exp1_2b}).
In fact, the Bernoulli design in Proposition~\ref{prop:wc-opt} is tailored to the worst case in which all delays equal $d_{\max}$, a scenario that becomes increasingly representative as the mean delay $\nu\theta$ approaches $d_{\max}=150$.
This is in line with Corollary~\ref{cor:wc-rate} and Corollary~\ref{cor:iid-rate}, whose rates depend on the maximum delay $d_{\max}$ and on the mean delay $\mathbb{E}_G[d_t]$, respectively, so that knowledge of the delay distribution $G$ is most valuable when $\mathbb{E}_G[d_t]$ is much smaller than $d_{\max}$.

\subsection{Open-Domain Question Answering}\label{sec:exp_QA}

In this subsection, we evaluate CA-OCRC on the open-domain question answering task, again under i.i.d. delays.
The benchmark settings are the same as in Sec.~\ref{sec:exp_tumor}.
We set the maximum delay to $d_{\max}=100$, the target level to $\alpha=0.3$ as in \cite{angelopoulos2024conformal}, and the time horizon to $T=3610$ rounds, with each question appearing once.
We set $\nu=8$ and $\theta=3.75$ in \eqref{eq:gamma-ar1}, yielding a mean delay of $\nu\theta=30$ rounds.

\begin{figure*}[t]
	\centering
	\subfigure[]{\label{Exp2_0a}
		\includegraphics[width=0.32\textwidth]{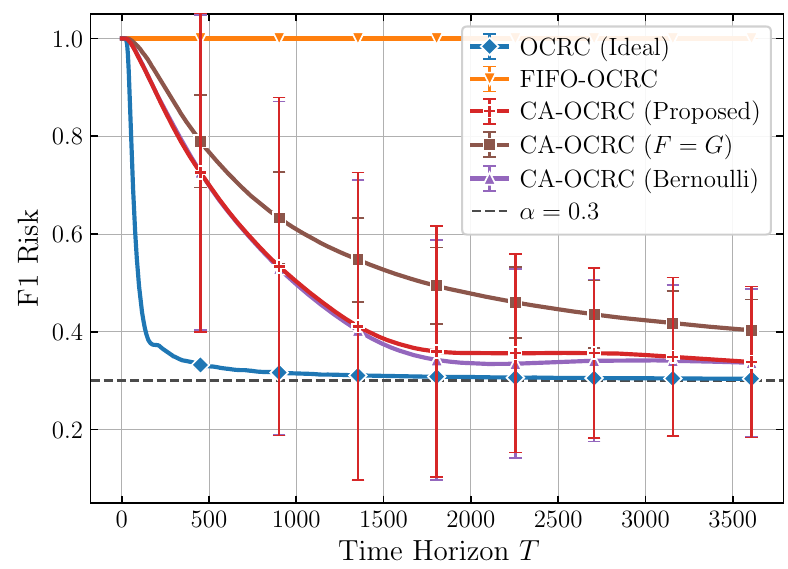}
	}
	\hspace{-1em}
	\subfigure[]{\label{Exp2_0b}
		\includegraphics[width=0.32\textwidth]{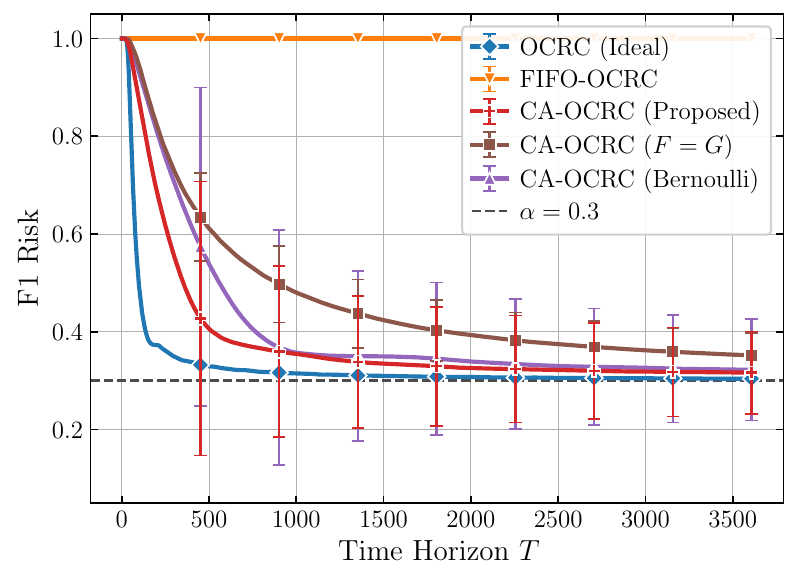}
	}
	\hspace{-1em}
	\subfigure[]{\label{Exp2_0c}
		\includegraphics[width=0.32\textwidth]{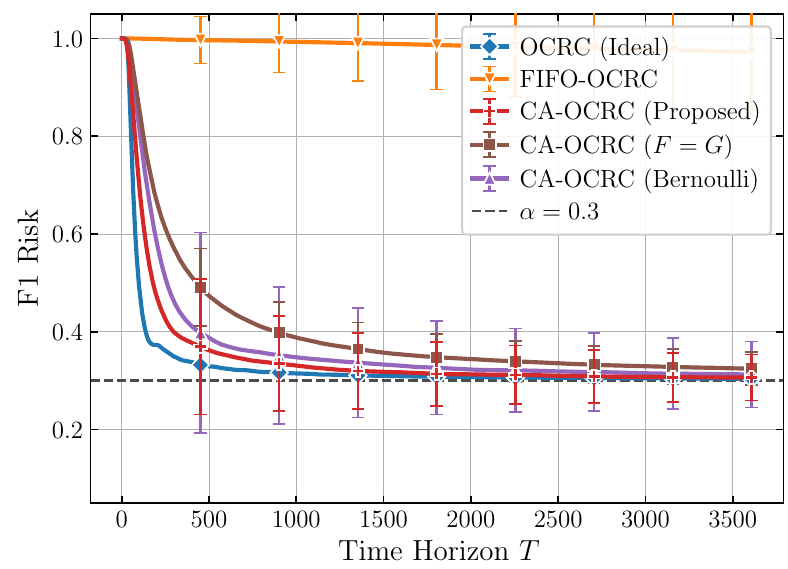}
	}\vspace{-0.5em}
	\caption{F1 risk as a function of the time horizon $T$ for the open-domain question answering task with $\nu=8$, $\theta=3.75$, $\rho =0$, $d_{\max}=100$, and target level $\alpha=0.3$ (dashed line), for capacities (a) $K=1$, (b) $K=2$, and (c) $K=4$. The results are averaged over $10^3$ independent trials, with the error bars denoting $\pm 1$ standard deviation.\vspace{-1em}}
	\label{Exp2_0}
\end{figure*}

\textit{1) Impact of the Capacity:}
Fig.~\ref{Exp2_0} shows the F1 risk as a function of the time horizon $T$ for capacities $K\in\{1,2,4\}$.
Since short delays are rare, FIFO-OCRC, which obtains almost no feedback for these capacities, thus, its F1 risk remains close to one.
In contrast, all versions of CA-OCRC converge faster as $K$ increases, in line with Corollary~\ref{cor:iid-rate}, with CA-OCRC (Proposed) converging fastest for $K\in\{2,4\}$ and performing comparably to CA-OCRC (Bernoulli) for $K=1$.

\begin{figure*}[t]
	\centering
	\subfigure[]{\label{Exp2_1a}
		\includegraphics[width=0.32\textwidth]{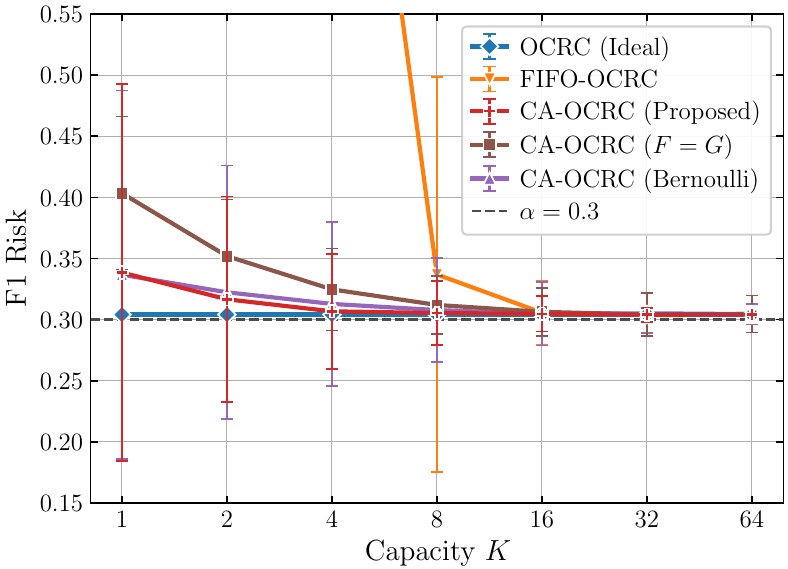}
	}
	\hspace{-1em}
	\subfigure[]{\label{Exp2_1b}
		\includegraphics[width=0.32\textwidth]{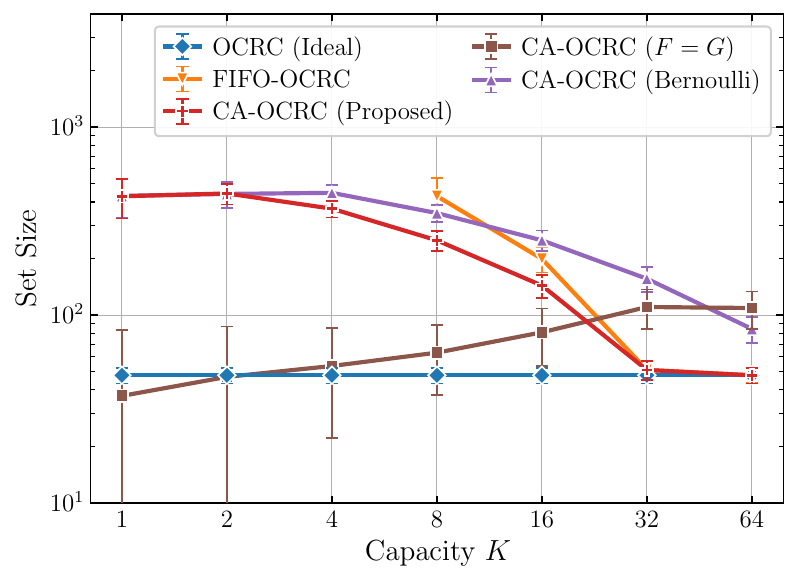}
	}
	\hspace{-1em}
	\subfigure[]{\label{Exp2_1c}
		\includegraphics[width=0.32\textwidth]{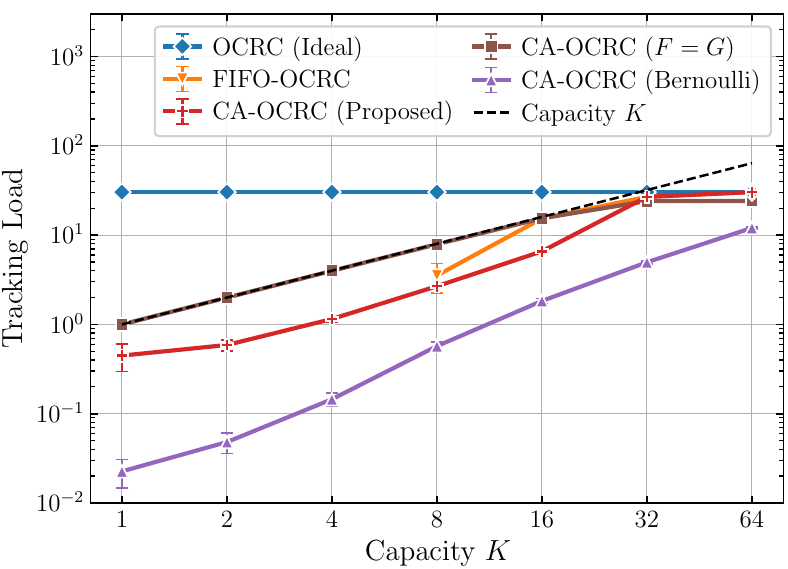}
	}\vspace{-0.5em}
	\caption{(a) F1 risk, (b) set size, and (c) tracking load as a function of the capacity $K$ for the open-domain question answering task with $\nu=8$, $\theta=3.75$, $\rho =0$, $d_{\max}=100$, target level $\alpha=0.3$, and time horizon $T=3610$. The results are averaged over $10^3$ independent trials, with the error bars denoting $\pm 1$ standard deviation.
	FIFO-OCRC, which fails for $K \le 4$ (see Fig. \ref{Exp2_1a}), is shown only for $K \ge 8$ in Figs. \ref{Exp2_1b} and \ref{Exp2_1c}.\vspace{-1em}}
	\label{Exp2_1}
\end{figure*}

Fig.~\ref{Exp2_1} shows that, as $K$ grows, the F1 risks of all versions of CA-OCRC approach the target level, while that of FIFO-OCRC does so only for $K\ge 16$.
CA-OCRC (Bernoulli) uses the least amount of capacity, since its design targets the worst case and is thus the most conservative.
By exploiting knowledge of the delay distribution $G$, CA-OCRC (Proposed) yields smaller sets than CA-OCRC (Bernoulli) for $K\ge 4$, and its set size approaches that of OCRC for $K=32$ (see Fig.~\ref{Exp2_1b}).
In contrast, CA-OCRC ($F=G$), whose stochastic tracking load $\ell_{F,G}\approx 24$ exceeds $K$ for $K\le 16$, operates at full capacity in this regime, which results in F1 risks well above the target level for $K\le 2$.

\begin{figure*}[t]
	\centering
	\subfigure[]{\label{Exp2_2a}
		\includegraphics[width=0.32\textwidth]{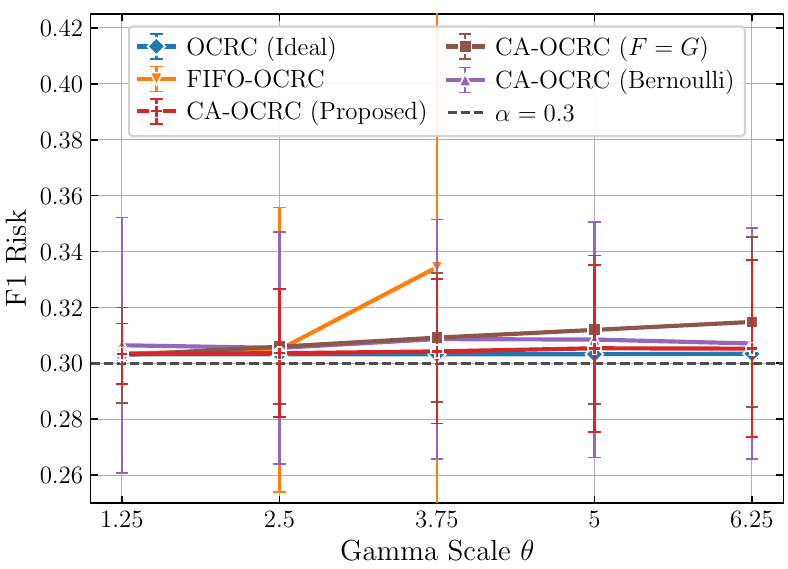}
	}
	\hspace{-1em}
	\subfigure[]{\label{Exp2_2b}
		\includegraphics[width=0.32\textwidth]{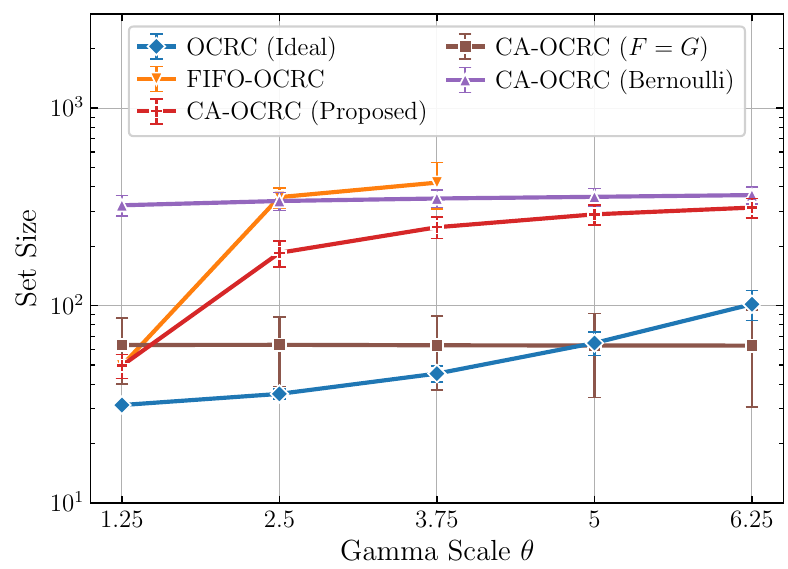}
	}
	\hspace{-1em}
	\subfigure[]{\label{Exp2_2c}
		\includegraphics[width=0.32\textwidth]{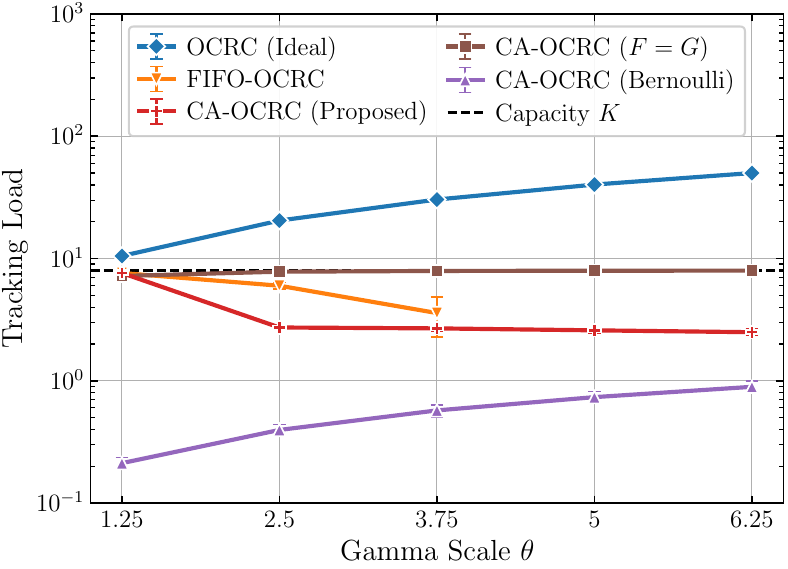}
	}\vspace{-0.5em}
	\caption{(a) F1 risk, (b) set size, and (c) tracking load as a function of the scale parameter $\theta$ for the open-domain question answering task with $\nu=8$, $\rho =0$, $d_{\max}=100$, capacity $K=8$, target level $\alpha=0.3$, and time horizon $T=3610$. The results are averaged over $10^3$ independent trials, with the error bars denoting $\pm 1$ standard deviation.
	FIFO-OCRC, which fails for $\theta \ge 5$ (see Fig. \ref{Exp2_2a}), is shown only for $\theta \le 3.75$ in Figs. \ref{Exp2_2b} and \ref{Exp2_2c}.\vspace{-1em}}
	\label{Exp2_2}
\end{figure*}

\textit{2) Impact of the Delay Distribution:}
Fig.~\ref{Exp2_2} shows the impact of the delay distribution for $K=8$.
Unlike in Sec.~\ref{sec:exp_tumor}, we fix the shape to $\nu=8$, and change the scale $\theta$ in \eqref{eq:gamma-ar1}, which stretches the delays without changing their relative spread, with mean $\nu\theta$ ranging from $10$ to $50$ rounds and variance $\nu\theta^2$ ranging from $12.5$ to $312.5$.
CA-OCRC (Proposed) and CA-OCRC (Bernoulli) keep the F1 risk close to the target level for all values of $\theta$, while FIFO-OCRC fails for $\theta\ge 5$.
Due to the concentration of the delays, the tracking load of CA-OCRC (Proposed) decreases sharply from nearly the full capacity to half capacity when $\theta$ increases from $1.25$ to $2.5$, and it then remains nearly constant as $\theta$ increases (see Fig.~\ref{Exp2_2c}).

With a constant tracking load, CA-OCRC (Proposed) receives fewer feedback signals as the delays become longer, and these signals are also more outdated, so that its set size grows with $\theta$.
In contrast, since the probability $p^\star$ in Proposition~\ref{prop:wc-opt} does not depend on $\theta$, CA-OCRC (Bernoulli) receives the feedback of about the same number of rounds for all values of $\theta$, producing large sets of nearly constant size, while its tracking load grows with the mean delay.
Finally, for $\theta \ge 2.5$, CA-OCRC ($F=G$) again produces smaller sets than CA-OCRC (Proposed) at the cost of operating close to saturation, so that its F1 risk moves away from the target level as $\theta$ increases.
Overall, the distribution-aware design in Proposition~\ref{prop:iid-opt} uses the capacity more effectively than the Bernoulli design in Proposition~\ref{prop:wc-opt}, yielding faster convergence and smaller sets.

\subsection{Hierarchical Image Classification}\label{sec:exp_Image}

In this subsection, we evaluate CA-OCRC on the hierarchical image classification task under temporally correlated feedback delays, i.e., with $\rho >0$ in \eqref{eq:gamma-ar1}, which violate Assumption~\ref{assum:delay}(b), unlike in Sec.~\ref{sec:exp_tumor} and Sec.~\ref{sec:exp_QA}.
The proposed scheme, referred to as ``CA-OCRC (Proposed)'', thus adopts the proxy delay distribution in Proposition~\ref{prop:wc-opt}, with $\bar{d}_{\max}=d_{\max}$ and $q_G(\bar{d}_{\max})=1$ in \eqref{eq_ca_ocp}, which requires no knowledge of the delay beyond $d_{\max}$.
The oracle benchmark ``CA-OCRC (best fixed $p$)'' adopts the same proxy delay distribution, but with the value of $p$ that minimizes the risk gap, obtained through a grid search.
Since the delays are identically distributed, with the survival function $G$ in \eqref{eq:gamma-G}, but not independent, a natural alternative, referred to as ``CA-OCRC (i.i.d. design)'', is to adopt the proxy delay distribution in Proposition~\ref{prop:iid-opt}.
However, CA-OCRC (i.i.d. design) is not covered by the guarantee in Theorem~\ref{thm_iid_miscoverage}, which relies on the independence of the delays.
We set $d_{\max}=100$, $\alpha=0.02$ as in \cite{angelopoulos2024conformal}, and $T=2\times 10^4$, along with $\nu=2$, $\theta=8$, and $\rho =0.999$ in \eqref{eq:gamma-ar1}, yielding a mean delay of $16$ rounds and a correlation time $1/(1-\rho )$ of $10^3$ rounds.

\begin{figure*}[t]
	\centering
	\subfigure[]{\label{Exp3_0a}
		\includegraphics[width=0.32\textwidth]{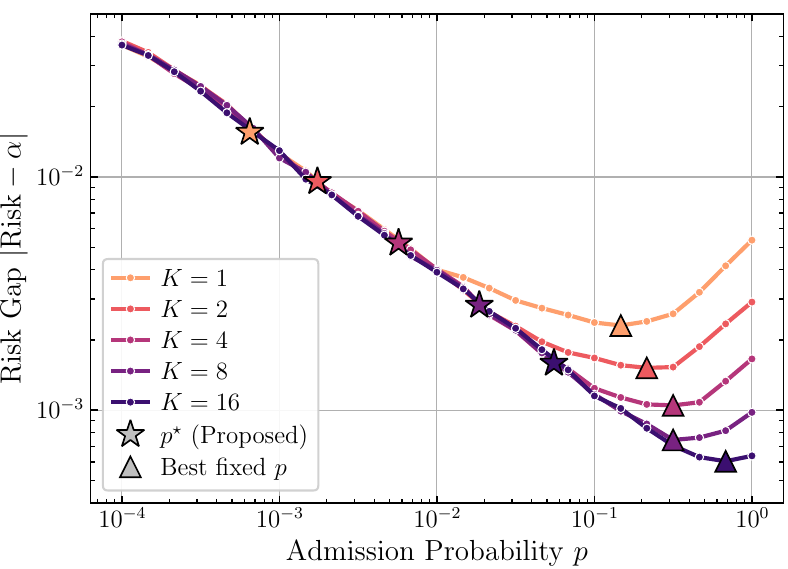}
	}
	\hspace{-1em}
	\subfigure[]{\label{Exp3_0b}
		\includegraphics[width=0.32\textwidth]{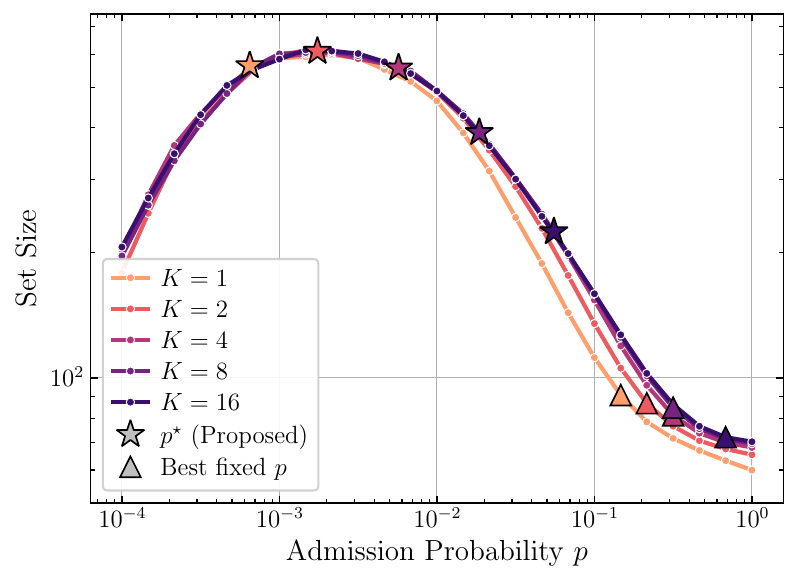}
	}
	\hspace{-1em}
	\subfigure[]{\label{Exp3_0c}
		\includegraphics[width=0.32\textwidth]{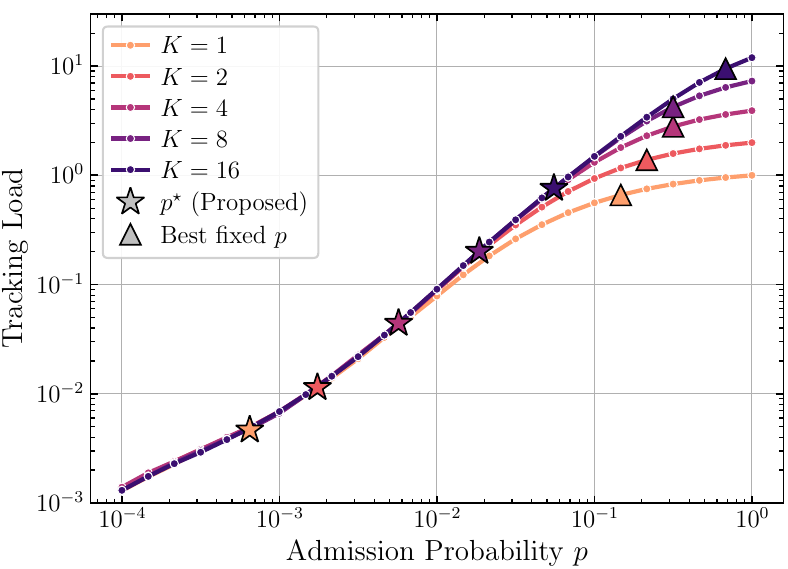}
	}\vspace{-0.5em}
	\caption{(a) Risk gap, (b) set size, and (c) tracking load of CA-OCRC with a fixed admission probability $p$ as a function of $p$ for the hierarchical image classification task with $\nu=2$, $\theta=8$, $\rho =0.999$, $d_{\max}=100$, capacities $K\in\{1,2,4,8,16\}$, target level $\alpha=0.02$, and time horizon $T=2\times 10^4$. Stars mark the probability $p^\star$ in Proposition~\ref{prop:wc-opt}, and triangles mark the minimum of the risk gap. The results are averaged over $10^3$ independent trials.\vspace{-1em}}
	\label{Exp3_0}
\end{figure*}

\textit{1) Choice of the Admission Probability:}
Fig.~\ref{Exp3_0} evaluates CA-OCRC with a fixed admission probability $p$, i.e., with $F(a)=p$ for $a\in\{1,\ldots,d_{\max}\}$.
In line with the bound in Theorem~\ref{thm:worstcase_miscoverage}, the risk gap first decreases with $p$, as more feedback is received, and, for small capacities, it then increases once capacity saturation becomes frequent.
The probability $p^\star$ lies to the left of the minimum, i.e., it is smaller than the value of $p$ minimizing the risk gap.
This conservative choice keeps the tracking load well below $K$ (see Fig.~\ref{Exp3_0c}), so that capacity saturation is rare, at the cost of larger sets (see Fig.~\ref{Exp3_0b}).
Such conservatism is consistent with the worst-case design in Proposition~\ref{prop:wc-opt}.

\begin{figure*}[t]
	\centering
	\subfigure[]{\label{Exp3_1a}
		\includegraphics[width=0.32\textwidth]{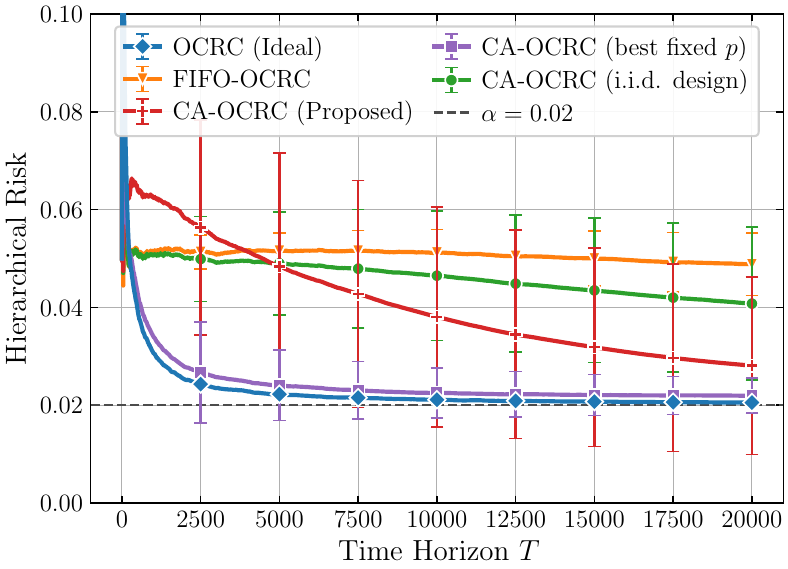}
	}
	\hspace{-1em}
	\subfigure[]{\label{Exp3_1b}
		\includegraphics[width=0.32\textwidth]{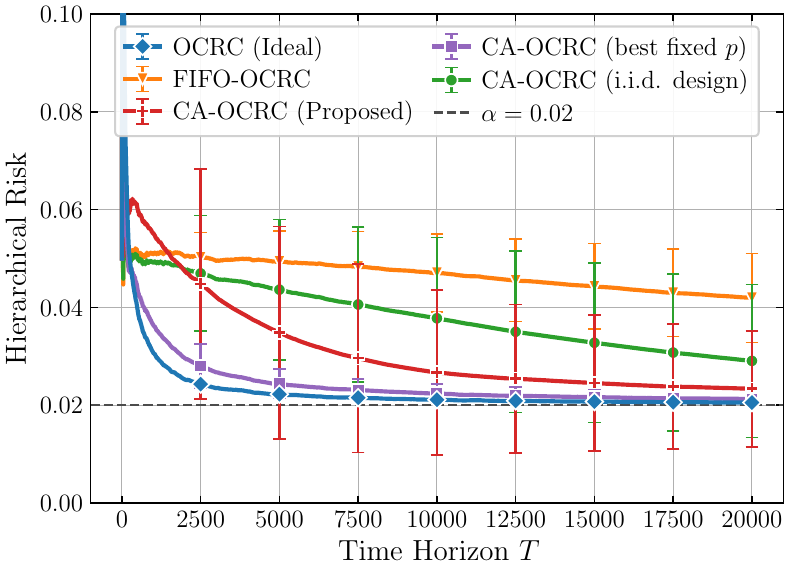}
	}
	\hspace{-1em}
	\subfigure[]{\label{Exp3_1c}
		\includegraphics[width=0.32\textwidth]{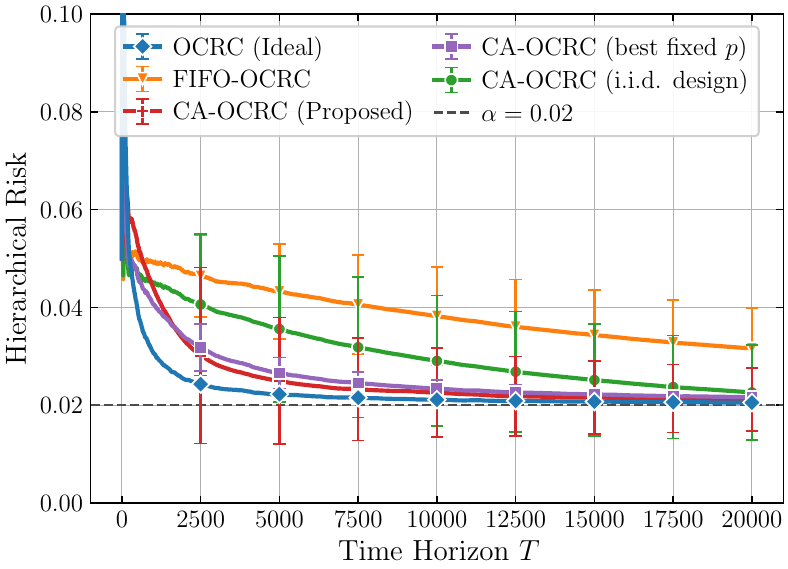}
	}\vspace{-0.5em}
	\caption{Hierarchical risk as a function of the time horizon $T$ for the hierarchical image classification task with $\nu=2$, $\theta=8$, $\rho =0.999$, $d_{\max}=100$, and target level $\alpha=0.02$ (dashed line), for capacities (a) $K=1$, (b) $K=2$, and (c) $K=4$. The results are averaged over $10^3$ independent trials, with the error bars denoting $\pm 1$ standard deviation.\vspace{-1em}}
	\label{Exp3_1}
\end{figure*}

\textit{2) Impact of the Capacity:}
Fig.~\ref{Exp3_1} reports the hierarchical risk as a function of the time horizon $T$ for capacities $K\in\{1,2,4\}$.
FIFO-OCRC remains well above the target level for all values of $K$.
CA-OCRC (i.i.d. design), whose design relies on the independence of the delays in Assumption~\ref{assum:delay}(b), also remains above the target level for $K\le 2$.
In contrast, CA-OCRC (Proposed) approaches the target level for all values of $K$, converging faster as $K$ increases, in line with Corollary~\ref{cor:wc-rate}.
For $K=1$, its convergence is slow, with a large standard deviation, since it admits fewer rounds, each with a large inverse probability weight $1/p^\star$, while, for $K=4$, it performs comparably to CA-OCRC (best fixed $p$) and to OCRC.

\begin{figure*}[t]
	\centering
	\subfigure[]{\label{Exp3_2a}
		\includegraphics[width=0.32\textwidth]{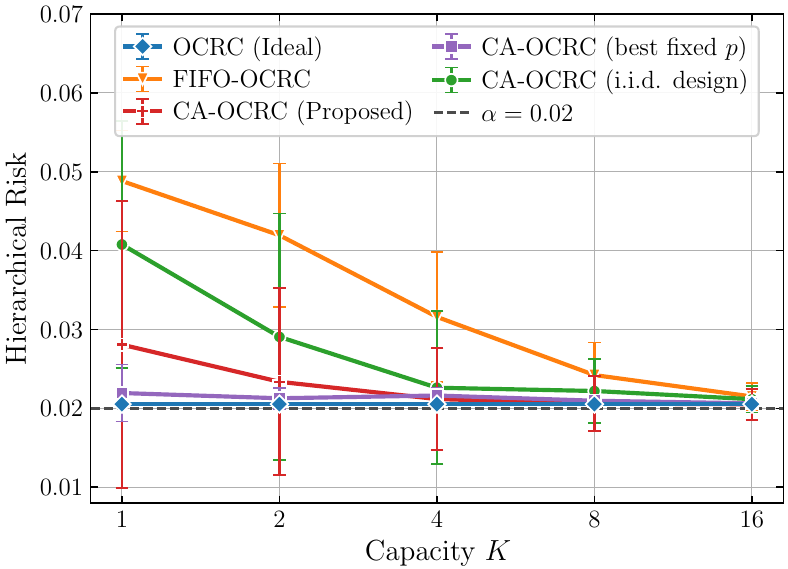}
	}
	\hspace{-1em}
	\subfigure[]{\label{Exp3_2b}
		\includegraphics[width=0.32\textwidth]{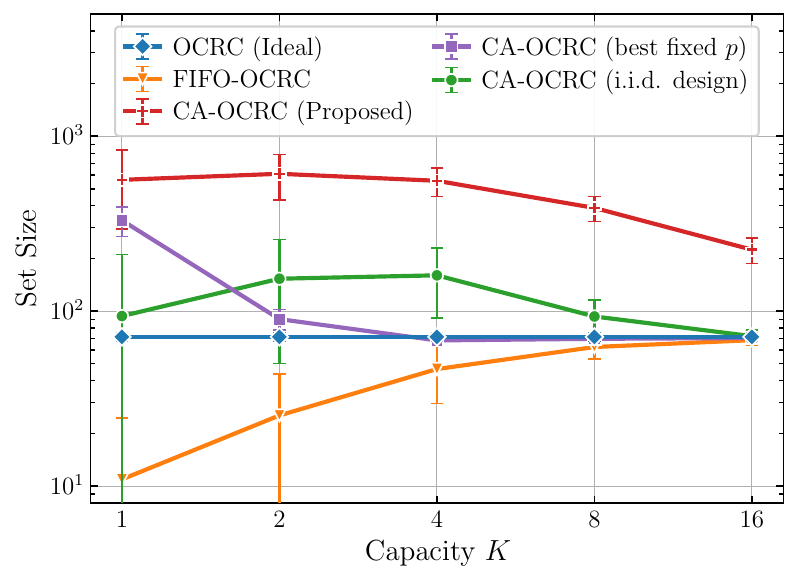}
	}
	\hspace{-1em}
	\subfigure[]{\label{Exp3_2c}
		\includegraphics[width=0.32\textwidth]{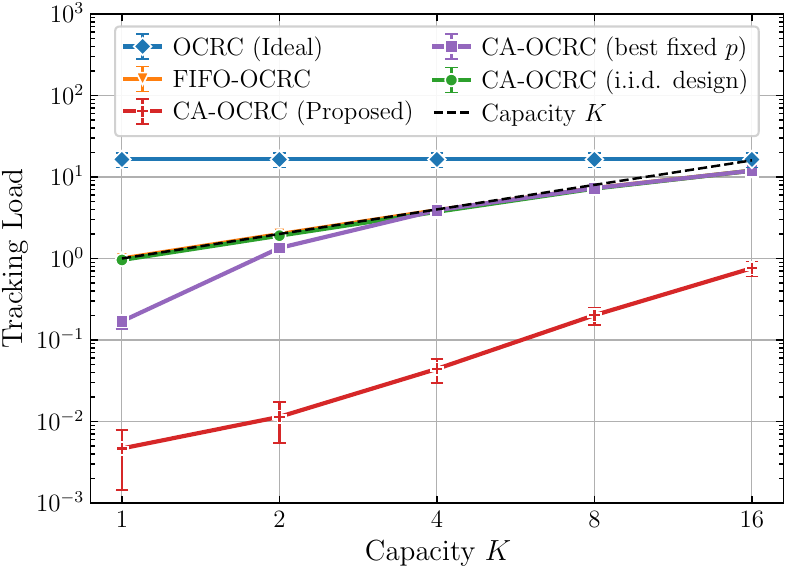}
	}\vspace{-0.5em}
	\caption{(a) Hierarchical risk, (b) set size, and (c) tracking load as a function of the capacity $K$ for the hierarchical image classification task with $\nu=2$, $\theta=8$, $\rho =0.999$, $d_{\max}=100$, target level $\alpha=0.02$, and time horizon $T=2\times 10^4$. The results are averaged over $10^3$ independent trials, with the error bars denoting $\pm 1$ standard deviation.\vspace{-1em}}
	\label{Exp3_2}
\end{figure*}

Fig.~\ref{Exp3_2} shows that, as $K$ grows, the hierarchical risks of all capacity-constrained schemes decrease towards the target level, with CA-OCRC (Proposed) and CA-OCRC (best fixed $p$) being the closest to it for all values of $K$.
Note that the small sets of FIFO-OCRC are not evidence of efficiency, as they come with hierarchical risks above the target level.
The set size of CA-OCRC (Proposed) decreases as $K$ grows, but it remains larger than that of OCRC for all considered capacities, since it keeps the lower tracking load.
In contrast, CA-OCRC (best fixed $p$) uses most of the capacity for $K\ge 4$, matching the set size of OCRC.

\begin{figure*}[t]
	\centering
	\subfigure[]{\label{Exp3_3a}
		\includegraphics[width=0.32\textwidth]{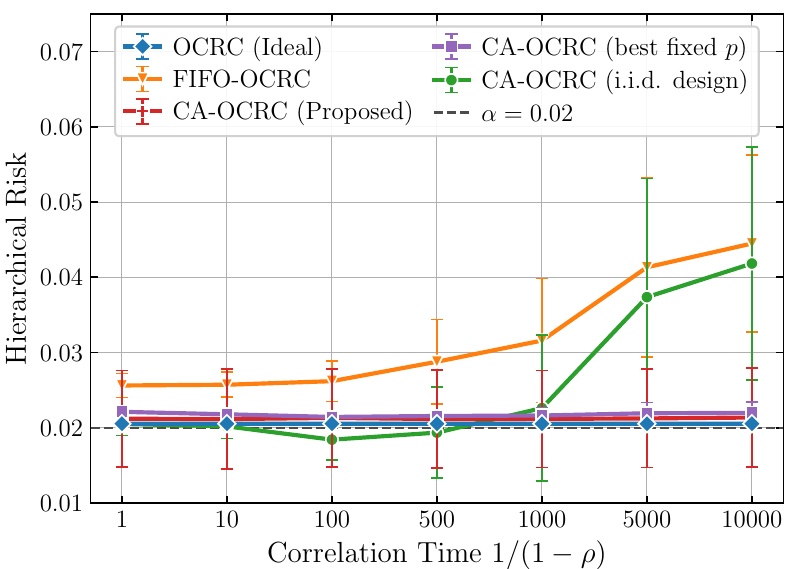}
	}
	\hspace{-1em}
	\subfigure[]{\label{Exp3_3b}
		\includegraphics[width=0.32\textwidth]{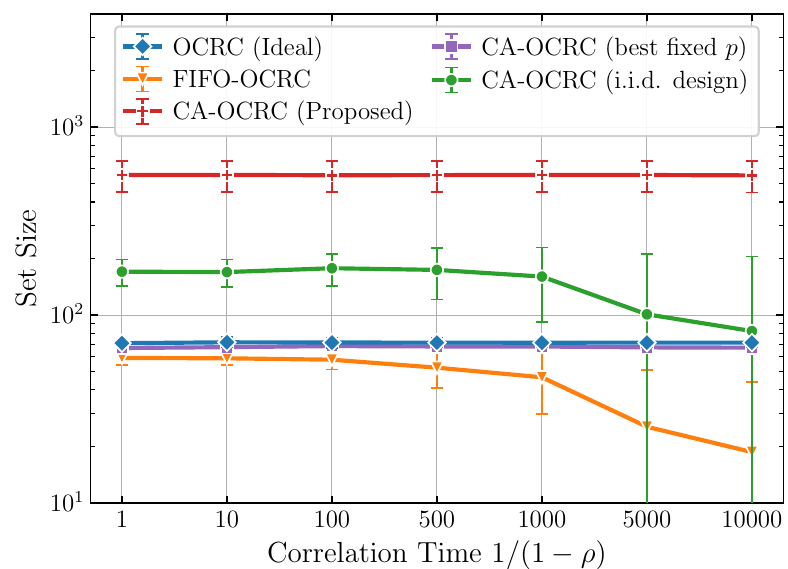}
	}
	\hspace{-1em}
	\subfigure[]{\label{Exp3_3c}
		\includegraphics[width=0.32\textwidth]{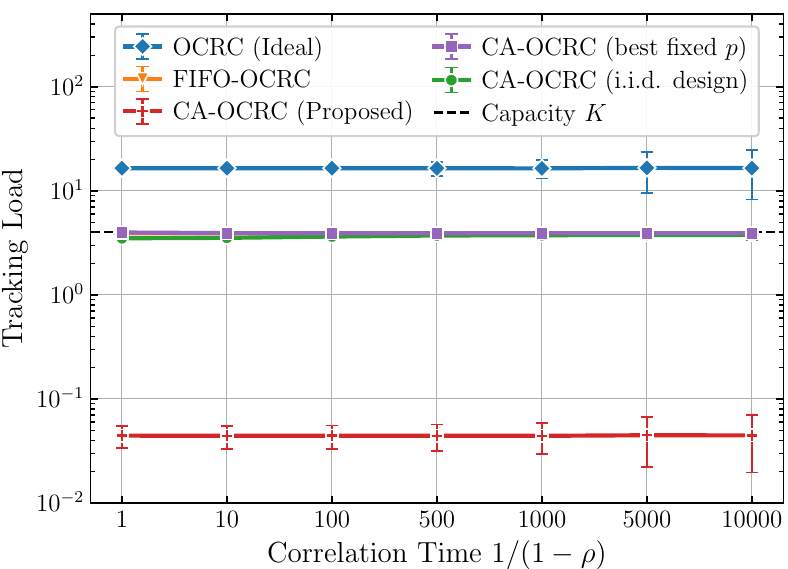}
	}\vspace{-0.5em}
	\caption{(a) Hierarchical risk, (b) set size, and (c) tracking load as a function of the correlation time $1/(1-\rho )$ for the hierarchical image classification task with $\nu=2$, $\theta=8$, $d_{\max}=100$, capacity $K=4$, target level $\alpha=0.02$, and time horizon $T=2\times 10^4$. The results are averaged over $10^3$ independent trials, with the error bars denoting $\pm 1$ standard deviation.\vspace{-1em}}
	\label{Exp3_3}
\end{figure*}

\textit{3) Impact of the Delay Correlation:}
We now fix the capacity to $K=4$, and vary the correlation time $1/(1-\rho )$ in \eqref{eq:gamma-ar1} from $1$, i.e., i.i.d. delays, to $10^4$ rounds.
CA-OCRC (Proposed) keeps the hierarchical risk close to the target level for all correlation times, in line with Theorem~\ref{thm:worstcase_miscoverage}, which holds for any sequence of delays.
In fact, since CA-OCRC (Proposed) admits each round with probability $p^\star$, irrespective of the delays, and tracks it until its feedback is received, its set size and tracking load are also unaffected by the temporal correlation of the delays.
In contrast, the hierarchical risk of FIFO-OCRC exceeds the target level for all correlation times, and it increases with the correlation time, as the feedback signals with delays $d_t\le K-1$ arrive in bursts during periods of short delays.
CA-OCRC (i.i.d. design) keeps the hierarchical risk close to the target level for correlation times up to about $10$ rounds.
For longer correlation times, however, the delays depart further from the i.i.d. model underlying its design, and its risk approaches twice the target level.

\subsection{Beam Selection}\label{sec:exp_beam}

In this subsection, we evaluate CA-OCRC on the beam selection task, also under temporally correlated feedback delays.
The benchmark settings are the same as in Sec.~\ref{sec:exp_Image}.
We set the maximum delay to $d_{\max}=100$, the target level to $\alpha=0.2$, and the time horizon to $T=2\times 10^4$ rounds, with each user appears once.
We set $\nu=2$, $\theta=10$, and $\rho =0.99$ in \eqref{eq:gamma-ar1}, yielding a mean delay of $20$ rounds.

\begin{figure*}[t]
	\centering
	\subfigure[]{\label{Exp4_0a}
		\includegraphics[width=0.32\textwidth]{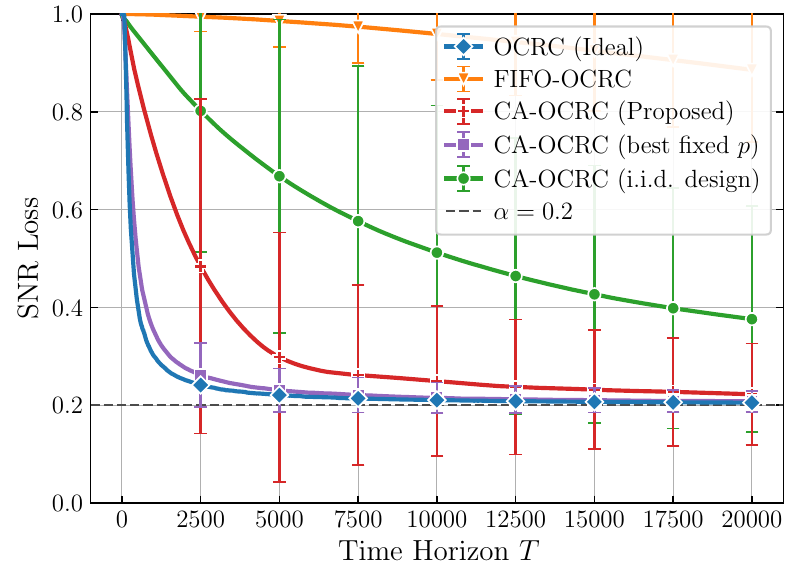}
	}
	\hspace{-1em}
	\subfigure[]{\label{Exp4_0b}
		\includegraphics[width=0.32\textwidth]{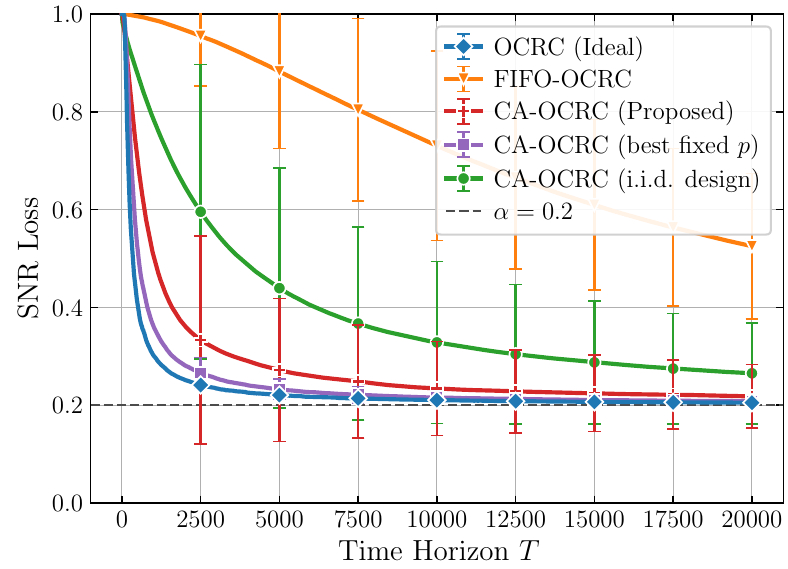}
	}
	\hspace{-1em}
	\subfigure[]{\label{Exp4_0c}
		\includegraphics[width=0.32\textwidth]{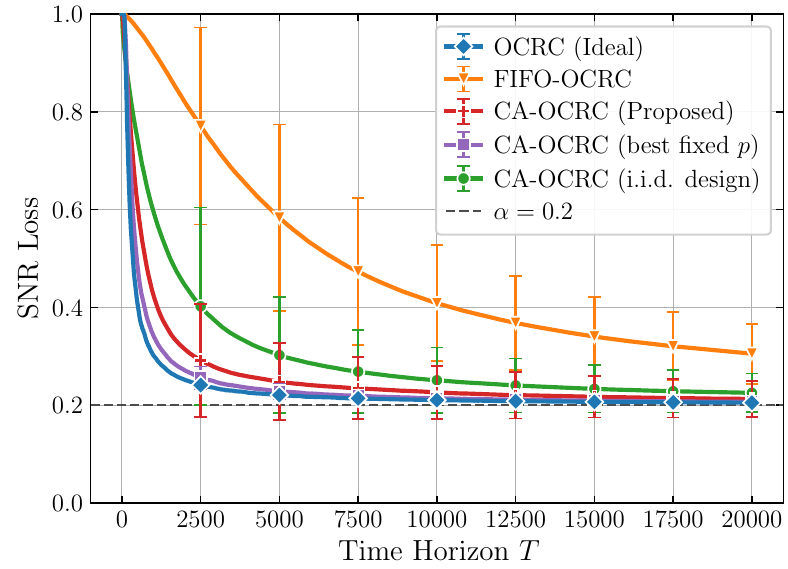}
	}\vspace{-0.5em}
	\caption{SNR loss as a function of the time horizon $T$ for the beam selection task with $\nu=2$, $\theta=10$, $\rho =0.99$, $d_{\max}=100$, and target level $\alpha=0.2$ (dashed line), for capacities (a) $K=1$, (b) $K=2$, and (c) $K=4$. The results are averaged over $10^3$ independent trials, with the error bars denoting $\pm 1$ standard deviation.\vspace{-1em}}
	\label{Exp4_0}
\end{figure*} 

Fig.~\ref{Exp4_0} shows the SNR loss as a function of the time horizon $T$ for capacities $K\in\{1,2,4\}$.
FIFO-OCRC and CA-OCRC (i.i.d. design) converge slowly, remaining above the target level for all values of $K$.
In contrast, CA-OCRC (Proposed) converges faster as $K$ increases, in line with Corollary~\ref{cor:wc-rate}, approaching the target level for $K\ge 2$.
For $K=1$, CA-OCRC (best fixed $p$) converges faster than CA-OCRC (Proposed), while both schemes perform comparably to OCRC for $K=4$.

\begin{figure*}[t]
	\centering
	\subfigure[]{\label{Exp4_1a}
		\includegraphics[width=0.32\textwidth]{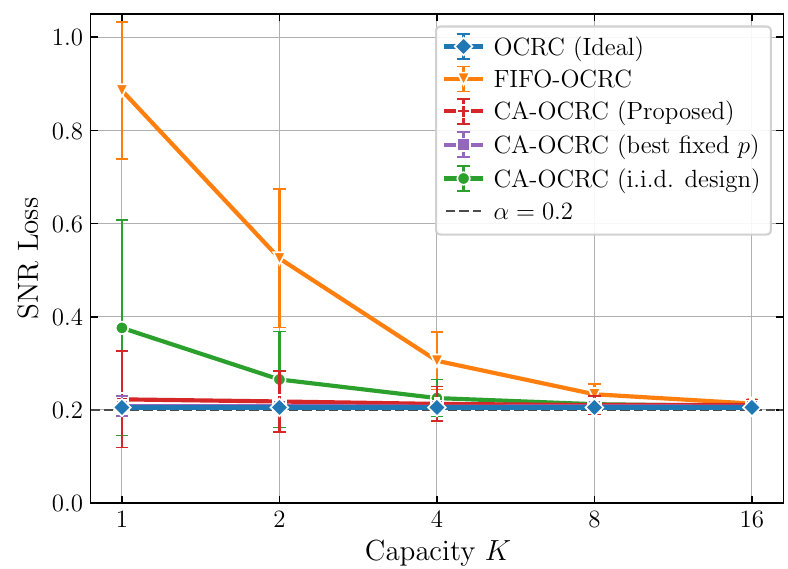}
	}
	\hspace{-1em}
	\subfigure[]{\label{Exp4_1b}
		\includegraphics[width=0.32\textwidth]{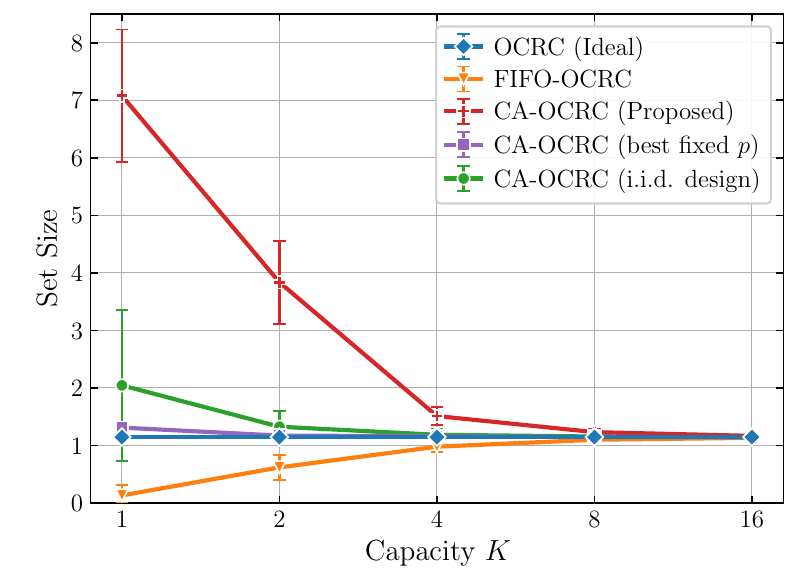}
	}
	\hspace{-1em}
	\subfigure[]{\label{Exp4_1c}
		\includegraphics[width=0.32\textwidth]{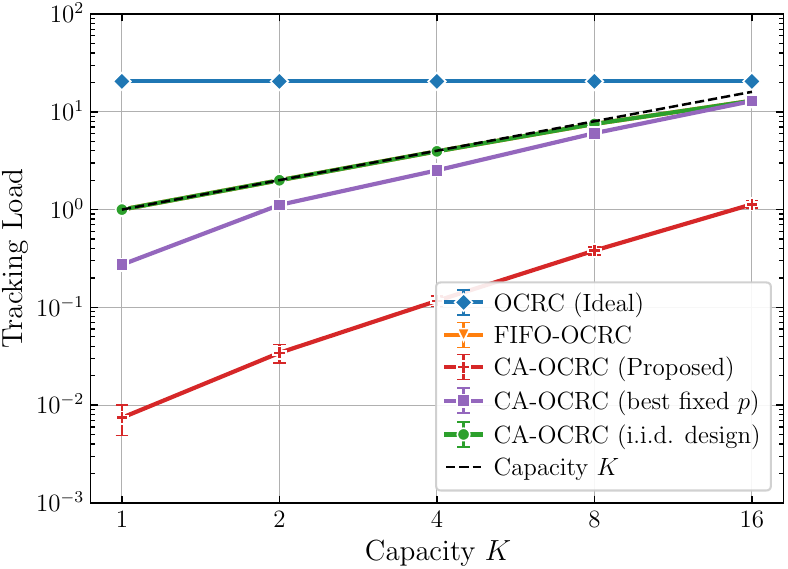}
	}\vspace{-0.5em}
	\caption{(a) SNR loss, (b) set size, and (c) tracking load as a function of the capacity $K$ for the beam selection task with $\nu=2$, $\theta=10$, $\rho =0.99$, $d_{\max}=100$, target level $\alpha=0.2$, and time horizon $T=2\times 10^4$. The results are averaged over $10^3$ independent trials, with the error bars denoting $\pm 1$ standard deviation.\vspace{-1em}}
	\label{Exp4_1}
\end{figure*}

Fig.~\ref{Exp4_1} shows that FIFO-OCRC exceeds the target level for all considered capacities, while CA-OCRC (i.i.d. design) does so for $K\le 4$.
In contrast, CA-OCRC (Proposed) approaches the target level as $K$ grows.
The set size of CA-OCRC (Proposed) decreases from about $7$ beams for $K=1$ to approximately that of OCRC for $K\ge 8$ (see Fig.~\ref{Exp4_1b}).
Finally, Fig.~\ref{Exp4_1c} shows that OCRC violates the capacity constraint for all considered values of $K$, while CA-OCRC (Proposed) keeps the tracking load well below $K$.
Overall, the worst-case design in Proposition~\ref{prop:wc-opt} enables CA-OCRC to approach the target level irrespective of the correlation time, unlike the distribution-aware design in Proposition~\ref{prop:iid-opt}, however, at the cost of larger sets than the oracle choice of $p$.

\section{Conclusion}\label{sec:8}

We have studied OCRC under capacity-constrained delayed feedback, in which at most $K$ rounds with pending feedback can be tracked at any time.
For this setting, we have proposed capacity-aware OCRC (CA-OCRC), which tracks each round, if capacity permits, until its feedback is received or a random proxy delay expires, and compensates for the lost feedback via a two-level form of inverse probability weighting.
We have analyzed the risk boundsof CA-OCRC under arbitrary and stochastic delays, and we have optimized the proxy delay distribution based on these bounds.
With the optimized designs, the risk gap decays as $\mathcal{O}((d_{\max}/T)^{K/(K+1)})$ under arbitrary delays, so that a capacity logarithmic in $T/d_{\max}$ suffices to match the convergence rate of OCRC with unlimited capacity.
Under stochastic delays, the maximum delay $d_{\max}$ in this rate is replaced by the mean delay $\mathbb{E}[d_t]$, by exploiting knowledge of the delay distribution.

Experiments on tumor segmentation, open-domain question answering, hierarchical image classification, and beam selection have verified the proposed methods and the theoretical findings.
In all tasks, CA-OCRC approached the target risk level while keeping the tracking load below the capacity.
Under stochastic delays, the distribution-aware design used the capacity more effectively than the worst-case design, yielding faster convergence and smaller prediction sets.

Future work includes the extension of CA-OCRC to general bounded losses that are non-increasing \cite{angelopoulos2024conformal}, the design of adaptive tracking policies that require no prior knowledge of the delay distribution, as well as the development of parameter-free capacity-aware algorithms \cite{orabona2019modern}.

\footnotesize
\bibliographystyle{IEEEtran}
\bibliography{IEEEabrv,./ref/ref.bib}

\begin{thebibliography}{10}
\providecommand{\url}[1]{#1}
\csname url@samestyle\endcsname
\providecommand{\newblock}{\relax}
\providecommand{\bibinfo}[2]{#2}
\providecommand{\BIBentrySTDinterwordspacing}{\spaceskip=0pt\relax}
\providecommand{\BIBentryALTinterwordstretchfactor}{4}
\providecommand{\BIBentryALTinterwordspacing}{\spaceskip=\fontdimen2\font plus
\BIBentryALTinterwordstretchfactor\fontdimen3\font minus
  \fontdimen4\font\relax}
\providecommand{\BIBforeignlanguage}[2]{{%
\expandafter\ifx\csname l@#1\endcsname\relax
\typeout{** WARNING: IEEEtran.bst: No hyphenation pattern has been}%
\typeout{** loaded for the language `#1'. Using the pattern for}%
\typeout{** the default language instead.}%
\else
\language=\csname l@#1\endcsname
\fi
#2}}
\providecommand{\BIBdecl}{\relax}
\BIBdecl

\bibitem{wisniewski2020application}
W.~Wisniewski, D.~Lindsay, and S.~Lindsay, ``Application of conformal
  prediction interval estimations to market makers’ net positions,'' in
  \emph{Proc. Symp. Conformal Probabilistic Prediction Appl.}, 2020, pp.
  285--301.

\bibitem{lekeufack2024conformal}
J.~Lekeufack, A.~N. Angelopoulos \emph{et~al.}, ``Conformal decision theory:
  Safe autonomous decisions from imperfect predictions,'' in \emph{Proc. IEEE
  Int. Conf. Robot. Automat.}, 2024, pp. 11\,668--11\,675.

\bibitem{Osvaldo2026COMG}
O.~Simeone, S.~Park, and M.~Zecchin, ``Conformal calibration: Ensuring the
  reliability of black-box {AI} in wireless systems,'' \emph{{IEEE} Commun.
  Mag.}, vol.~64, no.~8, pp. 179--187, 2026.

\bibitem{Park2026TPAMI}
Y.-J. Park, C.~Sobolewski, and N.~Azizan, ``Uncertainty quantification in
  detection transformers: Object-level calibration and image-level
  reliability,'' \emph{{IEEE} Trans. Pattern Anal. Mach. Intell.}, vol.~48,
  no.~11, pp. 13\,775--13\,788, 2026.

\bibitem{Wang2026TPAMI}
K.~Wang, F.~Cuzzolin, K.~Shariatmadar, D.~Moens, and H.~Hallez, ``A review of
  uncertainty representation and quantification in neural networks,''
  \emph{{IEEE} Trans. Pattern Anal. Mach. Intell.}, vol.~48, no.~3, pp.
  2476--2495, 2026.

\bibitem{Osvaldo2026JSAIT}
O.~Simeone and Y.~Romano, ``Uncertainty quantification and data efficiency in
  {AI}: An information-theoretic perspective,'' \emph{IEEE J. Sel. Areas Inf.
  Theory}, vol.~7, pp. 290--301, 2026.

\bibitem{angelopoulos2024theoretical}
A.~N. Angelopoulos, R.~F. Barber, and S.~Bates, ``Theoretical foundations of
  conformal prediction,'' \emph{arXiv preprint arXiv:2411.11824}, 2024.

\bibitem{Park2024TPAMI}
S.~Park, K.~M. Cohen, and O.~Simeone, ``Few-shot calibration of set predictors
  via meta-learned cross-validation-based conformal prediction,'' \emph{{IEEE}
  Trans. Pattern Anal. Mach. Intell.}, vol.~46, no.~1, pp. 280--291, 2024.

\bibitem{angelopoulos2024conformal}
A.~N. Angelopoulos, S.~Bates, A.~Fisch, L.~Lei, and T.~Schuster, ``Conformal
  risk control,'' in \emph{Proc. Int. Conf. Learn. Represent.}, 2024, pp.
  55\,198--55\,218.

\bibitem{Einbinder2025TPAMI}
B.-S. Einbinder, L.~Ringel, and Y.~Romano, ``Semi-supervised risk control via
  prediction-powered inference,'' \emph{{IEEE} Trans. Pattern Anal. Mach.
  Intell.}, vol.~47, no.~12, pp. 11\,080--11\,090, 2025.

\bibitem{blot2024automatically}
V.~Blot, A.~N. Angelopoulos, M.~Jordan, and N.~J.-B. Brunel, ``Automatically
  adaptive conformal risk control,'' in \emph{Proc. Int. Conf. Artif. Intell.
  Statist.}, 2025.

\bibitem{gibbs2021adaptive}
I.~Gibbs and E.~Candes, ``Adaptive conformal inference under distribution
  shift,'' in \emph{Proc. Adv. Neural Inf. Process. Syst.}, vol.~34, 2021, pp.
  1660--1672.

\bibitem{angelopoulos2024online}
A.~N. Angelopoulos, R.~Barber, and S.~Bates, ``Online conformal prediction with
  decaying step sizes,'' in \emph{Proc. Int. Conf. Mach. Learn.}, 2024, pp.
  1616--1630.

\bibitem{Xu2023TPAMI}
C.~Xu and Y.~Xie, ``Conformal prediction for time series,'' \emph{{IEEE} Trans.
  Pattern Anal. Mach. Intell.}, vol.~45, no.~10, pp. 11\,575--11\,587, 2023.

\bibitem{feldman2023achieving}
S.~Feldman, L.~Ringel, S.~Bates, and Y.~Romano, ``Achieving risk control in
  online learning settings,'' \emph{Trans. Mach. Learn. Res.}, 2023.

\bibitem{zecchin2024localized}
M.~Zecchin and O.~Simeone, ``Localized adaptive risk control,'' in \emph{Proc.
  Adv. Neural Inf. Process. Syst.}, vol.~37, 2024, pp. 8165--8192.

\bibitem{Zhu2025TSP}
M.~Zhu, M.~Zecchin, S.~Park, C.~Guo, C.~Feng, P.~Popovski, and O.~Simeone,
  ``Conformal distributed remote inference in sensor networks under reliability
  and communication constraints,'' \emph{{IEEE} Trans. Signal Process.},
  vol.~73, pp. 1485--1500, 2025.

\bibitem{Binucci2025TSP}
F.~Binucci, O.~Simeone, and P.~Banelli, ``Conformal {Lyapunov} optimization:
  Optimal resource allocation under deterministic reliability constraints,''
  \emph{{IEEE} Trans. Signal Process.}, vol.~73, pp. 4521--4538, 2025.

\bibitem{cifra2021bridging}
C.~L. Cifra, D.~F. Sittig, and H.~Singh, ``Bridging the feedback gap: a
  sociotechnical approach to informing clinicians of patients{\textquoteright}
  subsequent clinical course and outcomes,'' \emph{BMJ Qual. Saf.}, vol.~30,
  no.~7, pp. 591--597, 2021.

\bibitem{kwiatkowski2019natural}
T.~Kwiatkowski \emph{et~al.}, ``{Natural Questions}: A benchmark for question
  answering research,'' \emph{Trans. Assoc. Comput. Linguist.}, vol.~7, pp.
  452--466, 2019.

\bibitem{csaba2024label}
B.~Csaba, W.~Zhang, M.~M{\"u}ller, S.-N. Lim, M.~Elhoseiny, P.~H. Torr, and
  A.~Bibi, ``Label delay in online continual learning,'' in \emph{Proc. Adv.
  Neural Inf. Process. Syst.}, vol.~37, 2024, pp. 119\,976--120\,012.

\bibitem{pham2023continual}
Q.~Pham, C.~Liu, and S.~C. Hoi, ``Continual learning, fast and slow,''
  \emph{{IEEE} Trans. Pattern Anal. Mach. Intell.}, vol.~46, no.~1, pp.
  134--149, 2023.

\bibitem{wiesmayr2025salad}
R.~Wiesmayr, L.~Maggi, S.~Cammerer, J.~Hoydis, F.~A. Aoudia, and A.~Keller,
  ``{SALAD}: Self-adaptive link adaptation,'' \emph{arXiv preprint
  arXiv:2510.05784}, 2025.

\bibitem{chapelle2014modeling}
O.~Chapelle, ``Modeling delayed feedback in display advertising,'' in
  \emph{Proc. ACM SIGKDD Int. Conf. Knowl. Discov. Data Min.}, 2014, pp.
  1097--1105.

\bibitem{Hallberg2024PLMR}
J.~Hallberg~Szabadv\'{a}ry, ``Adaptive conformal inference for multi-step ahead
  time-series forecasting online,'' in \emph{Proc. Symp. Conformal
  Probabilistic Prediction Appl.}, 2024, pp. 250--263.

\bibitem{ding2025calibrated}
T.~Ding, I.~Gibbs, and R.~J. Tibshirani, ``Calibrated multi-level quantile
  forecasting,'' \emph{arXiv preprint arXiv:2512.23671}, 2025.

\bibitem{halabi2026adaptive}
L.~E. Halabi and A.~Brandt, ``Adaptive conformal inference under delayed
  feedback: Coverage guarantees and a delay-to-memory diagnostic,'' \emph{arXiv
  preprint arXiv:2609.07251}, 2026.

\bibitem{zhao2025conformalized}
M.~Zhao, H.~Admoni, R.~Simmons, A.~Ramdas, and A.~Bajcsy, ``Conformalized
  interactive imitation learning: Handling expert shift and intermittent
  feedback,'' in \emph{Proc. Int. Conf. Learn. Represent.}, 2025, pp.
  9238--9251.

\bibitem{wang2025mirror}
B.~Wang, M.~Zecchin, and O.~Simeone, ``Mirror online conformal prediction with
  intermittent feedback,'' \emph{{IEEE} Signal Process. Lett.}, vol.~32, pp.
  2888--2892, 2025.

\bibitem{orabona2019modern}
F.~Orabona, ``A modern introduction to online learning,'' \emph{arXiv preprint
  arXiv:1912.13213}, 2019.

\bibitem{shalev2025online}
S.~Shalev-Shwartz, ``Online learning and online convex optimization,''
  \emph{Found. Trends Mach. Learn.}, vol.~4, no.~2, pp. 107--194, 2025.

\bibitem{zinkevich2003online}
M.~Zinkevich, ``Online convex programming and generalized infinitesimal
  gradient ascent,'' in \emph{Proc. Int. Conf. Mach. Learn.}, 2003, pp.
  928--936.

\bibitem{langford2009slow}
J.~Langford, A.~Smola, and M.~Zinkevich, ``Slow learners are fast,''
  \emph{arXiv preprint arXiv:0911.0491}, 2009.

\bibitem{quanrud2015online}
K.~Quanrud and D.~Khashabi, ``Online learning with adversarial delays,'' in
  \emph{Proc. Adv. Neural Inf. Process. Syst.}, vol.~28, 2015.

\bibitem{wan2024non}
Y.~Wan, C.~Yao, M.~Song, and L.~Zhang, ``Non-stationary online convex
  optimization with arbitrary delays,'' in \emph{Proc. Int. Conf. Mach.
  Learn.}, 2024, pp. 49\,991--50\,011.

\bibitem{wu2024online}
P.~Wu, H.~Huang, and Z.~Liu, ``Online sequential decision-making with unknown
  delays,'' in \emph{Proc. ACM Web Conf.}, 2024, pp. 4028--4036.

\bibitem{qiu2025exploiting}
H.~Qiu, E.~Esposito, and M.~Zhang, ``Exploiting curvature in online convex
  optimization with delayed feedback,'' in \emph{Proc. Int. Conf. Mach.
  Learn.}, 2025, pp. 50\,448--50\,479.

\bibitem{ryabchenko2025capacity}
A.~Ryabchenko, I.~Attias, and D.~M. Roy, ``Capacity-constrained online learning
  with delays: Scheduling frameworks and regret trade-offs,'' in \emph{Proc.
  Conf. Learn. Theory}, 2025, pp. 4973--5014.

\bibitem{ryabchenko2026capacity}
------, ``Capacity-constrained online convex optimization with delayed
  feedback,'' \emph{arXiv preprint arXiv:2606.11711}, 2026.

\bibitem{angelopoulos2025gradient}
A.~N. Angelopoulos, M.~I. Jordan, and R.~J. Tibshirani, ``Gradient equilibrium
  in online learning: Theory and applications,'' \emph{J. Mach. Learn. Res.},
  vol.~26, no. 305, pp. 1--68, 2025.

\bibitem{Matteo2026JSAIT}
M.~Zecchin, U.~K. Ganesan, G.~Durisi, P.~Popovski, and O.~Simeone,
  ``Prediction-powered communication with distortion guarantees,'' \emph{IEEE
  J. Sel. Areas Inf. Theory}, vol.~7, pp. 33--45, 2026.

\bibitem{elvira2021advances}
V.~Elvira and L.~Martino, \emph{Advances in Importance Sampling}.\hskip 1em
  plus 0.5em minus 0.4em\relax John Wiley \& Sons, Ltd, 2021, pp. 1--14.

\bibitem{angelopoulos2023conformalge}
A.~N. Angelopoulos and S.~Bates, ``Conformal prediction: A gentle
  introduction,'' \emph{Found. Trends Mach. Learn.}, vol.~16, no.~4, pp.
  494--591, 2023.

\bibitem{deng2012hedging}
J.~Deng, J.~Krause, A.~C. Berg, and L.~Fei-Fei, ``Hedging your bets: Optimizing
  accuracy-specificity trade-offs in large scale visual recognition,'' in
  \emph{Proc. IEEE Conf. Comput. Vis. Pattern Recognit.}, 2012, pp. 3450--3457.

\bibitem{fan2020pranet}
D.-P. Fan, G.-P. Ji, T.~Zhou, G.~Chen, H.~Fu, J.~Shen, and L.~Shao, ``{PraNet}:
  Parallel reverse attention network for polyp segmentation,'' in \emph{Proc.
  Int. Conf. Med. Image Comput. Comput.-Assist. Intervent.}, 2020, pp.
  263--273.

\bibitem{karpukhin2020dense}
V.~Karpukhin, B.~Oguz, S.~Min, P.~Lewis, L.~Wu, S.~Edunov, D.~Chen, and W.-t.
  Yih, ``Dense passage retrieval for open-domain question answering,'' in
  \emph{Proc. Conf. Empir. Methods Nat. Lang. Process.}, 2020, pp. 6769--6781.

\bibitem{rajpurkar2016squad}
P.~Rajpurkar, J.~Zhang, K.~Lopyrev, and P.~Liang, ``{SQuAD}: 100,000+ questions
  for machine comprehension of text,'' in \emph{Proc. Conf. Empir. Methods Nat.
  Lang. Process.}, 2016, pp. 2383--2392.

\bibitem{deng2009imagenet}
J.~Deng, W.~Dong, R.~Socher, L.-J. Li, K.~Li, and L.~Fei-Fei, ``{ImageNet}: A
  large-scale hierarchical image database,'' in \emph{Proc. IEEE Conf. Comput.
  Vis. Pattern Recognit.}, 2009, pp. 248--255.

\bibitem{fellbaum1998wordnet}
C.~Fellbaum, \emph{WordNet: An electronic lexical database}.\hskip 1em plus
  0.5em minus 0.4em\relax MIT press, 1998.

\bibitem{he2016deep}
K.~He, X.~Zhang, S.~Ren, and J.~Sun, ``Deep residual learning for image
  recognition,'' in \emph{Proc. IEEE Conf. Comput. Vis. Pattern Recognit.},
  2016, pp. 770--778.

\bibitem{yao2019tsbatpro}
F.~Yao, J.~Wu, G.~Venkataramani, and S.~Subramaniam, ``{TS-BatPro}: Improving
  energy efficiency in data centers by leveraging temporal–spatial
  batching,'' \emph{IEEE Trans. Cogn. Commun. Netw.}, vol.~3, no.~1, pp.
  236--249, 2019.

\bibitem{chen2016dynamic}
W.~Chen, R.~F. da~Silva, E.~Deelman, and T.~Fahringer, ``Dynamic and
  fault-tolerant clustering for scientific workflows,'' \emph{IEEE Trans. Cloud
  Comput.}, vol.~4, no.~1, pp. 49--62, 2016.

\bibitem{kleinrock1975queueing}
L.~Kleinrock, \emph{Queueing Systems, Volume {I}: Theory}.\hskip 1em plus 0.5em
  minus 0.4em\relax New York: Wiley, 1975.

\bibitem{kalman2004modeling}
M.~Kalman and B.~Girod, ``Modeling the delays of successively-transmitted
  {Internet} packets,'' in \emph{Proc. IEEE Int. Conf. Multimedia and Expo
  (ICME)}, 2004, pp. 2015--2018.

\bibitem{severinson2022dsag}
A.~Severinson, E.~Rosnes, S.~El~Rouayheb, and A.~G. i~Amat, ``{DSAG}: A mixed
  synchronous-asynchronous iterative method for straggler-resilient learning,''
  \emph{{IEEE} Trans. Commun.}, vol.~71, no.~2, pp. 808--822, 2022.

\bibitem{liang2006network}
Y.~Liang and B.~Girod, ``Network-adaptive low-latency video communication over
  best-effort networks,'' \emph{{IEEE} Trans. Circuits Syst. Video Technol.},
  vol.~16, no.~1, pp. 72--81, 2006.

\bibitem{ktena2019addressing}
S.~I. Ktena, A.~Tejani, L.~Theis, P.~K. Myana, D.~Dilipkumar, F.~Husz{\'a}r,
  S.~Yoo, and W.~Shi, ``Addressing delayed feedback for continuous training
  with neural networks in {CTR} prediction,'' in \emph{Proc. ACM Conf. Recomm.
  Syst.}, 2019, pp. 187--195.

\bibitem{lewis1989gamma}
P.~A.~W. Lewis, E.~McKenzie, and D.~K. Hugus, ``Gamma processes,''
  \emph{Communications in Statistics. Stochastic Models}, vol.~5, no.~1, pp.
  1--30, 1989.

\end{thebibliography}


\begin{thebibliography}{1}
\providecommand{\url}[1]{#1}
\csname url@samestyle\endcsname
\providecommand{\newblock}{\relax}
\providecommand{\bibinfo}[2]{#2}
\providecommand{\BIBentrySTDinterwordspacing}{\spaceskip=0pt\relax}
\providecommand{\BIBentryALTinterwordstretchfactor}{4}
\providecommand{\BIBentryALTinterwordspacing}{\spaceskip=\fontdimen2\font plus
\BIBentryALTinterwordstretchfactor\fontdimen3\font minus
  \fontdimen4\font\relax}
\providecommand{\BIBforeignlanguage}[2]{{%
\expandafter\ifx\csname l@#1\endcsname\relax
\typeout{** WARNING: IEEEtran.bst: No hyphenation pattern has been}%
\typeout{** loaded for the language `#1'. Using the pattern for}%
\typeout{** the default language instead.}%
\else
\language=\csname l@#1\endcsname
\fi
#2}}
\providecommand{\BIBdecl}{\relax}
\BIBdecl

\bibitem{qiu2025exploiting}
H.~Qiu, E.~Esposito, and M.~Zhang, ``Exploiting curvature in online convex
  optimization with delayed feedback,'' in \emph{Proc. Int. Conf. Mach.
  Learn.}, 2025, pp. 50\,448--50\,479.

\bibitem{cao2020constrained}
X.~Cao, J.~Zhang, and H.~V. Poor, ``Constrained online convex optimization with
  feedback delays,'' \emph{{IEEE} Trans. Autom. Control}, vol.~66, no.~11, pp.
  5049--5064, 2020.

\bibitem{bowen2026distributed}
B.~Wang, M.~Zecchin, and O.~Simeone, ``Distributed dynamic associative memory
  via online convex optimization,'' \emph{{IEEE} J. Sel. Areas Commun.},
  vol.~44, pp. 5872--5887, 2026.

\bibitem{wang2025mirror}
------, ``Mirror online conformal prediction with intermittent feedback,''
  \emph{{IEEE} Signal Process. Lett.}, vol.~32, pp. 2888--2892, 2025.

\bibitem{zhao2025conformalized}
M.~Zhao, H.~Admoni, R.~Simmons, A.~Ramdas, and A.~Bajcsy, ``Conformalized
  interactive imitation learning: Handling expert shift and intermittent
  feedback,'' in \emph{Proc. Int. Conf. Learn. Represent.}, 2025, pp.
  9238--9251.

\bibitem{papaspiliopoulos2020high}
R.~Vershynin, \emph{High-Dimensional Probability: An Introduction with
  Applications in Data Science}.\hskip 1em plus 0.5em minus 0.4em\relax
  Cambridge University Press, 2018.

\bibitem{durrett2019probability}
R.~Durrett, \emph{Probability: theory and examples}.\hskip 1em plus 0.5em minus
  0.4em\relax Cambridge university press, 2019, vol.~49.

\end{thebibliography}

%
%

\end{document}


\title{Capacity-Constrained Delayed Feedback for Online Conformal Risk Control}

\author{
Bowen Wang,~\IEEEmembership{Student Member,~IEEE} and Osvaldo Simeone,~\IEEEmembership{Fellow,~IEEE}
}

\IEEEspecialpapernotice{(Supplementary Material)\vspace{-0.5em}}

\maketitle

\appendices
\counterwithin*{equation}{section}
\renewcommand{\theequation}{\thesection-\arabic{equation}}
\renewcommand{\theHequation}{\theHsection.\arabic{equation}}
\newcommand{\appendixnumbering}[1]{%
	\counterwithin*{#1}{section}%
	\expandafter\def\csname the#1\endcsname{\thesection-\arabic{#1}}%
	\expandafter\def\csname theH#1\endcsname{\theHsection.\arabic{#1}}%
}
\appendixnumbering{lemma}\appendixnumbering{theorem}\appendixnumbering{proposition}
\appendixnumbering{corollary}\appendixnumbering{definition}\appendixnumbering{remark}
\appendixnumbering{property}\appendixnumbering{insight}\appendixnumbering{assumption}
\section{Proofs of OCRC, I-OCRC, and CA-OCRC}\label{app:lem_baselines}

\subsection{Proof of Lemma~\ref{lem:unconstrained-coverage}}
Following \cite{qiu2025exploiting,cao2020constrained,bowen2026distributed}, we continue OCRC over $d_{\max}$ \emph{virtual rounds} $t=T+1,\ldots,T+d_{\max}$, where no new data point is observed, while the feedback of the rounds $t\le T$ with $t+d_t>T$ is received and the threshold is updated via \eqref{eq:unconstrained-update}.
Accordingly, $r_{T+1}, \cdots, r_{T+d_{\max}+1}$ are referred to as the \emph{virtual thresholds}.
Summing \eqref{eq:unconstrained-update} over rounds $t=1,\ldots,T+d_{\max}$ yields
\begin{equation}\label{eq:app-ocrc-identity}
	\sum_{t=1}^{T}(L_t-\alpha)=\frac{r_{T+d_{\max}+1}-r_1}{\eta} .
\end{equation}

Then, we turn to bounding $\{r_t\}_{t=1}^{T+d_{\max}+1}$.
Specifically, suppose that $r_{t+1}>B$ for some $t\le T+d_{\max}$, and let $\tau=\max\{j\le t:r_j\le B\}$.
For every round $j$ with $\tau<j\le \min\{t,T\}$, we have $r_j>B$, and hence $\mathcal{C}_j=\mathcal{Y}$ and $L_j=\mathcal{L}(\mathcal{Y},Y_j)=0$ by Assumption~\ref{assum:score} and Assumption~\ref{assum:loss}, so that the feedback of round $j$ decreases the threshold, while the virtual rounds $j>T$ produce no feedback of their own.
Therefore, every upward term $\eta(L_k-\alpha)$ of the updates \eqref{eq:unconstrained-update} applied in rounds $\tau,\ldots,t$ is due to a round $k\le\tau$ with $k+d_k\ge\tau$, i.e., $k\in\{\tau-d_{\max},\ldots,\tau\}$, and it is at most $\eta(1-\alpha)$, which yields
\begin{equation}
	\begin{aligned}
		r_{t+1} & \le r_\tau+\eta(1-\alpha)(d_{\max}+1) \\
		& \le B+\eta(1-\alpha)(d_{\max}+1) .
	\end{aligned}
\end{equation}

Symmetrically, we have the lower bound 
\begin{equation}
	r_{t+1}\ge-\eta\alpha(d_{\max}+1) .
\end{equation}

Since $r_1\in[0,B]$, combining the two bounds gives 
\begin{equation}\label{eq_A1_thbound}
	|r_{T+d_{\max}+1}-r_1|\le B+\eta(d_{\max}+1).
\end{equation}

Finally, dividing \eqref{eq:app-ocrc-identity} by $T$ and using \eqref{eq_A1_thbound} yields
\begin{equation}
	\left|\frac{1}{T}\sum_{t=1}^{T}L_t-\alpha\right|\le\frac{B}{\eta T}+ \frac{d_{\max}+1}{T} ,
\end{equation}
which completes the proof of Lemma~\ref{lem:unconstrained-coverage}.

\subsection{Proof of Lemma~\ref{lemm_i_ocp}}
Summing the update \eqref{eq_i_ocp} over rounds $t=1,\ldots,T$ yields
\begin{equation}\label{eq:app-iocrc-identity}
	\sum_{t=1}^{T}(L_t-\alpha)=\frac{r_{T+1}-r_1}{\eta}+\sum_{t=1}^{T}\left(1-\frac{\mathbbm{1}\{d_t=0\}}{{\Pr}_G[d_t=0]}\right)(L_t-\alpha) .
\end{equation}
Under Assumption~\ref{assum:delay}(b), each summand of the last sum in \eqref{eq:app-iocrc-identity} has zero mean given $d_1,\ldots,d_{t-1}$, and the last sum has zero mean \cite{wang2025mirror,zhao2025conformalized}.

Similarly to the proof of Lemma~\ref{lem:unconstrained-coverage}, the upward and downward terms of I-OCRC are $\eta(1-\alpha)/{\Pr}_G[d_t=0]$ and $\eta\alpha/{\Pr}_G[d_t=0]$, respectively, which yields the bound
\begin{equation}
	|r_{T+1}-r_1|\le B + \frac{\eta}{\Pr_G[d_t=0]} .
\end{equation}

Taking the expectation of \eqref{eq:app-iocrc-identity}, dividing by $T$, and using this bound yields \eqref{eq_unit_coverage} completes the proof of Lemma~\ref{lemm_i_ocp}.

\subsection{Proof of Theorem~\ref{thm_main_miscoverage}}\label{app:thm_main_miscoverage}

We first establish a general risk bound for CA-OCRC, which will be used in Appendix~\ref{app:thm_worstcase_miscoverage} and Appendix~\ref{app:thm_iid_miscoverage} to prove Theorem~\ref{thm:worstcase_miscoverage} and Theorem~\ref{thm_iid_miscoverage}, respectively.

\begin{theorem}\label{thm_main_miscoverage}
	Under Assumption~\ref{assum:delay}, for any proxy delay distribution $F$, the risk gap \eqref{eq_miscov} of CA-OCRC satisfies the upper bound
	\begin{equation}\label{eq:thm_main_miscoverage}
		\mathrm{RiskGap}(T)
		\le \frac{ \left| r_1 - \mathbb{E}\left[ r_{T+d_{\max}+1} \right] \right| }{\eta T}
		+ \frac{1}{T} \sum_{t = 1}^T \Pr\{|\mathcal{K}_t^{-}| = K\}
	\end{equation}
	for any sequence of data points.
	The expectation and the probability are taken with respect to the proxy delays under Assumption~\ref{assum:delay}(a), and also with respect to the feedback delays under Assumption~\ref{assum:delay}(b).

	$r_{T+d_{\max}+1}$ is the virtual threshold, as for OCRC in the proof of Lemma~\ref{lem:unconstrained-coverage} in Appendix~\ref{app:lem_baselines}.
	In virtual rounds, CA-OCRC only evicts the tracked rounds, receives the feedback of the rounds in the set $\mathcal{O}_t$, and updates the threshold as
	\begin{equation}\label{eq:r_tilde}
		r_{t+1} = r_t + \eta \sum_{k \in \mathcal{O}_t} \frac{L_k - \alpha}{q_G(\bar{d}_{\max})\,F(d_k)} , \text{ for } t = T+1, \ldots, T+d_{\max} .
	\end{equation}
\end{theorem}

\begin{proof}
Summing \eqref{eq_ca_ocp} over $t = 1,\dots,T$ and \eqref{eq:r_tilde} over the virtual rounds telescopes the threshold to
\begin{subequations}\label{eq:tele}
	\begin{align}
		r_{T + d_{\max} + 1} & = r_1 - \eta \sum_{t = 1}^T \varphi_t (\alpha - L_t)
		- \eta \sum_{t = 1}^{T + d_{\max}} \sum_{k \in \mathcal O_t} \frac{\alpha - L_k}{q_G(\bar{d}_{\max})\,F(d_k)}  \\
		& = r_1 - \eta \sum_{t = 1}^T w_t (\alpha - L_t) , \label{eq:tele3}
	\end{align}
\end{subequations}
where $w_t$ is the weight eventually assigned to the update $\alpha - L_t$ of round $t$, given by
\begin{equation}\label{eq:w}
	w_t = \varphi_t + (1 - \varphi_t) \frac{\mathbbm{1}\{|\mathcal{K}_t^-| < K\} \mathbbm{1}\{\bar{d}_t \ge d_t\}}{q_G(\bar{d}_{\max})\,F(d_t)} .
\end{equation}

Rearranging \eqref{eq:tele} gives
\begin{equation}\label{eq:identity}
	\sum_{t = 1}^T (\alpha - L_t) = \frac{r_1 - r_{T + d_{\max} + 1}}{\eta} + \Delta_T ,
\end{equation}
where 
\begin{equation}
	\Delta_T = \sum_{t = 1}^T (1 - w_t) (\alpha - L_t) = \sum_{i \in \mathcal{I}} \left(1 - \frac{\mathbbm{1}\{|\mathcal{K}_i^{-}| < K\} \mathbbm{1}\{\bar{d}_i \ge d_i\}}{q_G(\bar{d}_{\max})\,F(d_i)}\right) (\alpha - L_i) ,
\end{equation}
with $\mathcal{I} = \{t \in \{1, \ldots, T\} : \varphi_t = 0\}$ being the set of rounds with nontrivial prediction sets.

Noting that $\mathbbm{1} \{|\mathcal{K}_i^{-}| < K\} = 1 - \mathbbm{1} \{|\mathcal{K}_i^{-}| = K\}$, we split $\Delta_T = \Delta_{T}^{1} + \Delta_{T}^{2}$ with
\begin{align}
	\Delta_T^1 & = \sum_{i \in \mathcal{I}} \left( 1 - \frac{\mathbbm{1}\{\bar{d}_i \ge d_i\}}
	{q_G(\bar{d}_{\max})\,F(d_i)} \right) (\alpha - L_i) , \label{eq:MT} \\
	\Delta_{T}^{2} & = \sum_{i \in \mathcal{I}} \frac{\mathbbm{1}\{\bar{d}_i \ge d_i\} \mathbbm{1}\{|\mathcal{K}_i^{-}| = K\}}{q_G(\bar{d}_{\max})\,F(d_i)}(\alpha - L_i) . \label{eq:FT}
\end{align}

Let $\mathcal{H}_i$ denote the $\sigma$-field generated by the feedback delays $d_1, \ldots, d_{i-1}$ and by the proxy delays $\bar{d}_1, \ldots, \bar{d}_{i-1}$ of the previous rounds.
Then, the quantities $r_i$, $L_i$, $|\mathcal{K}_i^-|$, and $\mathbbm{1}\{i \in \mathcal{I}\}$ are determined by $\mathcal{H}_i$, while the proxy delay $\bar{d}_i$ is drawn from $F$ independently of $\mathcal{H}_i$ and of the feedback delay $d_i$, so that $\mathbb{E}[\mathbbm{1}\{\bar{d}_i \ge d_i\} \mid \mathcal{H}_i, d_i] = F(d_i)$.
Since $F(d_i)>0$ if and only if $d_i\le\bar{d}_{\max}$, and since $d_i$ is independent of $\mathcal{H}_i$, with ${\Pr}_G [d_i\le\bar{d}_{\max}]=q_G(\bar{d}_{\max})$, the two-level inverse probability weighting in \eqref{eq_ca_ocp} satisfies
\begin{equation}\label{eq:two-level-ipw}
	\mathbb{E}\left[\frac{\mathbbm{1}\{\bar{d}_i \ge d_i\}}{q_G(\bar{d}_{\max}) F(d_i)} \middle| \mathcal{H}_i\right]
	=\frac{{\Pr}_G [d_i\le\bar{d}_{\max}]}{q_G(\bar{d}_{\max})}
	=1 ,
\end{equation}
where the ratio inside the expectation is set to zero for $d_i>\bar{d}_{\max}$.
Under Assumption~\ref{assum:delay}(a), the feedback delay $d_i$ is deterministic, and \eqref{eq:two-level-ipw} holds since $\bar{d}_{\max}=d_{\max}$ and $q_G(\bar{d}_{\max})=1$.

Accordingly, each summand of \eqref{eq:MT} has zero mean given $\mathcal{H}_i$, and $\Delta_{T}^{1}$ is a sum of martingale differences with respect to $\{\mathcal{H}_i\}$.
Thus, we have $\mathbb{E}[\Delta_{T}^{1}] = 0$.

Similarly, by \eqref{eq:two-level-ipw}, given $\mathcal{H}_i$, each summand of \eqref{eq:FT} has mean $\mathbbm{1}\{|\mathcal{K}_i^-| = K\}(\alpha - L_i)$, whose absolute value is at most $\mathbbm{1}\{|\mathcal{K}_i^-| = K\}$, since $|\alpha - L_i| \le 1$ for $L_i\in[0,1]$.
Therefore, we have
\begin{equation}\label{eq:FT-bound}
	|\mathbb{E}[\Delta_{T}^{2}]| \le \sum_{t = 1}^T \Pr\{t \in \mathcal{I}, |\mathcal{K}_t^-| = K\}
	\le \sum_{t = 1}^T \Pr\{|\mathcal{K}_t^-| = K\} .
\end{equation}

Finally, taking the expectation of \eqref{eq:identity}, dividing by $T$, and using $\mathbb{E}[\Delta_T^1] = 0$ and \eqref{eq:FT-bound} completes the proof.
\end{proof}

\section{Proof of Theorem~\ref{thm:worstcase_miscoverage}}\label{app:thm_worstcase_miscoverage}

We prove the bound in Theorem \ref{thm:worstcase_miscoverage} by bounding the range of the threshold iterates and the saturation term in the following two subsections, respectively.
Throughout this appendix, we have $\bar{d}_{\max}=d_{\max}$ and $q_G(\bar{d}_{\max})=1$, as discussed in Sec.~\ref{sec:update}.

\subsection{Bounding the Range of Threshold Iterates}\label{app:lem_ca_ocp_bound}

\begin{lemma}\label{lem:ca_ocp_bound}
Under Assumption~\ref{assum:delay}(a), Assumption~\ref{assum:score}, and Assumption~\ref{assum:loss} with $r_1 \in [0, B]$, the thresholds of CA-OCRC satisfy the upper bound
\begin{equation}\label{eq:r_bound}
	\left| r_{t+1} - r_1 \right| \le B + \eta \frac{K + 1}{F(d_{\max})}
\end{equation}
for all $t \in \{1, \ldots, T + d_{\max}\}$ and for any sequence of data points, feedback delays, and proxy delays.
\end{lemma}

\begin{proof}
Each round $t \le T$ applies the update $-\eta(\alpha - L_t)$ if $\mathcal{C}_t \in \{\emptyset,\mathcal Y\}$, and the weighted update $-\eta(\alpha - L_k) / F(d_k)$ for each tracked round $k$ with $k + d_k = t$, while the virtual rounds apply only the latter updates.
Thus, each upward push is at most $\eta(1 - \alpha) / F(d_{\max})$, since $L \le 1$ and $F(d_k) \ge F(d_{\max})$.
Similarly to the proof of Lemma~\ref{lem:unconstrained-coverage}, at most $K + 1$ upward pushes can occur in any sequence of rounds, which gives
\begin{equation}
	\begin{aligned}
		r_{t + 1}
		& \le r_\tau + \eta(1 - \alpha) \frac{K + 1}{F(d_{\max})} \\
		& \le B + \eta(1 - \alpha) \frac{K + 1}{F(d_{\max})} .
	\end{aligned}
\end{equation}

Symmetrically, since $\mathcal{C}_j = \emptyset$ and $L_j = \mathcal{L}(\emptyset, Y_j) = 1$ whenever $r_j < 0$, we have the lower bound
\begin{equation}
	r_{t + 1} \ge -\eta\alpha \frac{K + 1}{F(d_{\max})}  .
\end{equation}

Since $r_1 \in [0, B]$, combining the upper and lower bounds yields \eqref{eq:r_bound}.
\end{proof}

\subsection{Bounding the Saturation Term}\label{app:lemma_sat_wc}

\begin{lemma}\label{lemma_sat_wc}
Under Assumption~\ref{assum:delay}(a) and Assumption~\ref{assum:wc-load}, the average probability of capacity saturation of CA-OCRC satisfies the upper bound
\begin{equation}
	\frac{1}{T}\sum_{t=1}^{T} {\Pr}_F \{|\mathcal{K}_t^{-}| = K\}
	\le e^{-K h(\ell_F / K)}
\end{equation}
for any sequence of data points and feedback delays, where $h(\cdot)$ is defined in Theorem~\ref{thm:worstcase_miscoverage}, and $\ell_F=\sum_{a=1}^{d_{\max}}F(a)=\mathbb{E}_F \left[(\bar{d})_+\right]$ is the tracking load.
\end{lemma}

\begin{proof}
For a given age $a$, round $t-a$ belongs to the set $\mathcal{K}_t^-$ only if it is admitted to tracking, i.e., $|\mathcal{K}_{t-a}^-| < K$,
its feedback is still pending, i.e., $d_{t-a} \ge a$, and its proxy delay has not expired, i.e., $\bar{d}_{t-a} \ge a$.
Accordingly, we have
\begin{subequations}
	\begin{align}
		|\mathcal{K}_t^{-}| & \le \sum_{a = 1}^{t - 1} \mathbbm{1}\{|\mathcal{K}_{t-a}^-| < K\} \mathbbm{1}\{d_{t - a} \ge a\} \mathbbm{1}\{\bar{d}_{t - a} \ge a\} \\
		& \le \sum_{a = 1}^{t - 1} \mathbbm{1}\{d_{t - a} \ge a\} \mathbbm{1}\{\bar{d}_{t - a} \ge a\} . \label{eq:PB-wc}
	\end{align}
\end{subequations}
Let $\bar{d}'_{t-a}\sim F$ denote the proxy delay that CA-OCRC draws at round $t-a$ if $\mathcal{C}_{t-a}$ were nontrivial, so that the variables $\bar{d}'_1,\ldots,\bar{d}'_{t-1}$ are independent, $\bar{d}_{t-a}=\bar{d}'_{t-a}$ for every round $t-a$ with a nontrivial prediction set, and $\mathbbm{1}\{\bar{d}_{t-a}\ge a\}=0\le\mathbbm{1}\{\bar{d}'_{t-a}\ge a\}$ otherwise.
Hence, the bound \eqref{eq:PB-wc} also holds with $\bar{d}_{t-a}$ replaced by $\bar{d}'_{t-a}$.
For a given sequence of feedback delays, the summands in \eqref{eq:PB-wc}, with $\bar{d}_{t-a}$ replaced by $\bar{d}'_{t-a}$, are independent Bernoulli random variables, whose sum has mean
\begin{equation}
	\mathbb{E}_F \left[ \sum_{a = 1}^{t - 1} \mathbbm{1}\{d_{t - a} \ge a\} \mathbbm{1}\{\bar{d}'_{t - a} \ge a\} \right]
	= \sum_{a = 1}^{t - 1} \mathbbm{1}\{d_{t - a} \ge a\} F(a)
	\le \sum_{a = 1}^{d_{\max}}F(a)
	= \ell_F .
\end{equation}
By Assumption~\ref{assum:wc-load} and the multiplicative Chernoff bound \cite{papaspiliopoulos2020high}, we have
\begin{equation}
	\begin{aligned}
		{\Pr}_F\{|\mathcal{K}_t^{-}| = K\}
		&\le {\Pr}_F\left\{\sum_{a = 1}^{t - 1}
		\mathbbm{1}\{d_{t - a} \ge a\}\mathbbm{1}\{\bar{d}'_{t - a} \ge a\} \ge K\right\} \\
		&\le e^{-\ell_F}
		\left(\frac{e\ell_F}{K}\right)^K \\
		&= \exp\left( -\ell_F + K + K \ln (\ell_F / K) \right) \\
		&= \exp\left( -K h(\ell_F / K) \right) .
	\end{aligned}
\end{equation}
Averaging this bound over $t$ completes the proof.
\end{proof}

\subsection{Overall Bound}
By Lemma~\ref{lem:ca_ocp_bound} with $t = T + d_{\max}$, the inequality
\begin{equation}
	|r_1 - r_{T + d_{\max} + 1}|
	\le B + \eta \frac{K + 1}{F(d_{\max})}
\end{equation}
holds for any realization of the proxy delays.
Hence, by Jensen's inequality \cite{durrett2019probability}, it also holds with $r_{T + d_{\max} + 1}$ replaced by $\mathbb{E}_F [r_{T + d_{\max} + 1}]$.

Furthermore, Lemma~\ref{lemma_sat_wc} gives
\begin{equation}
	\frac{1}{T} \sum_{t =1}^T {\Pr}_F \{|\mathcal{K}_t^{-}| = K\}
	\le e^{-K h(\ell_F / K)} .
\end{equation}

Substituting both bounds into \eqref{eq:thm_main_miscoverage} completes the proof.

\section{Proofs of Proposition~\ref{prop:wc-opt} and Corollary~\ref{cor:wc-rate}}\label{app:wc-design}


\subsection{Proof of Proposition~\ref{prop:wc-opt}}\label{app:prop_wc_opt}

We first rewrite the optimization problem as
\begin{subequations}\label{app_eq:wc-opt}
	\begin{align}
		\min_{\{F(a)\}_{a=-1}^{d_{\max}}} & \frac{K+1}{ F(d_{\max}) T } + e^{-K h \left( \ell_F / K\right)} \\
		\text{s.t.} \quad & \ell_F = \sum_{a=1}^{d_{\max}} F(a) \le K , \label{app_eq:wc-opt-b} \\
		& 1 = F(-1) \ge F(0) \ge \ldots \ge F(d_{\max}) \ge 0 .
	\end{align}
\end{subequations}

Then, we fix $F(d_{\max})=p\in(0,1]$.
Thus, the first term of the objective in problem~\eqref{app_eq:wc-opt} is fixed, while the second term depends on $F$ only through the tracking load $\ell_F$ in \eqref{app_eq:wc-opt-b}.
Since $F$ is nonincreasing, we have $F(a)\ge F(d_{\max})=p$ for all $a\in\{1, \cdots,d_{\max}\}$, and hence the tracking load satisfies the lower bound
\begin{equation}\label{eq:app-load-lb}
	\ell_F=\sum_{a=1}^{d_{\max}}F(a)\ge d_{\max} p .
\end{equation}
Furthermore, the value $F(0)\in[p,1]$ does not appear in the objective.

Since $h'(x)=1-1/x<0$ for $x\in(0,1)$, the function $h(x)$ is decreasing for $x\in(0,1]$, and the second term $e^{-Kh(\ell_F/K)}$ is thus nondecreasing in $\ell_F\in(0,K]$.
Accordingly, by \eqref{eq:app-load-lb}, for any fixed $p$, the objective of problem~\eqref{app_eq:wc-opt} is minimized by the proxy delay distribution with $F(1) = \cdots = F(d_{\max}) = p$, which yields $\ell_F=d_{\max}p$.

Therefore, problem~\eqref{app_eq:wc-opt} reduces to the one-dimensional problem
\begin{equation}\label{eq:app-h}
	J(p) = \frac{K+1}{pT}+e^{-Kh\left(d_{\max}p/K\right)} ,
\end{equation}
which completes the proof of Proposition~\ref{prop:wc-opt}.

\subsection{Proof of Corollary~\ref{cor:wc-rate}}\label{app:cor_wc_rate}

Using the definition of $J(p)$ in \eqref{eq:app-h}, we have
\begin{equation}\label{eq:app-cor-start}
	\mathrm{RiskGap}(T)\le\frac{B}{\eta T}+J(p^\star) .
\end{equation}
Since $p^\star$ minimizes $J(p)$ over $(0,K/d_{\max}]$, we have $J(p^\star)\le J(p)$ for the choice
\begin{equation}\label{eq:app-p-choice}
	p=\frac{K}{e\,d_{\max}}\left(\frac{d_{\max}}{T}\right)^{\frac{1}{K+1}},
\end{equation}
which satisfies $p\le K/d_{\max}$, since $d_{\max}\le T$.

For the choice \eqref{eq:app-p-choice}, the first term of $J(p)$ equals
\begin{equation}\label{eq:app-first}
	\frac{K+1}{pT}=\frac{e(K+1)}{K}\left(\frac{d_{\max}}{T}\right)^{\frac{K}{K+1}}\le 2e\left(\frac{d_{\max}}{T}\right)^{\frac{K}{K+1}},
\end{equation}
where the inequality follows from $(K+1)/K\le2$ for $K\ge1$.
Moreover, using the identity $e^{-Kh(x)}=(ex)^Ke^{-Kx}$ with $x=d_{\max}p/K=(d_{\max}/T)^{1/(K+1)}/e$, the second term of $J(p)$ satisfies
\begin{equation}\label{eq:app-second}
	e^{-Kh\left(d_{\max}p/K\right)}\le(ex)^K=\left(\frac{d_{\max}}{T}\right)^{\frac{K}{K+1}} .
\end{equation}
Substituting \eqref{eq:app-first} and \eqref{eq:app-second} into \eqref{eq:app-cor-start} yields
\begin{equation}
	\mathrm{RiskGap}(T) \le \frac{B}{\eta T} + (2e+1) \left( \frac{d_{\max}}{T} \right)^{\frac{K}{K+1}} ,
\end{equation}
which completes the proof of Corollary~\ref{cor:wc-rate}.

\section{Proof of Theorem~\ref{thm_iid_miscoverage}}\label{app:thm_iid_miscoverage}

Similar to the proof of Theorem~\ref{thm_main_miscoverage}, we first bound the range of the threshold iterates and the saturation term, and then combine the results to bound the risk gap.

\subsection{Bounding the Range of Threshold Iterates}\label{app:lem_Wbound}

\begin{lemma}\label{lem:dc_range}
	Under Assumption~\ref{assum:delay}(b), Assumption~\ref{assum:score}, and Assumption~\ref{assum:loss} with $r_1 \in [0, B]$, the thresholds of CA-OCRC satisfy the upper bound
	\begin{equation}\label{eq:dc-range}
		\left| \mathbb{E}_{F,G} [r_{t+1}] - r_1 \right| \le B + \eta R_{F,G}
	\end{equation}
	for all $t \in \{1, \ldots, T + d_{\max}\}$ and for any sequence of data points, where the expectation is taken with respect to both the proxy delays and the feedback delays,
	\begin{equation}\label{eq:dc-R}
		R_{F,G}=\bar{\alpha}\left(  \sum_{a = 0}^{\bar{d}_{\max}} a \frac{G(a) - G(a+1)}{q_G(\bar{d}_{\max})} +2\right)+\frac{K\bar{\alpha}^2}{2(1-\bar{\alpha})}\,Q_{F,G} ,
	\end{equation}
	with $\bar{\alpha} = \max\{\alpha, 1-\alpha\}$, and
	\begin{equation}\label{eq:dc-Q}
		Q_{F,G} =\sum_{a=0}^{\bar{d}_{\max}} \frac{ G(a) - G(a+1) }{q_G(\bar{d}_{\max})^2\,F(a)} .
	\end{equation}
\end{lemma}

\begin{proof}
	Let $\bar{d}'_k \sim F$ denote the proxy delay that CA-OCRC would draw at round $k$ if $\mathcal{C}_k$ were nontrivial, so that the pairs $(d_k, \bar{d}'_k)$ are independent across rounds, and let $\mathcal{G}_k$ denote the $\sigma$-field generated by the pairs $(d_i, \bar{d}'_i)$ with $i < k$.
	Thus, the loss $L_k$ and the indicator
	\begin{equation}
		\beta_k = \mathbbm{1}\{\mathcal{C}_k \notin \{\emptyset, \mathcal{Y}\}\} \mathbbm{1}\{|\mathcal{K}_k^-| < K\}
	\end{equation}
	are determined by $\mathcal{G}_k$. 
	Round $k$ is admitted if and only if $\beta_k = 1$ and $\bar{d}'_k \ge 0$, and its feedback is then received at round $k + d_k$ if and only if $\bar{d}'_k \ge d_k$.

	The feedback of an admitted round $k$ with $L_k > \alpha$ increases the threshold by $\eta(L_k-\alpha)\pi_k \le \eta(1-\alpha)\pi_k$, with 
	\begin{equation}
		\pi_k = \frac{\mathbbm{1}\{\bar{d}'_k \ge d_k\}}{q_G(\bar{d}_{\max})\,F(d_k)} ,
	\end{equation}
	where $\pi_k=0$ if $d_k>\bar{d}_{\max}$, since $\bar{d}'_k\le\bar{d}_{\max}$, while the feedback of an admitted round $k$ with $L_k \le \alpha$ does not increase the threshold.

	Fix $t \in \{1, \ldots, T + d_{\max}\}$, and let $T_t = \min\{t, T\}$ and $c = \alpha/(1-\alpha)$.
	Suppose that $r_{t+1} > B$, and let $\tau = \max\{j \le T_t : r_j \le B\}$, which is well defined since $r_1 \le B$.
	As in the proof of Lemma~\ref{lem:ca_ocp_bound}, every round $j$ with $\tau < j \le T_t$ applies the update $-\eta\alpha$, the virtual rounds apply no update of their own, and every upward push in rounds $\tau, \ldots, t$ is due either to round $\tau$ itself, if $\mathcal{C}_\tau = \emptyset$, or to the feedback of a round $k \in \mathcal{K}_\tau^+$ with $L_k > \alpha$, which is received at round $k + d_k \ge \tau$.
	Since $r_\tau \le B$ and $|\mathcal{K}_\tau^+| \le K$, we have
	\begin{equation}
		\frac{(r_{t+1} - B)_+}{\eta(1-\alpha)}
		\le 1 + \bigg[ \sum_{k \in \mathcal{K}_\tau^+} \pi_k - c \, (T_t - \tau) \bigg]_+
		\le 1 + \sum_{k \in \mathcal{K}_\tau^+} \left[ \pi_k - \frac{c}{K} (T_t - \tau) \right]_+ .
	\end{equation}
	Since every round $k \in \mathcal{K}_\tau^+$ has $\beta_k = 1$, and since $T_t - \tau \ge (T_t - k - d_k)_+$ if $\pi_k > 0$, extending the sum to all rounds $k \le T_t$ yields
	\begin{equation}\label{eq:dc-charge}
		\frac{(r_{t+1} - B)_+}{\eta(1-\alpha)} \le 1 + \sum_{k=1}^{T_t} \beta_k \left[ \pi_k - \frac{c}{K} (T_t - k - d_k)_+ \right]_+ ,
	\end{equation}
	which also holds trivially if $r_{t+1} \le B$.

	Since $\beta_k$ is determined by $\mathcal{G}_k$ and the pair $(d_k, \bar{d}'_k)$ is independent of $\mathcal{G}_k$, the expectation of the $k$th summand in \eqref{eq:dc-charge} is at most $\sum_{a=0}^{\bar{d}_{\max}} \Pr_G [d_t = a\mid d_t\le\bar{d}_{\max}] [1 - c q_G(\bar{d}_{\max}) F(a) (T_t - k - a)_+ / K]_+$, since, given $d_k=a\le\bar{d}_{\max}$, we have $\pi_k=1/(q_G(\bar{d}_{\max}) F(a))$ with probability $F(a)$, and $\pi_k=0$ otherwise.
	For each $a$, at most $a + 1$ rounds $k \le T_t$ satisfy $T_t - k - a \le 0$, while, for the remaining rounds, the bracket is evaluated at distinct positive integers $m = T_t - k - a$, and $\sum_{m \ge 1} [1 - \epsilon m]_+ \le 1/(2\epsilon)$ for $\epsilon > 0$.
	Summing over $k$ thus gives
	\begin{equation}\label{eq:dc-up}
		\mathbb{E}_{F,G} \left[ (r_{t+1} - B)_+ \right] \le \eta(1-\alpha) \left( \mathbb{E}_{G}[d_t\mid d_t\le\bar{d}_{\max}] + 2 + \frac{K (1-\alpha)}{2 \alpha} \, Q_{F,G} \right) .
	\end{equation}
	Symmetrically, with the roles of $\mathcal{C}_j = \emptyset$ and $\mathcal{C}_j = \mathcal{Y}$ exchanged and $c$ replaced by $1/c$, we have
	\begin{equation}\label{eq:dc-down}
		\mathbb{E}_{F,G} \left[ (-r_{t+1})_+ \right] \le \eta\alpha \left( \mathbb{E}_{G} [d_t\mid d_t\le\bar{d}_{\max}] + 2 + \frac{K \alpha}{2 (1-\alpha)} \, Q_{F,G} \right) .
	\end{equation}
	Finally, since $r_1 \in [0, B]$, we have $|\mathbb{E}_{F,G}[r_{t+1}] - r_1| \le B + \max\{\mathbb{E}_{F,G}[(r_{t+1} - B)_+], \mathbb{E}_{F,G}[(-r_{t+1})_+]\}$.
	The right-hand sides of \eqref{eq:dc-up} and \eqref{eq:dc-down} equal $\eta g(1-\alpha)$ and $\eta g(\alpha)$, respectively, with $g(x) = x(\mathbb{E}_{G}[d_t\mid d_t\le\bar{d}_{\max}] + 2) + K x^2 Q_{F,G}/(2(1-x))$, which is increasing in $x \in (0, 1)$, so that the maximum equals $\eta g(\bar{\alpha}) = \eta R_{F,G}$.
\end{proof}

\subsection{Bounding the Saturation Term}\label{app:lem_iid_saturation}

\begin{lemma}\label{lem_iid_saturation}
Under Assumption~\ref{assum:delay}(b) and Assumption~\ref{assum:stoch-load}, the average probability of capacity saturation of CA-OCRC satisfies the upper bound
\begin{equation}\label{eq:iid-sat}
	\frac{1}{T}\sum_{t=1}^{T} {\Pr}_{F,G}	\{|\mathcal{K}_t^{-}| = K\}
	\le \mathbbm{1}\{\bar{d}_{\max}\ge K\}\,e^{-K h(\ell_{F,G}/K)}
\end{equation}
for any sequence of data points, where $h(\cdot)$ is defined in Theorem~\ref{thm:worstcase_miscoverage}, and $\ell_{F,G}$ is defined in \eqref{eq:stoch-load}.
\end{lemma}
\begin{proof}

Based on the proxy delays, it holds that
\begin{equation}
	|\mathcal{K}_t^{-}| \le \sum_{a = 1}^{t - 1}
	\mathbbm{1}\{d_{t - a} \ge a\} \mathbbm{1}\{\bar d_{t - a} \ge a\} .
	\label{eq:PB}
\end{equation}
Since $\mathbbm{1}\{\bar{d}_{t-a}\ge a\}=0$ for $a>\bar{d}_{\max}$, the bound \eqref{eq:PB} implies $|\mathcal{K}_t^-|\le\bar{d}_{\max}$, and hence ${\Pr}_{F,G}\{|\mathcal{K}_t^-|=K\}=0$ if $\bar{d}_{\max}<K$.
In the following, we thus consider the case $\bar{d}_{\max}\ge K$.
As in the proof of Lemma~\ref{lemma_sat_wc}, the bound \eqref{eq:PB} also holds with $\bar{d}_{t-a}$ replaced by the proxy delay $\bar{d}'_{t-a}\sim F$ that CA-OCRC would draw at round $t-a$ if $\mathcal{C}_{t-a}$ were nontrivial, and, by Assumption~\ref{assum:delay}(b), the resulting summands are independent Bernoulli random variables.

The mean of the right-hand side of \eqref{eq:PB}, with this replacement, is $\sum_{a = 1}^{t - 1} G(a)F(a) \le \ell_{F,G}$.
The multiplicative Chernoff bound for sums of independent Bernoulli variables, $\Pr\{x \ge R\} \le e^{-\nu}(e\nu / R)^{R}$, is valid for $R \ge \nu$ and nondecreasing in $\nu \in [0,R]$.
So, since $\ell_{F,G} \le K$ by Assumption~\ref{assum:stoch-load}, applying the Chernoff bound with $\nu = \ell_{F,G}$ and $R = K$ gives
\begin{equation}
	\begin{aligned}
		{\Pr}_{F,G} \{|\mathcal{K}_t^{-}| = K\}
		& \le e^{-\ell_{F,G}}\left(\frac{e \ell_{F,G}}{K}\right)^{K} \\
		& = \exp\big(-\ell_{F,G} + K + K \ln (\ell_{F,G}/K)\big) \\
		& = \exp\big(-K h(\ell_{F,G}/K)\big) ,
	\end{aligned}
	\label{eq:chernoff-alg}
\end{equation}
for every $t$, and averaging over $t$ completes the proof.
\end{proof}

\subsection{Overall Bound}

Lemma~\ref{lem:dc_range} with $t = T + d_{\max}$ and $r_1 \in [0, B]$ therefore yields
\begin{equation}
	|r_1 - \mathbb{E}_{F,G} [r_{T + d_{\max} + 1}]|
	\le B + \eta R_{F,G} .
\end{equation}

By Lemma~\ref{lem_iid_saturation}, we have
\begin{equation}
	\frac{1}{T}\sum_{t = 1}^T {\Pr}_{F,G} \{|\mathcal{K}_t^-| = K\}
	\le \mathbbm{1}\{\bar{d}_{\max}\ge K\}\,e^{-K h\left( \ell_{F,G} / K \right)}
\end{equation}

Substituting the above two bounds into Theorem~\ref{thm_main_miscoverage} proves Theorem~\ref{thm_iid_miscoverage}.

\section{Proofs of Proposition~\ref{prop:iid-opt} and Corollary~\ref{cor:iid-rate}}\label{app:iid-design}
In this appendix, we prove Proposition~\ref{prop:iid-opt} and Corollary~\ref{cor:iid-rate}, which characterize the optimized proxy delay distribution and the resulting scaling law under stochastic delays, respectively.

\subsection{Proof of Proposition~\ref{prop:iid-opt}}
\label{app:prop_iid_opt_pareto}

If $\bar{d}_{\max}<K$, the optimization problem is given by
\begin{equation}
	\begin{aligned}
		\min_{F} \quad & Q_{F,G}=\sum_{a=0}^{\bar{d}_{\max}}\frac{{\Pr}_G [d_t=a]}{q_G(\bar{d}_{\max})^2\,F(a)}  \\
		\mathrm{s.t.} \quad & 1=F(-1)\ge F(0)\ge\cdots\ge F(\bar{d}_{\max})>0 ,
	\end{aligned}
\end{equation}
whose optimal solution is given by $F^\star (a)=1$ for $a\in\{-1,0,\ldots,\bar{d}_{\max}\}$.

In the rest of this proof, we consider the case $\bar{d}_{\max}\ge K$, where the optimization problem is rewritten as
\begin{subequations}\label{app_eq:iid-opt}
	\begin{align}
		\min_{\{F(a)\}_{a=-1}^{\bar{d}_{\max}}} \; & \frac{R_{F,G}}{T} + e^{-Kh\left(\ell_{F,G}/K\right)} \label{app_eq:iid-opt-obj} \\
		\text{s.t.} \quad & \ell_{F,G}=\sum_{a=1}^{\bar{d}_{\max}}F(a)\,G(a)\le K , \label{app_eq:iid-opt-cap} \\
		& 1=F(-1)\ge F(0)\ge\ldots\ge F(\bar{d}_{\max})>0 . \label{app_eq:iid-opt-mono}
	\end{align}
\end{subequations}
To solve problem~\eqref{app_eq:iid-opt}, we first reformulate it into an equivalent convex optimization problem and then derive its closed-form solution in the following four steps:


\subsubsection{Transformation}
The value $F(0)$ only appears in the objective \eqref{app_eq:iid-opt-obj} through $\Pr[d_t=0]/(q_G(\bar{d}_{\max})^2\,F(0))$ in $Q_{F,G}$. This term is nonincreasing in $F(0)$, hence, setting $F(0)=1$ is optimal.

Define the vector $\bm{x}=(x_1,\ldots,x_{\bar{d}_{\max}})$ with $x_a=F(a)$, which takes values in the convex set
\begin{equation}
    \mathcal{X} = \left\{ \bm{x}: 1 \ge x_1 \ge \cdots \ge x_{\bar{d}_{\max}} > 0 \right\}.
\end{equation}
Define the functions
\begin{align}\label{eq:app-fg}
    f(\bm{x}) & =q_G(\bar{d}_{\max})^2\,Q_{F,G} ={\Pr}_G [d_t=0] + \sum_{a=1}^{\bar{d}_{\max}} \frac{G(a) \bar{\kappa}(a)}{x_a} ,\\
    g(\bm{x}) & = \ell_{F,G} = \sum_{a=1}^{\bar{d}_{\max}}G(a) x_a ,
\end{align}
where $f(\bm{x})$ is convex and $g(\bm{x})$ is linear over $\mathcal{X}$.

\subsubsection{Convex Surrogate Problem}
For any $\mu>0$, consider the following surrogate problem:
\begin{subequations}\label{eq:app-surrogate}
    \begin{align}
        \min_{\bm{x}} \quad & f(\bm{x}) + \mu\,g(\bm{x}) \\
        \mathrm{s.t.} \quad & \bm{x}\in\mathcal{X} .
    \end{align}
\end{subequations}
Since $f(\bm{x})$ is convex, $g(\bm{x})$ is linear, and $\mathcal{X}$ is convex, problem~\eqref{eq:app-surrogate} is convex for any fixed $\mu>0$.

\subsubsection{Closed-Form Solution}
For fixed $\mu>0$, the objective of problem~\eqref{eq:app-surrogate} can be written as
\begin{equation}
    f(\bm{x}) = {\Pr}_G [d_t=0] + \sum_{a=1}^{\bar{d}_{\max}} G(a) \left( \frac{\bar{\kappa}(a)}{x_a} + \mu x_a \right).
\end{equation}

Define $z_a=\mu x_a^2$, which is strictly increasing in $x_a>0$. Each summand can be expressed as
\begin{equation}
    G(a)\left(\frac{\bar{\kappa}(a)}{x_a}+\mu x_a\right)
    =\sqrt{\mu}\,G(a)
    \left[D_{\Phi}(\bar{\kappa}(a),z_a)-\Phi(\bar{\kappa}(a))\right],
\end{equation}
where
\begin{equation}
    D_{\Phi}(y,z)
    = \Phi(y) - \Phi(z) - \Phi'(z)(y-z)
\end{equation}
is the Bregman divergence associated with the strictly convex function $\Phi(z)=-2\sqrt{z}$.

Since $\Phi(\bar{\kappa}(a))$ does not depend on $z_a$, problem~\eqref{eq:app-surrogate} is equivalent to the weighted Bregman isotonic regression problem
\begin{subequations}\label{eq:app-bregman-iso}
    \begin{align}
        \min_{\bm{z}} \quad & \sum_{a=1}^{\bar{d}_{\max}} G(a) D_{\Phi}(\bar{\kappa}(a),z_a)\\
        \mathrm{s.t.} \quad & \mu \ge z_1 \ge \cdots \ge z_{\bar{d}_{\max}}>0 ,
    \end{align}
\end{subequations}
whose optimal solution is given by
\begin{equation}
    z_a=\min\{\mu,\kappa(a)\},
\end{equation}
with 
\begin{equation}\label{appeq:hazard-iso}
		\kappa(a)=\min_{1\le s\le a} \; \max_{a\le s'\le\bar{d}_{\max}} \frac{ G(s) - G(s'+1) }{\sum_{i=s}^{s'}G(i)} .
	\end{equation}

Thus, the solution to problem~\eqref{eq:app-surrogate} is
\begin{equation}\label{eq:app-solution-mu}
    x_{\mu,a} = \min \left\{ 1 , \sqrt{\frac{\kappa(a)}{\mu}} \right\} .
\end{equation}

\subsubsection{Reduction to One-Dimensional Search}
We now show that the family of solutions in \eqref{eq:app-solution-mu} contains an optimal solution to the original problem.

Note that $G(a)\ge{\Pr}_G [d_t=\bar{d}_{\max}]>0$ for $a=1,\ldots,\bar{d}_{\max}$, which implies that  $\bar{\kappa}(a)\ge0$ and $\kappa(\bar{d}_{\max})>0$. 
Based on \eqref{appeq:hazard-iso}, it is straightforward to verify that $\kappa(a) > 0$.

By \eqref{eq:app-solution-mu}, the tracking load of the surrogate solution is
\begin{equation}\label{eq:app-load-mu}
    g(\bm{x}_{\mu}) = \sum_{a=1}^{\bar{d}_{\max}} G(a) \min\left\{ 1 , \sqrt{\frac{\kappa(a)}{\mu}} \right\}.
\end{equation}
This function is continuous and nonincreasing in $\mu>0$. 
Thus, we have
\begin{equation}
    \begin{aligned}
		\lim_{\mu\to0^+}g(\bm{x}_{\mu}) & =\sum_{a=1}^{\bar{d}_{\max}}G(a) = \mathbb{E}_G [\min\{d_t,\bar{d}_{\max}\}] , \\
		\lim_{\mu\to\infty}g(\bm{x}_{\mu}) & = 0 .
	\end{aligned}
\end{equation}
Besides, for any $\bm{x}\in\mathcal{X}$, we have $0 < g(\bm{x}) \le \mathbb{E}_G [\min\{d_t,\bar{d}_{\max}\}]$.
Therefore, by continuity, there exists $\mu>0$ such that
\begin{equation}\label{eq:app-load-match}
    g(\bm{x}_{\mu})=g(\bm{x}).
\end{equation}

Since $\bm{x}_{\mu}$ minimizes
$f(\bm{x})+\mu g(\bm{x})$ over $\mathcal{X}$, we have
\begin{equation}
    f(\bm{x}_{\mu}) + \mu g(\bm{x}_{\mu})
    \le f(\bm{x}) + \mu g(\bm{x}).
\end{equation}
Combining this inequality with \eqref{eq:app-load-match} yields
\begin{equation}\label{eq:app-dominance}
    f(\bm{x}_{\mu})\le f(\bm{x}).
\end{equation}

The term $R_{F,G}$ is nondecreasing in $f(\bm{x})$, while the exponential term depends on $\bm{x}$ only through $g(\bm{x})$. Hence, \eqref{eq:app-load-match} and \eqref{eq:app-dominance} imply
\begin{equation}\label{eq:app-objective-dom}
    \frac{R_{F_{\mu},G}}{T} + e^{-Kh(g(\bm{x}_{\mu})/K)}
    \le \frac{R_{F,G}}{T} + e^{-Kh(g(\bm{x})/K)}.
\end{equation}

Moreover, if $\bm{x}$ satisfies the tracking-load constraint
$g(\bm{x})\le K$, then $\bm{x}_{\mu}$ also satisfies this constraint by \eqref{eq:app-load-match}.
Thus, for every feasible solution to problem~\eqref{app_eq:iid-opt}, there exists a feasible member of the surrogate family with the same tracking load and no larger objective value. 

Finally, substituting $\theta=1/\sqrt{\mu}$ into \eqref{eq:app-solution-mu} gives
\begin{equation}
    x_{\theta,a} = \min\{ 1, \theta \sqrt{\kappa(a)} \}.
\end{equation}
Accordingly, problem~\eqref{app_eq:iid-opt} is equivalently reduced to the one-dimensional problem
\begin{subequations}\label{eq:app-line}
    \begin{align}
        \min_{\theta>0} \quad & \frac{R_{F_{\theta},G}}{T} + e^{-Kh\left(\ell_{F_{\theta},G}/K\right)} \\
        \mathrm{s.t.} \quad & \ell_{F_{\theta},G}\le K,
    \end{align}
\end{subequations}
where $F_{\theta}$ denotes the proxy law in \eqref{eq:iid_opt_F} with $\theta^\star=\theta$.

Therefore, a global optimizer $\theta^\star$ of \eqref{eq:app-line} yields a global optimizer of problem~\eqref{app_eq:iid-opt}, which completes the proof of Proposition~\ref{prop:iid-opt}.

\subsection{Proof of Corollary~\ref{cor:iid-rate}}\label{app:cor_iid_rate}
Since the optimized maximum proxy delay $\bar{d}_{\max}$ and the proxy delay distribution in Proposition~\ref{prop:iid-opt} minimize the bound \eqref{eq:iid-bound}, it suffices to bound \eqref{eq:iid-bound} for $\bar{d}_{\max}=d_{\max}$, $F(0)=1$, and $F(a)=p$ for $a\in\{1,\ldots,d_{\max}\}$, with $p\in(0,1]$ and $p\,\mathbb{E}_G[d_t]\le K$.
For this choice, we have $q_G(\bar{d}_{\max})=1$, $\mathbb{E}_G[d_t\mid d_t\le\bar{d}_{\max}]=\mathbb{E}_G[d_t]$, $Q_{F,G}\le1/p$, and $\ell_{F,G}=p\,\mathbb{E}_G[d_t]$.
Therefore, by Theorem~\ref{thm_iid_miscoverage}, we have
\begin{equation}\label{eq:app-iid-cor-start}
	\mathrm{RiskGap}(T)\le\frac{B}{\eta T}+\frac{\bar{\alpha}(\mathbb{E}_G[d_t]+2)}{T}+\frac{\bar{\alpha}^2}{2(1-\bar{\alpha})}\,\frac{K}{pT}+e^{-Kh\left(p\,\mathbb{E}_G[d_t]/K\right)} .
\end{equation}

Similar to the proof of Corollary~\ref{cor:wc-rate} in Appendix~\ref{app:cor_wc_rate}, we choose
\begin{equation}\label{eq:app-iid-p}
	p=\frac{K}{e \mathbb{E}_G[d_t]}\left(\frac{\mathbb{E}_G[d_t]}{T}\right)^{\frac{1}{K+1}} ,
\end{equation}
which satisfies $p\le1$ and $p\,\mathbb{E}_G[d_t]\le K$, since $K\le\mathbb{E}_G[d_t]\le T$, and which yields
\begin{equation}\label{eq:app-iid-sat}
	e^{-Kh\left(p\,\mathbb{E}_G[d_t]/K\right)}\le(ex)^K=\left(\frac{\mathbb{E}_G[d_t]}{T}\right)^{\frac{K}{K+1}} .
\end{equation}
Substituting these bounds into \eqref{eq:app-iid-cor-start}, together with $\mathbb{E}_G[d_t]/T\le(\mathbb{E}_G[d_t]/T)^{K/(K+1)}$, which completes the proof of Corollary~\ref{cor:iid-rate}.

\footnotesize
\bibliographystyle{IEEEtran}
\bibliography{IEEEabrv,./ref/ref.bib}
